\documentclass[10pt,twocolumn,letterpaper]{article}

\usepackage{iccv}
\usepackage{times}
\usepackage{epsfig}
\usepackage{graphicx}
\usepackage{amsmath}
\usepackage{amssymb}

\usepackage{listings}
\usepackage{numprint}
\usepackage{nicefrac}
\usepackage{hyphenat}
\usepackage{url}
\usepackage{multirow}
\usepackage{multicol}
\usepackage{gensymb}
\usepackage{diagbox}
\usepackage{multicol} 
\usepackage{rotating}
\usepackage{pifont}
\usepackage{relsize}
\usepackage[dvipsnames, svgnames, table, xcdraw]{xcolor}
\usepackage{colortbl}
\usepackage{multirow}
\usepackage{wrapfig}
\usepackage[labelfont=bf]{caption}
\usepackage{tabularx}
\usepackage{utfsym}
\usepackage{comment}
\usepackage{lipsum}  
\usepackage{booktabs}
\usepackage{balance}
\usepackage{float}
\usepackage{subcaption}
\usepackage{graphicx}

\definecolor{Yellow}{rgb}{1.0, 1.0, 0.6}
\definecolor{Orange}{rgb}{1.0, 0.8, 0.6}
\definecolor{Red}{rgb}{1.0, 0.6, 0.6}
\definecolor{brightpink}{rgb}{1.0, 0.0, 0.5}
\definecolor{jade}{rgb}{0.0, 0.66, 0.42}
\definecolor{pastelbrown}{rgb}{0.51, 0.41, 0.33}
\definecolor{palegreen}{HTML}{CFFDBC}

\definecolor{brightpink}{rgb}{1.0, 0.0, 0.5}
\definecolor{brightgreen}{rgb}{0.4, 1.0, 0.0}
\definecolor{cadmiumyellow}{rgb}{1.0, 0.96, 0.0}
\definecolor{canaryyellow}{rgb}{1.0, 0.94, 0.0}
\definecolor{caribbeangreen}{rgb}{0.0, 0.8, 0.6}
\definecolor{citrine}{rgb}{0.89, 0.82, 0.04}
\definecolor{chromeyellow}{rgb}{1.0, 0.65, 0.0}
\definecolor{classicrose}{rgb}{0.98, 0.8, 0.91}
\definecolor{cherryblossompink}{rgb}{1.0, 0.72, 0.77}
\definecolor{carnationpink}{rgb}{1.0, 0.65, 0.79}
\definecolor{candypink}{rgb}{0.89, 0.44, 0.48}
\definecolor{applegreen}{rgb}{0.55, 0.71, 0.0}
\definecolor{ao}{rgb}{0.0, 0.5, 0.0}
\definecolor{brightgreen}{rgb}{0.4, 1.0, 0.0}
\definecolor{caribbeangreen}{rgb}{0.0, 0.8, 0.6}

\definecolor{lightgreen}{RGB}{197,224,180}
\definecolor{lightblue}{RGB}{222,235,247}
\definecolor{lightpurple}{RGB}{238,229,241}
\definecolor{lightorg}{RGB}{251,229,214}
\definecolor{lightorange}{rgb}{1.0, 0.8, 0.6}
\definecolor{olive}{rgb}{0.6, 0.6, 0.2}
\definecolor{sand}{rgb}{0.8666666666666667, 0.8, 0.4666666666666667}
\definecolor{wine}{rgb}{0.5333333333333333, 0.13333333333333333, 0.3333333333333333}
\definecolor{deblue}{RGB}{11,132,147}
\definecolor{ocra}{RGB}{204, 119, 34}
\definecolor{electricindigo}{rgb}{0.44, 0.0, 1.0}
\definecolor{lightblue}{RGB}{240,245,255}
\definecolor{darkblue}{RGB}{40,40,85}
\definecolor{babyblue}{rgb}{0.54, 0.81, 0.94}
\definecolor{pearDark}{HTML}{2980B9}
\definecolor{pearDarker}{HTML}{1D2DEC}
\definecolor{indigo(web)}{rgb}{0.29, 0.0, 0.51}
\definecolor{carouselpink}{HTML}{F9E0ED}
\definecolor{paleblue}{HTML}{CFECEC}
\definecolor{tablelightyellow}{rgb}{1,1, 0.8}
\definecolor{tableyellow}{HTML}{FFF1B5}
\definecolor{tableorange}{HTML}{CBDBD6}
\definecolor{tablered}{HTML}{FAC5DD}
\definecolor{soma}{HTML}{DC9043}
\definecolor{ours}{rgb}{0.77, 0.33, 0.77}
\definecolor{plain}{HTML}{FF889F}
\definecolor{slerp}{HTML}{1864ab}
\definecolor{lerp}{HTML}{a61e4d}
\definecolor{random}{HTML}{cdc988}
\definecolor{unrealistic_results}{HTML}{0b7285}
\definecolor{missing}{HTML}{c92a2a}

\lstdefinestyle{mystyle}{
    commentstyle=\color{codegreen},
    keywordstyle=\color{magenta},
    numberstyle=\tiny\color{codegray},
    stringstyle=\color{codepurple},
    basicstyle=\ttfamily\footnotesize,
    breakatwhitespace=false,         
    breaklines=true,                 
    captionpos=b,                    
    keepspaces=true,                 
    numbers=left,                    
    numbersep=5pt,                  
    showspaces=false,                
    showstringspaces=false,
    showtabs=false,                  
    tabsize=2
}

\newcommand{\first}[1]{\textbf{#1}\cellcolor{tablered}}
\newcommand{\second}[1]{#1\cellcolor{tableorange}}
\newcommand{\third}[1]{#1\cellcolor{tableyellow}}
\newcommand{\up}[1]{#1$\uparrow$}
\newcommand{\down}[1]{#1$\downarrow$}

\usepackage{float}

\usepackage[pagebackref=true,breaklinks=true,letterpaper=true,colorlinks,bookmarks=false]{hyperref}

\iccvfinalcopy %

\def\iccvPaperID{21} %

\newcommand{\pose}{\boldsymbol{\theta}}
\newcommand{\shape}{\boldsymbol{\beta}}
\newcommand{\transform}{\mathbf{T}}

\newcommand{\Error}{\mathcal{E}}
\newcommand{\Loss}{\mathcal{L}}
\newcommand{\Heatmap}{\mathbf{H}}
\newcommand{\body}{\mathcal{B}}
\newcommand{\vertices}{\mathbf{v}}
\newcommand{\normals}{\mathbf{n}}
\newcommand{\faces}{\mathbf{f}}
\newcommand{\mesh}{(\vertices, \faces)}
\newcommand{\regress}{r}
\newcommand{\regressor}{\mathbf{R}}
\newcommand{\encoder}{\mathcal{E}}
\newcommand{\anchors}{\mathcal{A}}
\newcommand{\anchor}{\mathbf{a}}
\newcommand{\sample}{\mathcal{S}}
\newcommand{\generator}{\mathcal{G}}
\newcommand{\latent}{\mathbf{z}}
\newcommand{\landmark}{\ell}
\newcommand{\landmarks}{\boldsymbol{\ell}}
\newcommand{\joint}{\ell^j}
\newcommand{\joints}{\boldsymbol{\ell}^j}
\newcommand{\markers}{\boldsymbol{\ell}^m}

\newcommand{\ST}[1]{\textcolor{black}{#1}}

\newcommand{\affiliationone}{\textsuperscript{1}}
\newcommand{\affiliationtwo}{\textsuperscript{2}}
\newcommand{\affiliationcombo}{\textsuperscript{1}\textsuperscript{,}\textsuperscript{2}}

\DeclareMathOperator*{\argmin}{\arg\!\min}

\usepackage{adjustbox}%
\usepackage{enumitem}
\newcommand{\fcircle}[2][red,fill=red]{\tikz[baseline=-0.5ex]\draw[#1,radius=#2] (0,0.03) circle ;}
\usepackage{pifont}
\usepackage{fontawesome5}
\usepackage{soul}
\newcommand{\xmark}{\ding{55}}%
\newcommand{\cmark}{\ding{51}}%
\usepackage[capitalize]{cleveref}
\crefname{section}{Sec.}{Secs.}
\Crefname{section}{Section}{Sections}
\Crefname{table}{Table}{Tables}
\crefname{table}{Tab.}{Tabs.}
\Crefname{figure}{Figure}{Figures}
\crefname{figure}{Fig.}{Figs.}

\usepackage[framemethod=TikZ]{mdframed}
\newcounter{prf}[section]
\newenvironment{prf}[2][]{%
\refstepcounter{prf}%
\ifstrempty{#1}%
{\mdfsetup{%
frametitle={%
\tikz[baseline=(current bounding box.east),outer sep=0pt]
\node[anchor=east,rectangle,fill=red!10]
{\strut \textit{Are high-end MoCap data necessary?}};}}
}%
{\mdfsetup{%
frametitle={%
\tikz[baseline=(current bounding box.east),outer sep=0pt]
\node[anchor=east,rectangle,fill=red!10]
{\strut Proof~\theprf:~#1};}}%
}%
\mdfsetup{innertopmargin=2pt,linecolor=red!10,%
linewidth=1pt,topline=true,%
frametitleaboveskip=\dimexpr-\ht\strutbox\relax
}
\begin{mdframed}[]\relax%
\label{#2}}{\end{mdframed}}
\newcounter{propb}[section] 
\renewcommand{\thepropb}{\arabic{propb}}
\newenvironment{propb}[2][]{%
\refstepcounter{propb}%
\ifstrempty{#1}%
{\mdfsetup{%
frametitle={%
\tikz[baseline=(current bounding box.east),outer sep=0pt]
\node[anchor=east,rectangle,fill=green!20]
{\strut \textit{Addressing the bias and long-tail}};}}
}%
{\mdfsetup{%
frametitle={%
\tikz[baseline=(current bounding box.east),outer sep=0pt]
\node[anchor=east,rectangle,fill=green!20]
{\strut Proposition~\thepropb:~#1};}}%
}%
\mdfsetup{innertopmargin=1pt,linecolor=green!20,%
linewidth=2pt,topline=true,%
frametitleaboveskip=\dimexpr-\ht\strutbox\relax
}
\begin{mdframed}[]\relax%
\label{#2}}{\end{mdframed}}
\newcounter{theob}[section] 

\newenvironment{theob}[2][]{%
\refstepcounter{theob}%
\ifstrempty{#1}%
{\mdfsetup{%
frametitle={%
\tikz[baseline=(current bounding box.east),outer sep=0pt]
\node[anchor=east,rectangle,fill=blue!20]
{\strut \textit{Direct joint solving}};}}
}%
{\mdfsetup{%
frametitle={%
\tikz[baseline=(current bounding box.east),outer sep=0pt]
\node[anchor=east,rectangle,fill=blue!20]
{\strut Theorem~\thetheo:~#1};}}%
}%
\mdfsetup{innertopmargin=1pt,linecolor=blue!20,%
linewidth=2pt,topline=true,%
frametitleaboveskip=\dimexpr-\ht\strutbox\relax
}
\begin{mdframed}[]\relax%
\label{#2}}{\end{mdframed}}

\newcounter{lem}[section] 
\renewcommand{\thelem}{\arabic{lem}}
\newenvironment{lem}[2][]{%
\refstepcounter{lem}%
\ifstrempty{#1}%
{\mdfsetup{%
frametitle={%
\tikz[baseline=(current bounding box.east),outer sep=0pt]
\node[anchor=east,rectangle,fill=violet!20]
{\strut \textit{Explicit vs implicit labeling}};}}
}%
{\mdfsetup{%
frametitle={%
\tikz[baseline=(current bounding box.east),outer sep=0pt]
\node[anchor=east,rectangle,fill=violet!20]
{\strut Corollary~\thelem:~#1};}}%
}%
\mdfsetup{innertopmargin=1pt,linecolor=violet!20,%
linewidth=2pt,topline=true,%
frametitleaboveskip=\dimexpr-\ht\strutbox\relax
}
\begin{mdframed}[]\relax%
\label{#2}}{\end{mdframed}}
\newcounter{corb}[section] 
\renewcommand{\thelem}{\arabic{corb}}
\newenvironment{corb}[2][]{%
\refstepcounter{corb}%
\ifstrempty{#1}%
{\mdfsetup{%
frametitle={%
\tikz[baseline=(current bounding box.east),outer sep=0pt]
\node[anchor=east,rectangle,fill=orange!20]
{\strut \textit{Addressing input noise}};}}
}%
{\mdfsetup{%
frametitle={%
\tikz[baseline=(current bounding box.east),outer sep=0pt]
\node[anchor=east,rectangle,fill=orange!20]
{\strut Corollary~\thelem:~#1};}}%
}%
\mdfsetup{innertopmargin=1pt,linecolor=orange!20,%
linewidth=2pt,topline=true,%
frametitleaboveskip=\dimexpr-\ht\strutbox\relax
}
\begin{mdframed}[]\relax%
\label{#2}}{\end{mdframed}}

\begin{document}

\title{Noise-in, Bias-out: Balanced and Real-time MoCap Solving}

\author{Georgios Albanis \affiliationcombo
\quad \quad \quad
Nikolaos Zioulis \affiliationone
\quad \quad \quad
Spyridon Thermos \affiliationone\\
\quad
Anargyros Chatzitofis \affiliationone
\quad \quad \quad
Kostas Kolomvatsos \affiliationtwo
\\
\\
\affiliationone  Moverse {\tt\small \{giorgos,nick,spiros,argyris\}@moverse.ai}\\
\affiliationtwo  Dept. of Informatics and Telecommunications, University of Thessaly {\tt\small kostasks@uth.gr}
}

\twocolumn[{%
\renewcommand\twocolumn[1][]{#1}%
\maketitle
\ificcvfinal\thispagestyle{empty}\fi

\begin{center}
    \centering
    \includegraphics[width=\textwidth]{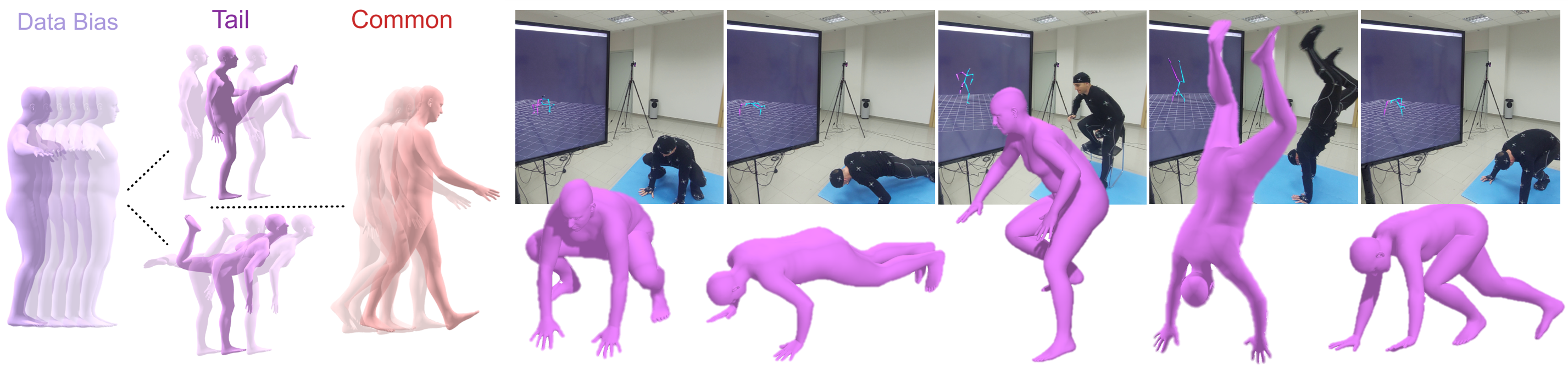}
    \captionof{figure}{
        Effective use of human motion data needs to overcome their inherent temporal bias and long-tailed distribution (\textit{left}).
        Our model uses a novel balanced regression technique to improve robustness and accuracy to challenging poses, de-noise markers and solve joints using raw unstructured marker positions as input.
        It runs in real-time and can handle higher noise levels (\textit{right}), producing high-quality body fits even when deployed in a system using just $3$ consumer-grade sensors.
    }
    \label{fig:teaser}
\end{center}
}]

\begin{abstract}
Real-time optical Motion Capture (MoCap) systems have not benefited from the advances in modern data-driven modeling.
In this work we apply machine learning to solve noisy unstructured marker estimates in real-time and deliver robust marker-based MoCap even when using sparse affordable sensors.
To achieve this we focus on a number of challenges related to model training, namely the sourcing of training data and their long-tailed distribution.
Leveraging representation learning we design a technique for imbalanced regression that requires no additional data or labels and improves the performance of our model in rare and challenging poses.
By relying on a unified representation, we show that training such a model is not bound to high-end MoCap training data acquisition, and exploit the advances in marker-less MoCap to acquire the necessary data.
Finally, we take a step towards richer and affordable MoCap by adapting a body model-based inverse kinematics solution to account for measurement and inference uncertainty, further improving performance and robustness. Project page:  \href{https://moverseai.github.io/noise-tail}{moverseai.github.io/noise-tail}.
\end{abstract}

\section{Introduction}
\label{sec:intro}
Human Motion Capture (MoCap) technology has benefited from the last decade's data-driven breakthroughs mostly due to significant research on the human-centric visual understanding that focuses on unencumbered capture using raw color inputs.
The golden standard of MoCap technology -- referred to as ``optical'' -- still uses markers attached to the body, often through suits, for robust and accurate captures, and has received little attention in the literature.
These scarce works \cite{han2018online,ghorbani2019auto,soma,mocap-solver,holden,democap} mainly focus on processing (raw) archival MoCap data for direct marker labeling \cite{ghorbani2019auto,soma} or labeling through regression \cite{han2018online}, solving the skeleton's joints \cite{mocap-solver,democap} or transforms \cite{holden}, while \cite{democap} also addressing the case of commodity sensor captures and the noise levels associated with it.

As even high-end systems produce output with varying noise levels, be it either information- (swaps, occlusions, and ghosting), or measurement-related (jitter, positional shifts), these works exploit the plain nature of raw marker representation to add synthetic noise during training.
Still, for data-driven systems, the variability of marker placements comprises another challenge that needs to be addressed.
Some works \cite{democap,holden} address this implicitly, relying on the learning process, while others \cite{mocap-solver} address this quasi-explicitly, considering them as input to the model.
Another way to overcome this involves fitting the raw data to a parametric model after manually \cite{han2018online,mosh,amass}, or automatically \cite{ghorbani2019auto,soma} labeling and/or annotating correspondences, standardizing the underlying representation.

In this work, we explore the next logical step stemming from prior work, bridging standardized representations and consumer-grade sensing, and delivering real-time data-driven MoCap that is robust to tracking errors.
Most works \cite{soma,mocap-solver,ghorbani2019auto,holden,democap} leverage high-end MoCap to acquire training data, a process that is expensive, laborious and difficult to scale, apart from \cite{han2018online} that used low-cost sensor acquired data, but nonetheless, applied the model to a high-end capturing system.

Instead, by relying on a standardized representation using a parametric human body model, we benefit from modern markerless capture technology, greatly increasing data acquisition rates at a fraction of the costs and labor.
Still, there are certain challenges that need to be addressed, such as the distribution of MoCap data and the input optical sensing noise.

The nature of human motion, albeit high-dimensional, instills a significant level of data redundancy in MoCap datasets.
Indeed, standing still or walking poses dominate most captures and affect the training data distribution in two ways.
First by introducing bias in the learning process, and second, by further skewing the long-tailed distribution.
The latter is an important problem \cite{rong2022chasing} that data-driven methods need to overcome as rare poses exist, not only due to their reduced appearance frequency, but also due to biomechanical limitations of the captured subjects in fast movements, body balancing, and striking challenging poses.
Prior work crucially neglects this, resorting to uniform temporal downsampling, which only helps in reducing data samples, yet not redundancy nor long-tailed distribution.

Another typical assumption is that the raw marker data are relatively high quality, most common to labeling works \cite{soma,ghorbani2019auto} that solve using the raw positions.
Even though synthetic noise is added during training, this is mostly to regularize training as the noisy nature of inputs is not taken into account post-labeling.
Those works that directly infer solved estimates \cite{democap,mocap-solver,holden} solely rely on the model's capacity to simultaneously denoise the inputs and solve for the joints' positions.
Nonetheless, even the models' outputs are uncertain, a situation that will be increasingly magnified when the raw marker input is affected by higher noise levels, as common when relying on consumer-grade sensors.
This lack of solutions that increase noise robustness hinders the adoption of more accessible sensing options.

To that end, we present techniques to address MoCap dataset challenges as well as noisy inputs, resulting in a MoCap framework that \fcircle[fill=red]{2pt} \underline{does not} necessarily require data from high-end MoCap systems, \fcircle[fill=red]{2pt} \underline{does not} require additional data to boost long-tail performance, and \fcircle[fill=red]{2pt} \underline{does not} require specialised hardware.
More specifically we:
\begin{itemize} [label=\ding{212}] %
    \item Leverage representation learning to jointly oversample and perform utility-based regression, addressing the redundancy and long-tailed MoCap data distribution.
    \item Introduce a noise-aware body shape and pose solver that models the measurement uncertainty region during optimization.
    \item Demonstrate a real-time inference capable and artifact-free MoCap solving model, running at $60Hz$ on a system comprising just 3 consumer-grade sensors.
    \item Harness a human parametric representation to cold-start data-driven optical MoCap models using data through markerless acquisition methods.
\end{itemize}

\section{Related Work}
\label{sec:related}
\subsection{MoCap Solving}
\label{subsec:related_mocap}

Solving the joints' positions or transforms from marker data is a cascade of numerous (sometimes optional) steps.
The markers need to be labeled, ghost markers need to be removed, occluded markers should be predicted and then an articulated body structure needs to be fit to the observed marker data.
Various works address errors at different stages of MoCap solving, with contemporary ones relying on smoothness and bone-related (angles, offsets and lengths) constraints \cite{herda2001using,ringer2004procedure,hornung2005self,aristidou2013real,feng2014mining,meyer2014online,steinbring2016real}.
Recent approaches started resorting to existing data for initialization \cite{schubert2015automatic} or marker cleaning \cite{aristidou2018self}.
MoSh \cite{mosh} moved one step ahead and instead of relying on plain structures employed a parametric human body to solve labeled marker data and estimate pose articulation and joint positions, even accounting for marker layout inconsistencies and/or soft tissue motion.

Nonetheless the advent of modern -- deep -- data-driven technologies have stimulated new approaches for MoCap solving.
A label-via-regression approach was employed in \cite{han2018online} where a deep model was used to regress marker positions and then perform maximum assignment matching for labeling the input.
Labeling was also formulated as permutation learning problem \cite{ghorbani2019auto}, albeit with constraints on the input, which were then relaxed in \cite{soma} by adding a ghost category.
However, labeling assumes that the raw data are of a certain quality as the raw measurements are then used to solve for the joints' transforms or extra processing steps are required to denoise the input.

Consequently, end-to-end data-driven approaches that can simultaneously denoise and solve have been a parallel line of research.
While end-to-end cleaning and solving is possible using solely a single feed-forward network \cite{holden}, the process naturally benefits from using two cascaded autoencoders \cite{pavllo2019real}, the first operating on marker data and cleaning them for the subsequent joint regressor.
The staging from markers to joints was also shown to be important from a performance perspective in \cite{democap} which trained a convolutional network with coupled noisy and clean data captures to address noisy inputs.
Recently, graph convolutional models were employed in \cite{mocap-solver} allowing for the explicit encoding of marker layout and skeleton hierarchy, two crucial factors of variation that were only implicitly handled in prior end-to-end solvers.

\subsection{MoCap Data}
\label{subsec:related_data}

Learning to solve MoCap marker data requires supervision provided by collecting data using professional high-end MoCap systems \cite{holden,soma,mocap-solver,democap}.
SOMA \cite{soma} standardized the representation using the AMASS dataset \cite{amass} which, in turn, relied on an extension of MoSh \cite{mosh} to fit a parametric human body model to markers.
All other works suffer from inconsistent marker layouts which is a problem that was either implicitly addressed \cite{holden,democap} or quasi-explicitly \cite{mocap-solver} using the layouts as inputs.
Marker data can be (re-)synthesized in different layouts when higher-level information is available (\textit{e.g.}~marker-to-joint offsets, meshes) \cite{holden,soma}.
Yet, it has been also shown that fitting a synthetic hand model to depth data acquired by consumer-grade sensors can also produce usable training data \cite{han2018online} for deploying a model to a high-end marker capturing system for data-driven MoCap.
Compared to \cite{han2018online}, we experimentally demonstrate this feasibility and even extend it to noisy inputs at run-time, something not considered in \cite{han2018online} as it relied on a high-end system for live capture.

Statistical parametric models \cite{loper2015smpl,pavlakos2019expressive,osman2020star,osman2022supr,wang2021panoman,xu2020ghum,alldieck2021imghum,yan2021ultrapose} are more expressive alternatives than the skinned mesh \cite{libhand} used in \cite{han2018online} as, apart from realistic shape variations, deformation corrective factors can also be employed.
They have been used to synthesize standardized training data before \cite{varol2017learning,hoffmann2019learning,kaufmann2021pose} but crucially rely on preceeding high-end MoCap acquisition.
We also explore this path using multi-view markerless capture \cite{iskakov2019learnable,cheng2022generalizable,thuman2} to produce parametric model fits and synthesize marker positions as a solution to the cold-start problem of data-driven MoCap solving.
Even though such data can be fit to marker data as done in AMASS \cite{amass} and Fit3D \cite{fit3d}, the potential of acquiring them using less expensive capture solutions is very important, as long as it is feasible to train high quality models.

Still, one also needs to take into account the nature of human performance data and their collection processes. 
As seen in AMASS \cite{amass} and Fit3D \cite{fit3d}, both contain significant redundancies and suffer from the long-tail distribution effect.
Rare poses are challenging for regression models to predict, mainly stemming from the combined effect of the selected estimators and stochastic optimization with mini-batches.
Various solutions have been surfacing in the literature, some tailored to the nature of the problem \cite{rong2022chasing}, leveraging a prototype classifier branch to initialize the learned iterative refinement, and others adapting works from imbalanced classification to the regression domain.
Traditional approaches fall into either the re-sampling or re-weighting category, with the former focusing on balancing the frequency of samples and the latter on properly adjusting the parameter optimization process.
Re-sampling strategies involve common sample under-sampling \cite{torgo2015resampling}, rare sample over-sampling by synthesizing new samples via interpolation \cite{torgo2013smote}, re-sampling after perturbing with noise \cite{branco2016ubl}, and hybrid approaches that simultaneously under- and over-sample 
\cite{branco2017smogn}.
Yet interpolating high-dimensional samples like human pose is non-trivial or even defining the rare samples that need to be re-sampled.

Utility-based \cite{torgo2007utility} -- or otherwise, cost-sensitive -- regression assigns different weights -- or relevance -- to different samples.
Defining a utility function is also essential to re-sampling strategies for regression \cite{torgo2015resampling}.
Recent approaches employ kernel density estimation \cite{steininger2021density}, adapt evaluation metrics as losses \cite{silva2022model}, or resort to label/feature smoothing and binning \cite{yang2021delving}.
Another family of methods that are now explored can be categorized as contrastive, with \cite{gong2022ranksim} regularizing training to enforce feature and output space proximity.
BalancedMSE \cite{ren2022balanced} is also a contrastive-like objective that employs intra-batch minimum error sample classification using a cross-entropy term that corresponds to an L2 error from a likelihood perspective.
However, most approaches rely on stratified binning of the output space using distance measures that lose significance in higher dimensions. Further, binning can only be used with specific networks/architectures (proper feature representations for classifying bins or feature-based
losses). It has not been shown to be applicable in high-performing dense networks relying on heatmap representations.
Instead, we introduce a novel technique that can jointly over-sample and assign higher relevance to rare samples by leveraging representation learning and its synthesis and auto-encoding traits.

\section{Approach}
\label{sec:approach}
\begin{figure*}
    \centering
    \includegraphics[width=\textwidth]{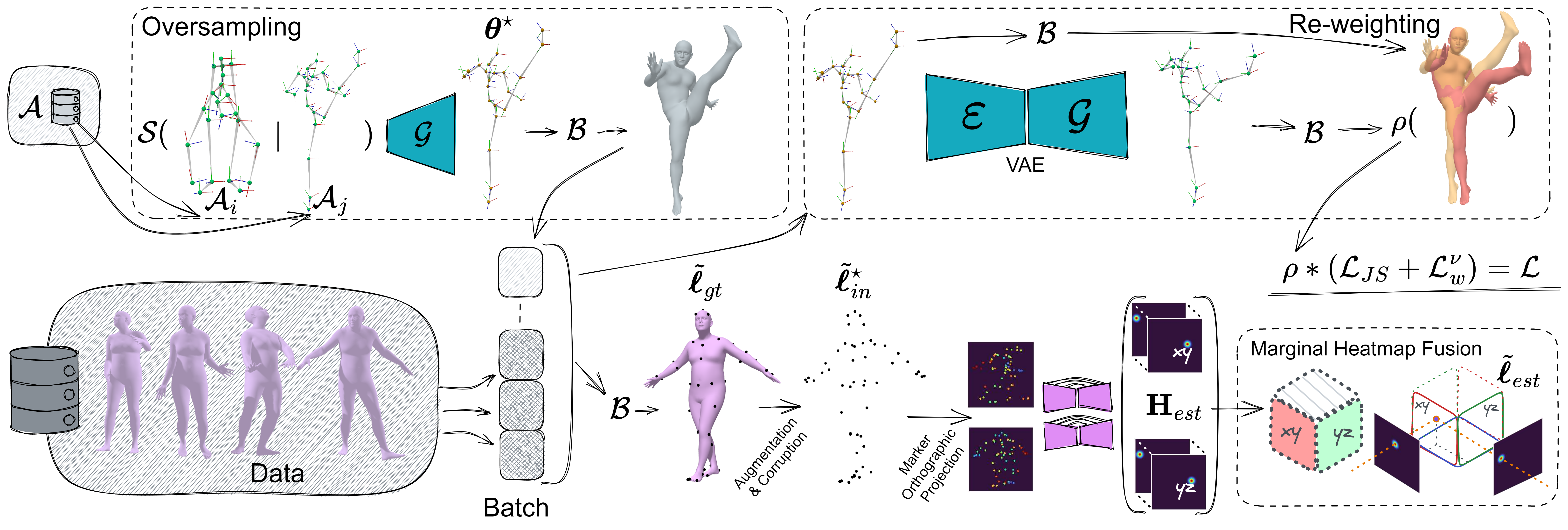}
    \caption{
    Overview of the balanced and real-time MoCap solving training model.
    Starting from an existing data corpus (\textit{bottom left}), a set of encoded tail anchor poses $\anchors$ are selected (\cref{subsec:data_distribution} - \textit{top left}) and randomly blended via $\mathcal{S}$ and a generator $\mathcal{G}$.
    This oversamples the tail, adding extra synthetic rare samples during training.
    A UNet model (\cref{subsec:model_design} - bottom middle) receives two orthographic depth map renders ($xy$ and $yz$ planes) of augmented and corrupted marker 3D positions $\landmarks_{in}^{\star}$ extracted from the body's $\mathcal{B}$ surface, producing $2$ orthogonal heatmaps which are marginally fused along the $y$ coordinate, producing 3D positions $\tilde{\landmarks}_{est}$ (\cref{subsec:model_design} - \textit{bottom right}).
    The loss for each batch item is re-weighted by its relevance $\rho$,  computed after calculating the joint reconstruction error of its pose's $\pose$ generative autoencoder reconstruction  (\cref{subsec:data_distribution} - \textit{top right}).
    }
    \label{fig:model_overview}
\end{figure*}

The MoCap representation we use is a parametric human body model $\body$.
Different variants exist, all data-driven, some relying on stochastic representations \cite{xu2020ghum}, others on explicit ones \cite{loper2015smpl,osman2020star}, with a notable exception using an artist-made one \cite{yan2021ultrapose} and all typically employ linear blend skinning \cite{jacobson2014skinning} and pose corrective factors \cite{loper2015smpl,xu2020ghum} to overcome its artifacts.
Generally, we consider it as a function $\mesh = \body(\shape, \pose, \transform)$, where $\mesh$ are the vertices $\vertices \in \mathbb{R}^{V \times 3}$ and faces $\faces \in \mathbb{N}^{F \times 3}$ of a triangular mesh surface that is defined by $S$ blendshape coefficients $\shape \in \mathbb{R}^S$, articulated by $P$ pose parameters $\pose \in \mathbb{SO}(3)^P$, and globally positioned by the transform $\transform \in \mathbb{SE}(3)$.
Using linear functions $\regress$ expressed as matrices $\regressor$ it is possible to extract $L$ different body landmarks $\landmarks := \regress(\vertices) = \regressor \times \vertices$, with $\landmarks \in \mathbb{R}^{L \times 3}$ and $\regressor \in \mathbb{R}^{L \times V}$.
This way, surface points $\landmarks^v$ can be extracted using delta (vertex picking) or barycentric (triangle interpolation) functions and joints $\joints$ using weighted average functions.
Since markers are extruded by the marker radius $d$ they correspond to $\markers = \landmarks^v + d (\regressor \times \normals)$, with $\normals$ being the vertices' normals.

Following prior art \cite{soma}, the input data are the parameters of a body model that synthesize markers, which due to their synthetic nature can be augmented, and corrupted with artifacts and noise \cite{holden,mocap-solver,democap}.
\cref{fig:model_overview} illustrates our model's training framework which is followingly explained starting with the technique addressing the redundancy and long-tailed nature of the data (\cref{subsec:data_distribution}), the marker denoising and joint solving model's design choices (\cref{subsec:model_design}), and finally the noise-aware body parameter solver (\cref{subsec:likelihood_mosh}).

\subsection{Balancing Regression}
\label{subsec:data_distribution}
Relevance functions drive utility regression and guide the re-/over-/inter-sample selection/generation \cite{branco2017smogn,torgo2015resampling,torgo2007utility,torgo2013smote}.
Instead of defining relevance or sample selection based on an explicit formula or set of rules, we employ representation learning to learn it from the data.
Autoencoding synthesis models \cite{Kingma2014,rezende2015variational} jointly learn a reconstruction model as well as a generative sampler:
\begin{equation}
\label{eq:prior_model}
    \pose^{\ddagger} = \generator(\encoder(\pose)), \qquad \pose^{\star} = \generator(\sample(\cdot)),
\end{equation}
with varying constraints on the input $\pose$ and latent $\latent = \encoder(\pose), \latent \in \mathbb{R}^Z$ spaces.
An encoder $\encoder(\pose)$ maps input $\pose$ to a latent space $\latent$ which gets reconstructed to $\pose^{\ddagger}$ by a generator $\generator(\latent)$.
Using a sampling function $\sample$ to sample the latent space it is also possible to generate novel output samples $\pose^{\star}$.
We exploit the hybrid nature of such models to design a novel imbalanced regression solution that simultaneously over-samples the distribution at the tail and adjusts the optimization by re-weighting rarer samples. Our solution is based on a deep Variational AutoEncoder (VAE)~\cite{Kingma2014}.

\begin{figure}[]
    \centering
    \includegraphics[width=\columnwidth]{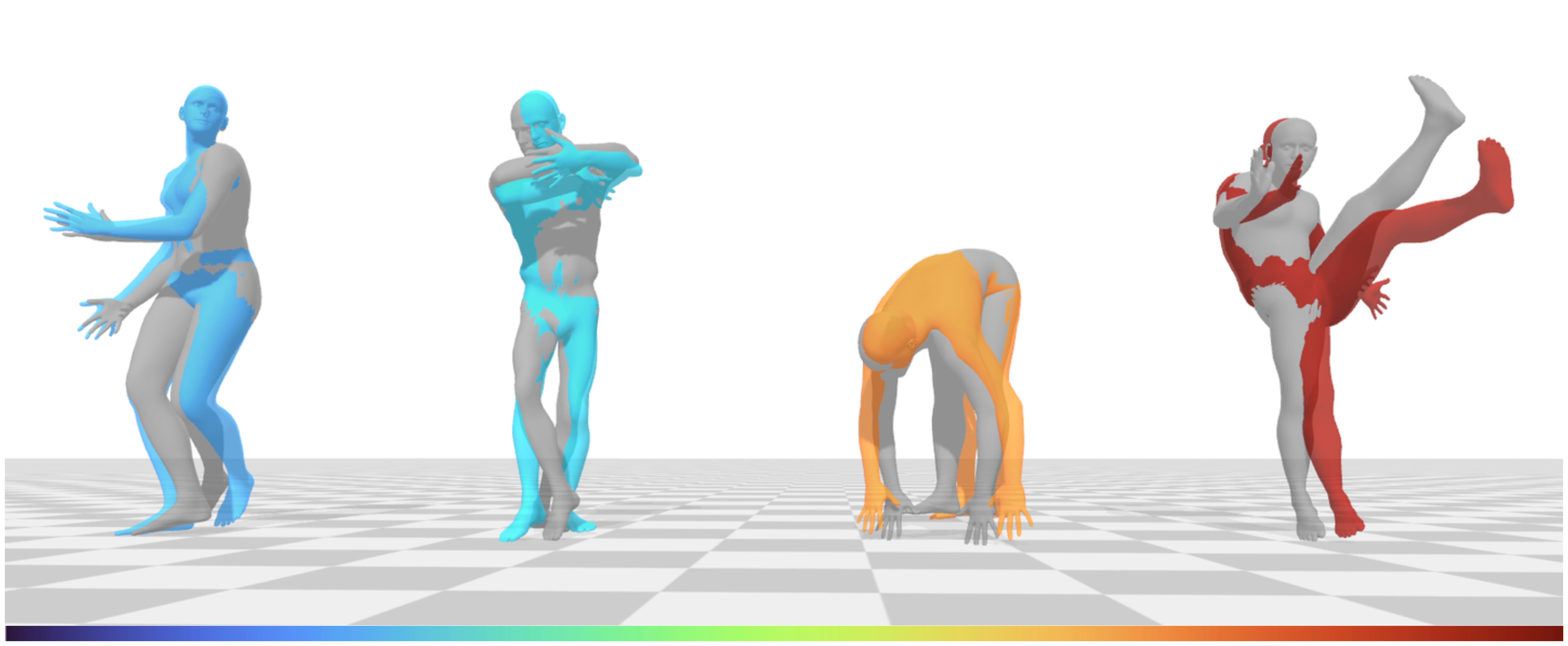}
    \caption{
    Color-coded (turbo colormap \cite{turbo} at the bottom) autoencoding relevance $\rho$ of various poses.
    }
    \label{fig:relevance_function}
\end{figure}

\textbf{Relevance via Reconstructability.}
Autoencoding models are expected to reflect the bias of their training data, with redundant/rare samples being easier/harder to properly reconstruct respectively.
This bias in reconstructability can be used to assign relevance to each sample as those more challenging to reconstruct properly are more likely to be tail samples.
We define a relevance function $\rho$ (see Fig.~\ref{fig:model_overview} re-weighting) using a reconstruction error $\epsilon$:
\begin{equation}
\label{eq:relevance}    
    \rho(\theta) = 1 + exp(\nicefrac{\epsilon}{\sigma}), \quad \epsilon = \sqrt{\frac{1}{J}\sum_{i=1}^{J} ||\bar{\joint_i} -\bar{\joint_i}^{\ddagger}||_2},
\end{equation}
with $(\bar{\cdot})$ denoting unit normalization using the input joints' bounding box diagonal, $\epsilon$ the normalized-RMSE over the reconstructed and original joints, and $\sigma$ a scaling factor controlling the relevance $\rho$.
Using landmark positions we can preserve interpretable semantics in $\rho$ and $\sigma$ as they are unidirectionally interchangeable (linear mapping) with the pose $\pose$ given fixed shape $\shape$.
Fig.~\ref{fig:relevance_function} shows exemplary poses as scored by our relevance function.

\begin{figure}
    \centering
    \includegraphics[width=\columnwidth]{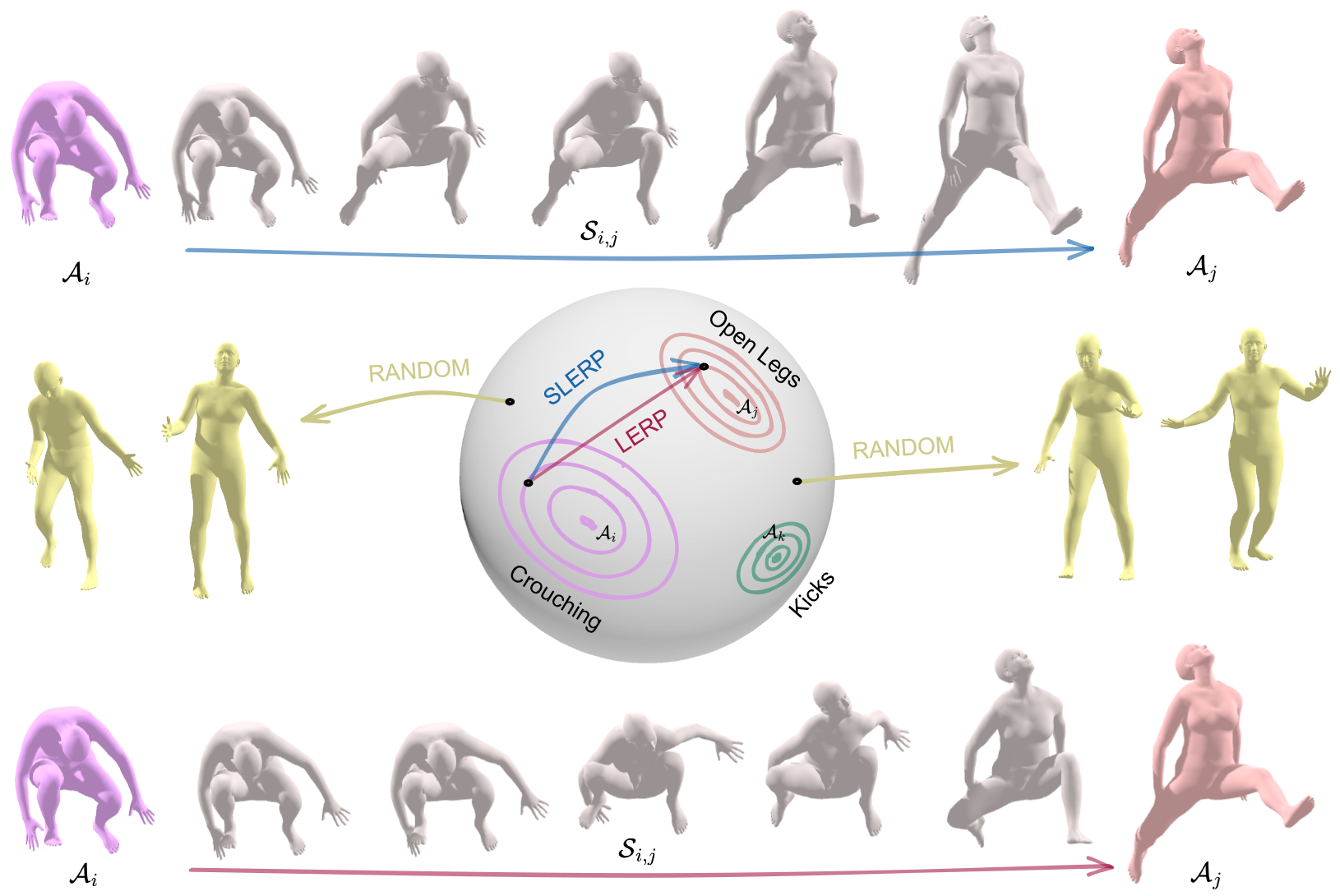}
    \caption{
    Tail oversampling using latent anchors $\anchors$.
    Random latent vector blending using \textcolor{slerp}{\textbf{non-linear}} interpolation generates diverse and realistic tail samples, compared to the \textcolor{lerp}{\textbf{linear}} one which produces less diverse or unrealistic samples, or to \textcolor{random}{\textbf{random}} sampling which produces more biased samples.
    }
    \label{fig:latent_sampling}
\end{figure}

\textbf{Balance via Controlled Synthesis.}
Even though the tail samples are not reconstructed faithfully, the generative and disentangling nature of modern synthesis models shape manifolds that map inputs to the underlying factors of data variation, effectively mapping similar poses to nearby latent codes which can be traversed across the latent space dimensions.
Based on this, we define a controlled sampling scheme for synthesizing new tail samples (see Fig.~\ref{fig:model_overview} oversampling).
Using the relevance function from \cref{eq:relevance}, it is possible to identify tail samples $\pose^{\dagger}$ via statistical thresholding that serve as anchor latent codes $\anchors = \{\,\latent^{\dagger} \mid \latent^{\dagger} = \encoder(\pose^{\dagger})\,\}$.
This process adapts to the training data distribution instead of risking a mismatch via empiric manual picking when using a purely generative model (\eg \cite{tiwari2022pose}). We then sample using the following function:
\begin{equation}
\label{eq:sampling}
    \sample_{i,j}(\cdot) = \varsigma(\mathcal{N}(\anchor_i, \mathbf{s}), \mathcal{N}(\anchor_j, \mathbf{s}), b), \quad \anchor_{i,j} \in_R \anchors.
\end{equation}
Specifically, we sample from a normal distribution centered around two random anchors $i$ and $j, i\neq j,$ from $\anchors$ using a standard deviation $\mathbf{s}$, and blend them using spherical linear interpolation \cite{shoemake1985siggraph} $\varsigma$ with a uniformly sampled blending factor $b \in \mathcal{U}(0, B), B \in [0, 1]$.
Non-linear interpolation between samples avoids dead manifold regions as not all directions lead to meaningful samples \cite{jahanian2020iclr,karras2019cvpr} and increases our samples' plausibility \cite{white2016arxiv}, as illustrated in \cref{fig:latent_sampling}.

\subsection{ Real-time Landmark Estimation}
\label{subsec:model_design}
Compared to pure labeling \cite{soma,ghorbani2019auto} or pure solving approaches \cite{holden,mocap-solver} we design our model around simultaneous denoising, solving and hallucination.

While some approaches use the raw marker positions as input, we opt to leverage the maturity of structured heatmap representations and employ a convolutional model, similar to \cite{han2018online,democap} instead of relying on unstructured regression \cite{mocap-solver,holden} using MLPs.
This improves the convergence of the model and by using multi-view fusion we can also improve accuracy via robust regression.
First, we augment and corrupt the input markers $\landmarks_{gt}$ into $\tilde{\landmarks}_{in}^{\star}$. 
Then, we normalize and render $\tilde{\landmarks}_{in}^{\star}$ from two orthographic viewpoints as in \cite{democap}, but with a notable difference when processing the model's output; instead of predicting the $3^{rd}$ dimension, we 
manage to predict normalized 3D coordinates by learning to solve a single 2D task. To achieve that, we use the two rendered views as input to the model, predict the corresponding view's heatmaps, and fuse them with a variant of marginal heatmap regression \cite{nibali20193d,ye2022faster} (see Fig.~\ref{fig:model_overview} fusion). We assume the gravity direction along the $y$ axis and use the orthogonal and orthographic views denoted as $xy$ and $yz$ which share the $y$ axis. To estimate the landmarks' normalized positions $\tilde{\landmarks}_{est}$, we employ center-of-mass regression \cite{softargmax,integral,nibali2018numerical,com} taking the average expectation \cite{nibali20193d,ye2022faster} for $y$ from the two views.
The model is supervised by:
\begin{equation}
\label{eq:model_supervision}
    \Loss = \rho(\lambda \Loss_{JS}(\Heatmap_{gt}, \Heatmap_{est}) + \Loss_w^{\nu}(\tilde{\landmarks}_{gt}, \tilde{\landmarks}_{est})),
\end{equation}
where $\Loss_{JS}$ is the $\lambda-$weighted Jensen-Shannon divergence \cite{menendez1997jensen} between the normalized ground truth and soft-max normalized predicted heatmaps, while $\Loss_w^{\nu}$ is the robust Welsch penalty function \cite{holland1977robust,dennis1978techniques}, with the support parameter $\nu$, between the normalized landmark ground-truth $\tilde{\landmarks}_{gt}$ and estimated $\tilde{\landmarks}_{est}$ coordinates.
Overall, $\Loss_{JS}$ accelerates training while $\Loss_w^{\nu}$ facilitates higher levels of sub-pixel accuracy since even though we reconstruct the heatmaps $\Heatmap$ using the normalized -- un-quantized -- coordinates \cite{zhang2020distribution}, discretization artifacts can never be removed entirely.

Note that the fusion outcome $\tilde{\landmarks}_{est}$ comprises both marker and joint estimations, essentially estimating a complete, labeled, and denoized marker set, as well as solving for the joints' positions.

Finally, we use U-Net \cite{ronneberger2015u} as a regression backbone for its runtime performance and its efficiency in high-resolution regression.

\subsection{Noise-aware Fitting}
\label{subsec:likelihood_mosh}
Given the denoised and complete set of landmarks $\tilde{\landmarks}_{est} \in \mathbb{R}^{L \times 3}$, we can fit the body to these estimates and obtain the pose $\pose$ and shape $\shape$ which implies an articulated skeleton and mesh surface.
This is a non-linear optimization problem with the standard solution being MoSh \cite{mosh} and its successor MoSh++ \cite{amass}.
However, MoSh(++) also solves for the marker layout which in our case is known apriori as the model was trained with a standard $53$ marker configuration.
Compared to prior works that assume the estimates are of high-quality or low signal-to-noise ratios, we seek to relax this assumption to support additional sensing options.
The solution to this is robust optimization but typical approaches that involve robust kernels/estimators require confident knowledge about the underlying data distribution.
This is not easily available in practice, and moreover, it varies with different sensing options but more importantly, when involving a data-driven model, it is skewed by another challenging-to-model distribution. 
The Barron loss \cite{barron2019general} is a robust variant that also adapts to the underlying distribution and interpolates/generalizes many known variants by adjusting their shape and scale jointly.

Following likelihood-based formulations \cite{kendall2018multi,hamilton2020likely} that have been presented for multi-task/robust stochastic optimization, we formulate a noise-aware fitting objective that is adaptive and optimizes the Gaussian uncertainty region $\boldsymbol{\sigma} \in \mathbb{R}^L$ jointly with the data and prior terms: 
\begin{equation}
\label{eq:optimization}
    \argmin_{\pose^{\ast}, \shape^{\ast}, \transform^{\ast}, \boldsymbol{\sigma}^{\ast}} \Error_{data} + \Error_{prior}.
\end{equation}

We use standard prior terms \cite{mosh,amass,pavlakos2019expressive} $\Error_{prior} = \lambda_{\shape} \sum ||\shape||_2 + \lambda_{\latent} \sum ||\latent||_2$, and a data term formulated as:
\begin{equation}
\label{eq:data_uncertainty}
     \Error_{data} = \sum\limits_i^L \frac{1}{2 \sigma_i^2} || \tilde{\landmark}_{est, i} - \tilde{\landmark}_i^{\ast} ||_2 + log\sigma_i.
\end{equation}

As in MoSh(++) we perform staged annealed optimization but with only 2 stages as there is no marker layout optimization.
The first stage optimizes over $\shape^{\ast},\pose^{\ast},\transform^{\ast}$, while the second stage fixes $\shape$ and $\transform$ and optimizes $\pose^{\ast},\boldsymbol{\sigma}^{\ast}$.

\section{Results}
\label{sec:results}
We base our implementation on the SMPL(-X) body model $\body$ \cite{loper2015smpl,pavlakos2019expressive}.
Our models are implemented using PyTorch \cite{paszke2019pytorch}, optimized with Adam \cite{kingma2014adam}, initialized with Kaiming init. \cite{he2015delving}, and trained for a fixed number of epochs and with a fixed seed, with the best parameters selected using the performance indicators presented in Sec.~5 of the supplement.
UNet receives $160 \times 160$ depth maps and outputs heatmaps of the same resolution for all landmarks ($53$ markers and $18$ joints in all cases apart from the experiments in \cref{tab:solving_vs_sota} where $56$ markers and $24$ joints are used for consistency).
The autoencoding generator is implemented as a robust variant of VPoser \cite{pavlakos2019expressive}\footnote{Description and comparison can be found in Sec.~7.1 of the suppl.}.
To fit the body to the estimated landmarks we use quasi-Newton optimization \cite{nocedal2006nonlinear}. For the evaluation, the $\tilde{\landmarks}_{est}$ are denormalized to $\landmarks_{est}$. Finally, the Tables are color-coded with the best result being visualized in pink and bolded, the second in green, and the third (where it is needed) in yellow.

We use a variety of datasets that provide corresponding parametric body $\body$ parameters from which we can extract input (markers) and ground truth (joints and markers).
We additionally curate a custom test set comprising $4$ categories of tail samples. 
Note that all models' performance is validated using \textit{unseen} data comprising entire datasets, thus, ensuring different capturing contexts.
For a lack of space, we moved all preprocessing (see supp. Sec.~3), datasets (see supp. Sec.~4), and metrics (see supp. Sec.~5) details in the supplement, as well as an in-the-wild \href{https://youtu.be/Eu8j8fGeO_o}{supp. video}.

\begin{table}[t]
\resizebox{\columnwidth}{!}{%
\begin{tabular}{ccccc}
              & \down{RMSE} & \up{PCK1}        & \up{PCK3}        & \up{PCK7}        \\ \hline
Optical\#1 & \first{50.4} $mm$ & \second{36.14\%} & \first{84.89\%} & \first{90.90\%} \\
 Optical\#2 & \third{89.9} $mm$               &  \first{41.11\%}  & \second{81.18\%} & \third{86.24\%}  \\
 Optical\#3 & 92.9 $mm$                       & \second{39.16\%} & 79.74\%          & 86.08\%         \\ \cline{1-2}
 Markerless       & \second{59.4} $mm$              &  21.70\%          & \third{79.96\%}  & \second{90.08\%} \\ \hline
\end{tabular}%
}
\caption{
Markerless vs optical data tested on ACCAD.
}
\label{tab:real_vs_mocap_data}
\vspace{-0.05in}
\end{table}

\begin{prf}{res:data}

Relying on an intermediate body model $\body$ representation opens up new opportunities for data acquisition.
We seek to validate the hypothesis that training an optical MoCap model does not necessarily require data acquired by high-end optical MoCap systems.
Recent multi-view datasets \cite{thuman2,cheng2022generalizable,peng2021neural} rely on markerless capturing technology to fit parametric body models to estimated keypoint observations.
We train our model (without the imbalanced regression adaptation) on the combined GeneBody \cite{cheng2022generalizable} and THuman2.0 (TH2) \cite{thuman2} multi-view marker-less data (\textit{Markerless}), and on $3$ high-end MoCap dataset combinations from AMASS \cite{amass}, specifically, EKUT \cite{mandery2015kit}, HumanEva \cite{sigal2010humaneva}, MoSh \cite{mosh}, and SOMA \cite{soma} (\textit{Optical \#1}); CNRS and HumanEva (\textit{Optical \#2}); and, solely HumanEva (\textit{Optical \#3}) to progressively reduce the diversity of the samples.
We equalize the different markerless and optical training data via temporal downsampling to a total of $9$mins of MoCap.
By evaluating these models using ACCAD \cite{ACCAD} (see \cref{tab:real_vs_mocap_data}), we observe a correlation between pose diversity and performance, and that the markerless data result in comparable performance to the high-end MoCap data. The latter indicates that it is possible to acquire data for optical MoCap without having access to any high-end system.

\end{prf}

\begin{propb}{res:tail}

To evaluate our novel imbalanced regression discussed in Sec.~\ref{subsec:data_distribution}, we design an experiment simulating a progressive data collection process by aggregating the DFaust~\cite{dfaust:CVPR:2017}, EYES~\cite{eyes}, EKUT, HumanEva, MoSh, PosePrior~\cite{akhter2015pose}, SFU~\cite{sfu}, SOMA, SSM, and Transitions parts from AMASS, captured with varying acquisition protocols and settings.
\cref{tab:ddist_mix} presents the results compared to a baseline model trained without re-weighting/oversampling, and the BMSE \cite{ren2022balanced} imbalanced regression loss, which is properly adapted to consider joint distances and not scalars.

\cref{tab:ddist_mix} (top) presents the results on TH2, a dataset of diverse static poses that also includes challenging poses (\textit{e.g.}~extreme bending, inversion, etc.), where our approach improves overall performance compared to BMSE that presents inferior results to the baseline model.
\cref{tab:ddist_mix} (bottom) presents the results on our ``tail" (rare) poses that include ``\textit{high kicks}", ``\textit{crouching}", ``\textit{crossed arms}", and ``\textit{crossed legs}".
Both imbalanced regression approaches improve the long-tail performance, with our oversampling and re-weighting method achieving the best results almost horizontally. These results highlight that our approach overcomes the known weakness of the BMSE balancing the data distribution at the expense of performance on more common poses.
Ablation experiments showcasing the orthogonality of oversampling and re-weighting can be found in the supplementary material~(Sec.~7.2, Tab.~4).

\end{propb}

\begin{table}[t]
\small
\centering
\begin{tabular}{@{}llcccc@{}}
                           &          & \down{RMSE}           & \up{PCK1}      & \up{PCK3}      & \up{PCK7}      \\ \midrule
\multirow{3}{*}{\rotatebox{90}{TH2}} & Base & \second{21.4} $mm$          & \second{28.69}\%          & \second{92.08}\%          & 98.60\%          \\
                           & \cite{ren2022balanced}     & 22.0 $mm$          & 25.51\%          & 91.90\%          & \second{98.62}\%          \\
                           & Ours     & \first{19.1} $mm$ & \first{32.38}\% & \first{93.55}\% & \first{99.11}\% \\ \midrule
\multirow{3}{*}{\rotatebox{90}{Tail}}      & Base & 35.8 $mm$         & 22.04\%          & 80.27\%          & 94.31\%          \\
                           & \cite{ren2022balanced}     & \second{32.9} $mm$          & \first{27.66}\% & \second{81.98}\% & \second{94.92}\%          \\
                           & Ours     & \first{29.3} $mm$ & \second{23.42}\%          & \first{84.70}\%          & \first{97.24}\% \\ \bottomrule
\end{tabular}
\caption{
Imbalanced regression results.
}
\label{tab:ddist_mix}
\vspace{-0.1in}
\end{table}
\begin{table}[t]
\resizebox{\columnwidth}{!}{%
\begin{tabular}{c|ccccc}
& \down{RMSE} & \down{JPE} & \up{PCK1}       & \up{PCK3}       & \up{PCK7}       \\ \hline
\cite{mocap-solver} & \second{21.1} $mm$          & \second{17.4} $mm$          & 38.11\%          & 84.70\%          & \first{99.17}\%          \\
                                     \cite{democap}      & 27.0 $mm$ & 17.5 $mm$ & \first{51.08}\% & \second{89.39}\% & 97.24\% \\
 Ours & \first{20.1} $mm$               & \first{15.9} $mm$              & \second{50.14}\% & \first{92.23}\% & \second{98.14}\%
\end{tabular}%
}
\caption{
Direct joint solving on CMU test set \cite{cmuWEB}.
}
\label{tab:solving_vs_sota}
\vspace{-0.1in}
\end{table}

\begin{theob}{res:solve}

We proceed with evaluating our model's ability to accurately estimate the skeleton joints $\landmarks^{j}$ from the input markers (\ie joint-solving). We compare our model against two SotA joint-solving approaches: a) MoCap-Solver \cite{mocap-solver} that uses graph convolutions and temporal information, and b) DeMoCap \cite{democap} that employs an HRNet \cite{wang2020deep} backbone and frontal-back fusion.
All models are trained and evaluated on the CMU \cite{cmuWEB} dataset as  in~\cite{mocap-solver}.
For MoCap-Solver we rerun the evaluation without normalizing the markers and the skeletons as this information should be unknown during testing. At the same time, we employ the joint position error (JPE) from \cite{mocap-solver} for a more fair comparison.
From the results in \cref{tab:solving_vs_sota} we observe that our model outperforms the SotA in both positional metrics (RMSE, JPE) while having the best or the second-best accuracy for different PCK.

\end{theob}

\begin{figure}[t]
    \centering
    \includegraphics[width=\columnwidth]{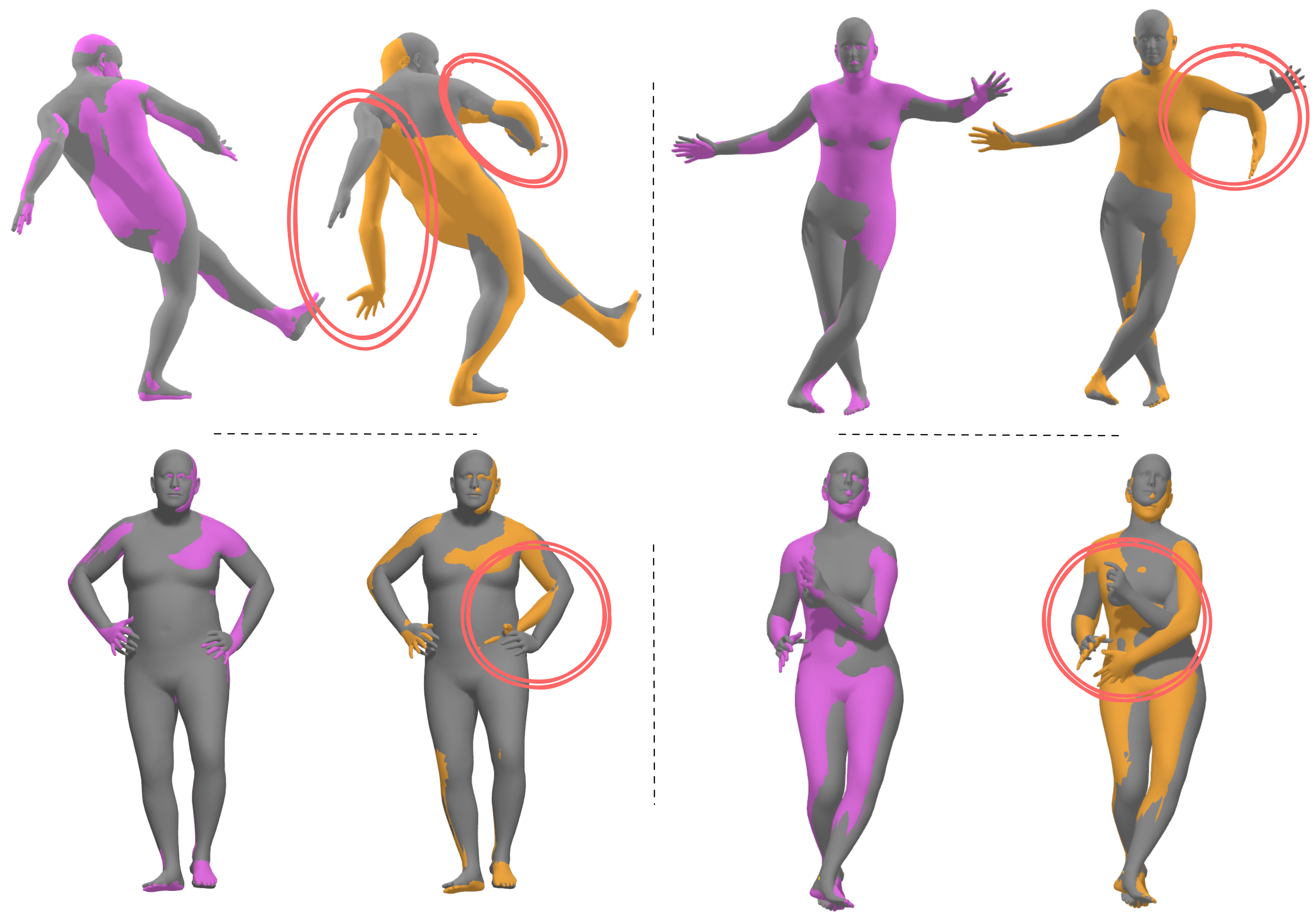}
    \caption{Fits to our \textcolor{ours}{regressed} vs SOMA \textcolor{soma}{labeled} markers. Incorrect labeling results in highly erroneous fits.}
    \label{fig:ours_vs_soma}
\end{figure}

\begin{lem}{res:label}
Our next experiment aims to showcase the advances of fitting a parametric body model on landmarks estimated with regression instead of explicitly labeling them. We compare our model that de-noises, completes, and implicitly labels landmarks via regression with SOMA, a SotA explicitly labeling method, by fitting the body to the markers similar to \cite{mosh}. Note that in order to have a fair comparison we solve \textbf{only for markers} and not for markers $\&$ joints (as discussed in Sec.~\ref{subsec:model_design}).
We train our model using the same datasets that SOMA was trained on, and then test on TH2 and our ``Tail" test set using the clean body-extracted markers, and the same MoSh-like fitting without uncertainty region optimization and without considering latent markers as the marker layout is fixed to the nominal one.
\cref{tab:labelling_vs_sota} showcases that the fits on our model's markers $\landmarks^{m}$ deliver better performance, 
a fact that is mainly attributed to the robustness of regression compared to the larger error margin of fitting to incorrectly labeled markers.
This is evident in all test sets but more pronounced in the tail (rare) poses. Indicative qualitative examples are depicted in \cref{fig:ours_vs_soma}.
For completion (not direct comparison with SOMA), we include the results for solving both markers and joints ($\landmarks$) estimated by our model, which clearly achieves the best overall performance.

\end{lem}

\begin{table}[t]
\resizebox{\columnwidth}{!}{%
\begin{tabular}{lc|ccccc}
 &                       & \down{RMSE} & \down{MAE} & \up{PCK1}       & \up{PCK3}       & \up{PCK7}        \\ \hline
\multirow{3}{*}{\rotatebox{90}{TH2}}           & \cite{soma} & 29.7 $mm$          & 3.49$^{\circ}$         & 28.33\%          & 87.78\%          & 96.11\%          \\
                                  & Ours $(\boldsymbol{\mathcal{\ell}}^{m, \ast})$        & \second{19.1} $mm$ & \first{2.68}$^{\circ}$ & \second{26.49}\% & \second{93.72}\% & \second{99.26}\% \\
 & Ours $(\boldsymbol{\mathcal{\ell}})$ & \first{17.6} $mm$                   & -                   & \first{33.92}\% & \first{95.13}\% & \first{99.35}\%  \\ \hline

\multirow{3}{*}{\rotatebox{90}{Tail}} & \cite{soma} & 68.6 $mm$          & 6.76$^{\circ}$          & 11.78\%          & 60.87\%          & 84.84\%          \\
                                  & Ours $(\boldsymbol{\mathcal{\ell}}^{m, \ast})$        & \second{30.1} $mm$ & \first{2.89}$^{\circ}$ & \second{12.11}\% & \second{73.13}\% & \first{96.87}\%  \\
 & Ours $(\boldsymbol{\mathcal{\ell}})$ & \first{28.3} $mm$                   & -                   & \first{27.31}\% & \first{83.12}\% & \second{95.35}\%
\end{tabular}%
}
\caption{
Explicit (SOMA \cite{soma}) vs implicit (Ours) labeled marker fits and direct landmarks' $\landmarks$ solving comparison. 
}
\label{tab:labelling_vs_sota}
\end{table}
\begin{table}[t]
\resizebox{\columnwidth}{!}{%
\begin{tabular}{c|ccccc}
\hline
                  & \down{RMSE}        & \down{MAE}              & \up{PCK1}        & \up{PCK3}        & \up{PCK7}        \\ \hline
\cite{mosh,amass} & \second{30.1} $mm$ & 3.49$^{\circ}$          & 11.79\%          & 66.85\%          & \first{98.34}\%  \\
\cite{barron2019general}     & 30.8 $mm$          & \second{3.10}$^{\circ}$ & \second{12.71}\% & \second{67.06}\% & 97.71\%          \\
Ours $(\landmarks^{m})$ & \first{28.9} $mm$  & \first{2.98}$^{\circ}$  & \first{14.71}\%  & \first{69.86}\%  & \second{98.18}\% \\ \hline
\end{tabular}%
}
\caption{
Noisy landmark fits comparison on TH2.
}
\label{tab:fit_results}
\end{table}

\begin{figure}
    \centering
    \includegraphics[width=\columnwidth]{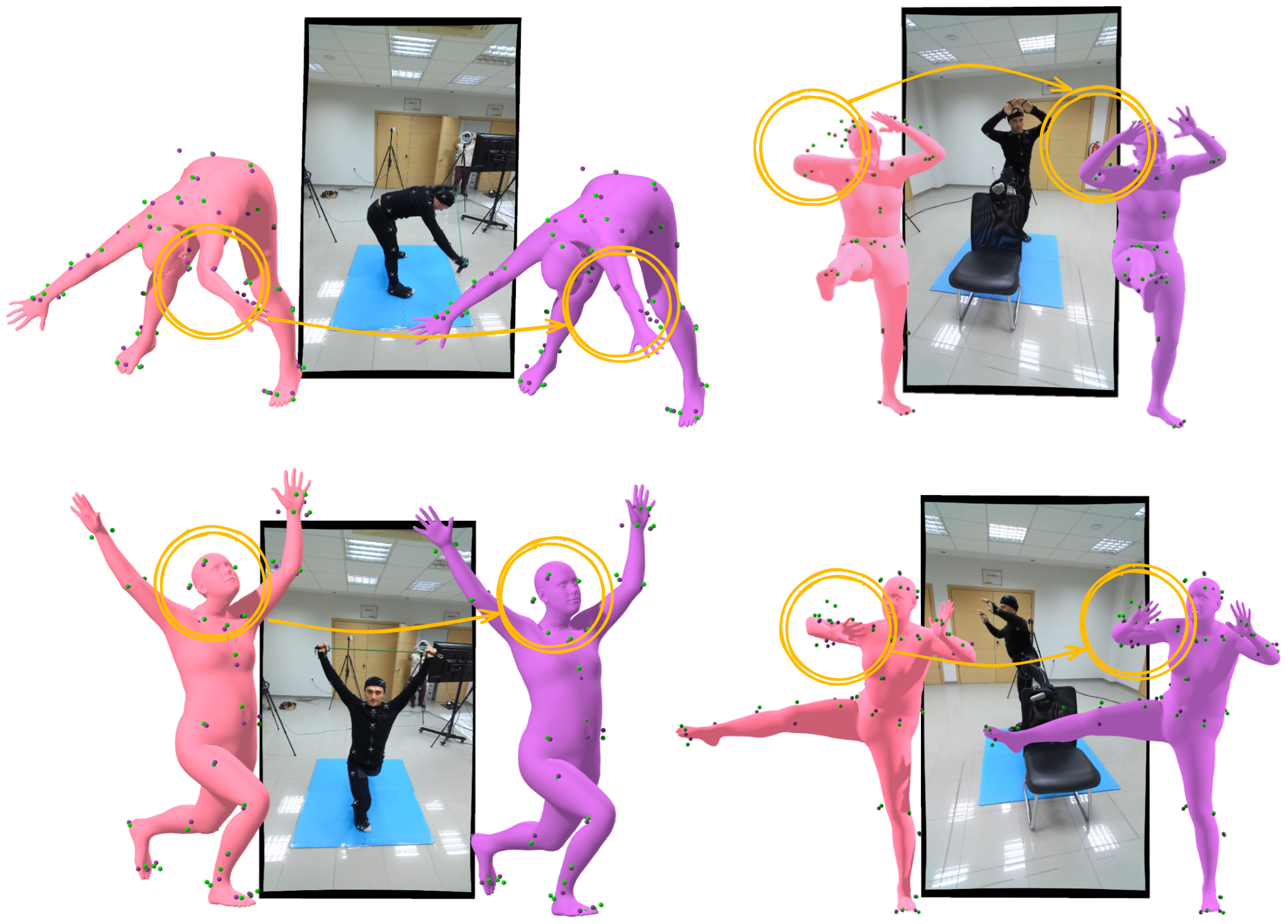}
    \caption{\textcolor{plain}{Plain} vs \textcolor{ours}{uncertainty}-based fit. Input markers from the consumer-grade system and the model inferred ones are colored with 
    with \textcolor{green}{green}, and \textcolor{violet}{violet} respectively.}
    \label{fig:naive_vs_uncertain}
\end{figure}

\begin{corb}{res:fit}

Finally, we design an experiment for showcasing our model's fitting robustness to noisy marker input as discussed in Sec.~\ref{subsec:likelihood_mosh}.
\cref{tab:fit_results} presents results when fitting to noisy landmarks between the uncertainty optimization method and  MoSh(++) like fitting (ignoring the latent marker optimization as the markers are extracted from the body's surface and placed using the nominal layout).
The TH2 dataset is used for evaluation, with the body extracted input markers corrupted with high levels of noise (see Sec.~3.2 of the supp. for the applied types of noise) prior to fitting the body model to them.
Naturally, optimizing the uncertainty region improves fitting performance to noisy observations.
Compared to a more complex optimization objective that also considers the shape of the data distribution \cite{barron2019general} we find that the proposed Gaussian uncertainty region optimization delivers improved fits.
This can be attributed to the complexity of tuning it, as well as the increased parameter count.
\cref{fig:naive_vs_uncertain} depicts qualitative examples with body fits in the noisy inputs acquired with just $3$ viewpoints (same capture session as \cref{fig:teaser}) and shows that jointly optimizing the uncertainty region allows for robustness to input-related measurement noise, as well as model-related information noise. Some interesting noise-aware fitting ablations along with visualizations can be found in Sec.~9 of the supplementary material.

\end{corb}

\noindent\textbf{Real-time performance.} We validate our end-to-end method by implementing a real-time system using sparse consumer-grade sensors (see details in Sec.~11 of the supp.).
Leveraging the orthogonal view two-pass approach we deploy an optimized ONNX \cite{onnx} model where we flatten the two passes across the batch dimension, performing only the light-weight marginal heatmap fusion in a synchronized manner.
Our system achieves under $16$ms inference even on a laptop equipped with a mobile-grade RTX 2080. Nonetheless, we understand that high-quality MoCap requires greater efficiency to achieve processing rates of at least 120Hz and we set this rate as the next goal.

\section{Conclusion}
\label{sec:conclusion}
MoCap data are highly imbalanced and in this work we have presented a novel technique for imbalanced regression.
Still we believe we have but scratched the surface of exploiting representation learning for addressing the long-tail and bias, as different architectures, samplers and relevance functions can be explored.
At the same time, this work contributes to integrating machine learning in real-time optical MoCap, while also making it more accessible.
However, there is room for improvements, as temporal information is not integrated in our approach, and a single, fixed marker layout is only supported.

\newpage
\balance
{\small
\bibliographystyle{ieee_fullname}
\bibliography{refs}
}

\appendix

\section{Intro}
\label{sec:intro}
In this supplementary material we provide additional quantitative and qualitative results to accompany the main paper.
In addition, a set of ablation studies are presented to offer extra insights into the inner workings of the methods and techniques presented in the main paper.
Finally, due to the lack of space in the main paper, we provide more details with respect to the implementation of the proposed models, the experimental protocol with respect to the datasets and metrics that were used, visualizations of related data points, and details regarding the experiments comparing to the state-of-the-art.
It should be noted that no additional training or optimization was performed in any of these experiments with respect to that presented in the main paper. 

\textit{Along with this supplementary material, we share a} \href{https://youtu.be/Eu8j8fGeO_o}{short video} \textit{that showcases the real-time performance of our MoCap system in a challenging input context, captured with only 3 Microsoft Kinect for Azure sensors.}

\cref{sec/sup:model} provides the implementation details of the UNet model used to predict landmarks $\landmarks_{est}$.
\cref{sec/sup:augmentations} clarifies the augmentation and corruptions used when training and when experimenting with noisy fits.
\cref{sec/sup:datasets} presents the different datasets that were used for the main paper's experiments and the accompanying experiments found in this supplementary material.
\cref{sec/sup:metrics} defines the metrics used to evaluate performance and the performance indicators used to select the best performing models.

Following the experimental results structure of the main paper, the remaining sections supplement the already presented analysis with additional experiments, results and insights.
\cref{sec/sup:fit} provides visualizations comparing the distribution of the markerless and marker-based data used to assess the efficacy of the former as a training corpus.
Complementary experiments are also presented to support the main paper claims.
\cref{sec/sup:balance} provides further analysis with respect to the inner workings of the balanced regression approach presented in the main paper, specifically, the VAE model's details (\cref{subsec/sup:vae}), a relevance function ablation (\cref{subsec/sup:relevance}), an investigation of the orthogonality between the different techniques (\cref{subsec/sup:orthogonality}), and an ablation of the different sampling components (\cref{subsec/sup:synthesis}).
\cref{sec/sup:solving} presents an extra experiment supplementing the solving comparison experiment conducted in the main paper.
\cref{sec/sup:landmarks} offers extra insights with respect to the landmarks regressed by our model, by ablating the fitting process across various noise levels and input landmark types.
Finally, \cref{sec/sup:qualitative} includes additional qualitative results, while \cref{sec/sup:system} describes the implementation details related to the real-time MoCap system used to capture and provide in-the-wild results.

\section{MoCap Solving Model}
\label{sec/sup:model}
Our proposed model is designed to work with any method capable of inferring markers and joints from an input markers' point cloud. 
However, for the presented study, we utilized a light-weight convolutional model that can preserve high resolution outputs, exploiting the quasi-autoencoding nature of regressing pre-defined markers (and, when applicable, joints) from unstructured marker position inputs.

Specifically, a modified version of the UNet \cite{ronneberger2015u} architecture was used to simultaneously predict $53$ markers and $18$ joints landmarks. 
It should be noted that since MoCap-Solver \cite{mocap-solver} was trained with $56$ markers and $24$ joints on the CMU data, for the experiment comparing direct solving performance, our model was adapted to the same outputs.
The model consists of $5$ convolutional blocks, with each block consisting of $32$, $64$, $128$, $256$, and $512$ features, respectively.
Each encoder block comprises $2$ convolution layers, with a kernel size of $3$, a stride and padding of $1$, followed by ReLU activations and batch normalization \cite{ioffe2015batch}.
When downscaling anti-aliased max pooling \cite{zhang2019shiftinvar} is used, while upscaling uses bilinear interpolation. 
The bottleneck of the model consists of a single convolution block, utilizing the same parameters as the encoder blocks. 
The decoder includes the same convolution blocks, and the output of each block is concatenated with the corresponding encoder's output.
Finally, the prediction layer consists of a convolution block with a kernel size of $1$, a stride of $1$, and padding of $0$, activated by the ReLU function.
Training runs for $30$ epochs with a batch size of $16$, a learning rate of $2 \times 10^{-4}$ accompanied by a step-wise schedule reducing it to $95\%$ every $4$ epochs.

As mentioned in the main paper the model is supervised by the following loss summed over all landmarks (batch notation is omitted for brevity):

\begin{equation}
\label{eq:model_supervision}
    \Loss = \sum_{i = 1}^{L}(\lambda_{JS} \Loss_{JS}(\Heatmap_{gt}, \Heatmap_{est}) + \lambda_{w} \Loss_w^{\nu}(\tilde{\landmarks}_{gt}, \tilde{\landmarks}_{est})).
\end{equation}

\noindent $\Loss_{JS}$ is the Jensen-Shannon divergence defined in \cref{eq:JS}:

\begin{equation}
\label{eq:JS}
\Loss_{JS}(\Heatmap_{gt},\Heatmap_{est}) = \frac{1}{2} D_{KL}(\Heatmap_{gt},M) + \frac{1}{2} D_{KL}(\Heatmap_{est}, M),
\end{equation}
where $D_{KL}$ is the Kullback-Leibler divergence, $M = \frac{1}{2}(\Heatmap_{gt}+\Heatmap_{est})$ is the average of $\Heatmap_{gt}$ and $\Heatmap_{est}$.

$\Loss_w^{\nu}$ is the robust Welsch penalty function, applied to the normalized $\landmarks$ coordinates,  defined by \cref{eq:welsch}, with $\nu > 0$ being a user-specified parameter set to $0.05$:

\begin{equation}
\label{eq:welsch}
\Loss_w^{\nu}(\tilde{\landmarks}_{gt}, \tilde{\landmarks}_{est}) = 1 - \exp{(-\frac{|\tilde{\landmarks}_{gt} - \tilde{\landmarks}_{est}|^2}{2v^2})}
\end{equation}

\section{Pre-processing}
\label{sec/sup:augmentations}
We use a pre-processing pipeline to augment and then corrupt the input training data.
Augmentations exploit the parametric nature of the data to increase their variance.
Similar to \cite{holden,soma,mocap-solver}, corruption exploits the simple and synthetic nature of motion capture (MoCap) to closely approximate real-world MoCap settings with noisy inputs and marker-/viewpoint- related artifacts like ghost markers, occluded markers, and varying levels of measurement noise.

\subsection{Augmentations}
First, we perform an augmentation to account for subject body shape variations.
A two-step process is employed that starts with a controlled shifting of the shape coefficients, with random values $u$ sampled from a uniform distribution $u \sim \mathcal{U}(-1,1)$:
\begin{equation}
\label{eq.betas_shifting}
\beta' = \beta + u
\end{equation}
Then, a small random subset of the shape coefficients are randomly sampled from a normal distribution:
\begin{equation}
\label{eq.random_betas}
\beta'_i = \begin{cases} \beta_i, & \text{if } i \notin S \\ \mathcal{N}(0, 1), & \text{if } i \in S \end{cases}
\end{equation}
where $S$ is a set of $n'$ indices sampled uniformly from the set of indices, with our experiments randomly shifting between $[0, 2]$ coefficients.

Then, using the rotation symmetry of the body, we randomly perform a handedness flipping augmentation by flipping the parameters of the left/right arms/legs. 

\begin{figure*}[!ht]
    \centering
    \includegraphics[width=\textwidth]{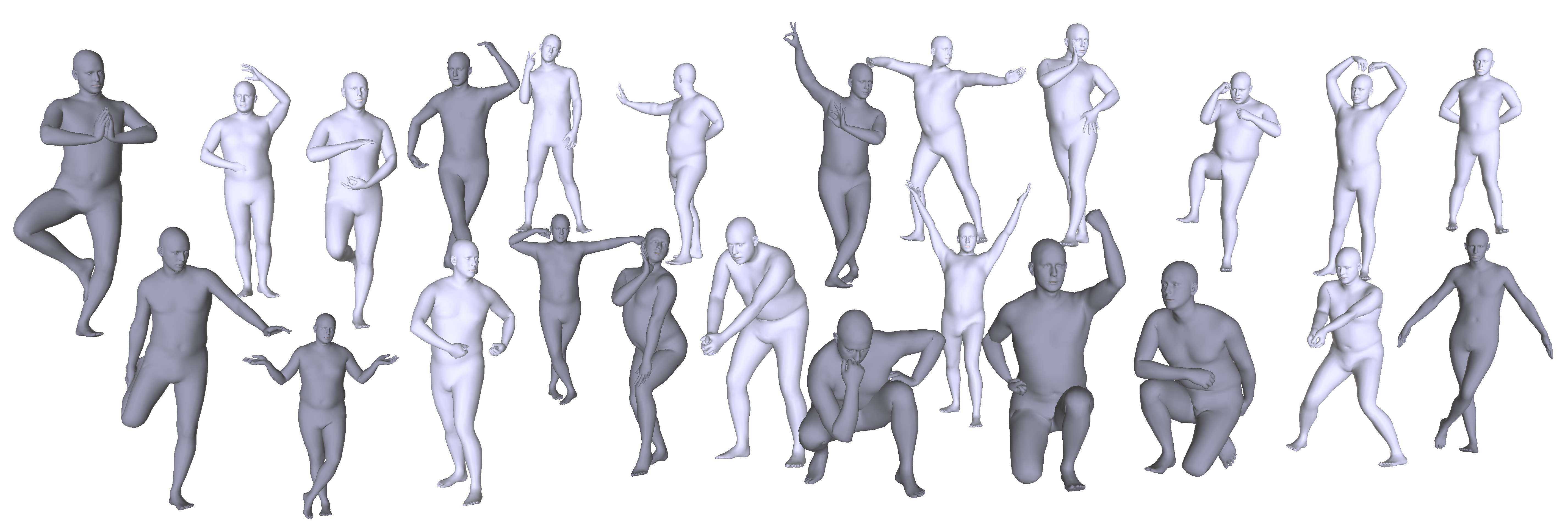}
    \caption{
        A set of random samples from the THuman2.0 \cite{thuman2} dataset. 
        The darker meshes indicate more challenging poses.
    }
    \label{fig:thuman2}
\end{figure*}

\subsection{Corruption}
We simulate marker occlusions with the following process. 
Let $\mathbf{p} = (p_1, p_2, \ldots, p_n)$ be the vector of marker positions, where $p_i$ is the position of the $i$-th marker. 
We randomly select a subset of markers for occlusion by determining the number of markers to be occluded, denoted as $k$. 
We draw a random sample from a discrete uniform distribution to determine $k$, $k \sim \mathcal{U}(m, n'), \qquad m \leq k \leq n'\leq n$, where $\mathcal{U}(m, n')$ is the uniform distribution over the range of integers $\{m_{1}, m_{2}, \ldots, n'\}$, and $n'$ defines the maximum number of markers to be occluded. 
Next, we draw another random sample from a uniform distribution to determine the indices of the markers to be occluded, \textit{i.e.}~$\mathbf{m} = (m_1, m_2, \ldots, m_k) \sim \mathcal{U}(1, n),\qquad k \leq n$ where $\mathcal{U}(1, n)$ is the uniform distribution over the markers' set of indices. 
The resulting vector ${m}$ contains the indices of the markers to be occluded and is used to exclude these markers from $\mathbf{p}$. 

As a next step, the ghosting of markers is emulated by extracting samples from a Gaussian distribution with mean and standard deviation values equivalent to the original marker positions, following \cite{soma}. 
In more detail, we first compute the median position for each spatial dimension of the marker positions, $\mu_j$, (\textit{i.e.}~the median value for the $j$-th spatial dimension of the marker positions), and the sample covariance matrix $\Sigma$. 
We then draw samples $g \sim \mathcal{G}(\mu, \Sigma)$, which are appended to the original markers' positions $\mathbf{p}$.

Finally, to simulate marker noise, we randomly select a set of markers to shift and generate a random offset for each selected marker. 
Particularly, with $N$ being the number of markers to shift, and $M$ being the maximum allowable shift distance, we randomly sample from a uniform distribution to determine the indices of the markers to which the noise will be added $I \sim \mathcal{U}(1, N)$. 
For each index $i_j \in I$, we generate a random offset vector $o \sim \mathcal{U}(-M, M)$, and add this offset to the original marker position to obtain the noisy position $\mathbf{p}' = \mathbf{p} + \mathbf{o}$.

The proposed prepossessing pipeline is randomly applied in each epoch, with specific probabilities assigned to each of the augmentation and corruption functions. In more detail, we apply the aforementioned augmentation functions with $0.5$ probability each, meaning that they will be applied to half of the instances of input data. Similarly, we apply the ghosting and occlusion corruption functions with $0.7$ probability, while the shifting one with $0.8$.

\section{Datasets}
\label{sec/sup:datasets}
\subsection{Marker-based}
For our experiments we used a variety of MoCap datasets unified within AMASS \cite{amass} to body model parameters.
The datasets we use for our experiments include the CMU dataset, which is one of the largest motion capture datasets containing a wide variety of motion types, such as walking, running, dancing, and more.
We also use the Transitions dataset, which focuses on the transitions between different activities, such as sitting down and standing up, or picking up and carrying an object.
 Additionally, we use the PosePrior dataset developed by \cite{akhter2015pose} to train a statistical model of human pose, the HumanEva dataset \cite{sigal2010humaneva}, which includes various activities performed by multiple subjects, and the ACCAD dataset \cite{ACCAD}, consisting of more action motion types such as dancing, martial arts, and sports. 
 Moreover, we use the TotalCapture dataset \cite{joo2018total}, which includes data from $5$ different subjects performing $37$ motion actions, the DFaust dataset \cite{dfaust:CVPR:2017} that includes motion data from $10$ subjects performing $129$ different types of motion, and the CNRS dataset consisting of data from $2$ subjects performing $79$ different motions.

\subsection{Markerless}
Apart from these, which were all acquired with high-end marker-based optical MoCap systems, we additionally use a number of datasets that were collected with markerless methods, using body models and fitting them to observations.
These include the THuman 2.0 \cite{thuman2} dataset, including $5$ subjects in extreme poses, the GeneBody dataset \cite{cheng2022generalizable} consisting of $50$ subjects performing various short duration activities, and the ZJU-MoCap dataset \cite{peng2021neural} that includes data from $10$ sequences of human performances. \cref{fig:thuman2} depicts an indicative subset from the THuman 2.0 dataset, which consists of both common and challenging-to-understand poses (shown with darker meshes).

\begin{figure}[]
    \centering
    \includegraphics[width=\columnwidth]{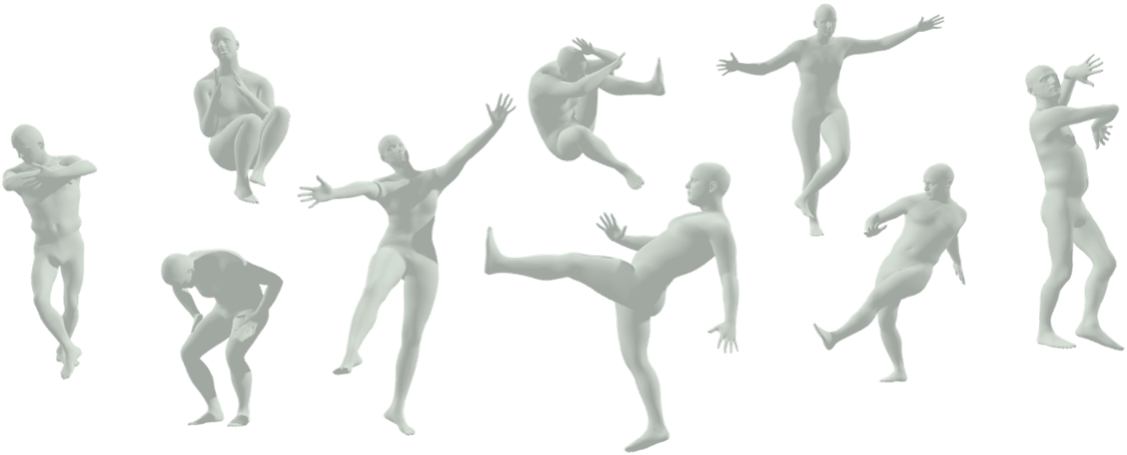}
    \caption{
        Exemplar rare and complex poses from our custom tail dataset.
    }
    \label{fig:tail}
\end{figure}
\subsection{Long-Tail}
We have manually curated a small test set comprising 274 challenging poses, including extreme and rare ones, and was used as our ``Tail" dataset for assessing long-tail regression performance.
These were coarsely grouped into 4 categories, ``\textit{crossed legs}", ``\textit{crossed arms}", ``\textit{kicks}" and ``\textit{crouching}". Indicative examples are shown in \cref{fig:tail}.

\subsection{Qualitative Distribution}
An overview of these datasets in terms of some qualitative variance indicators is presented in \cref{tab:datasets_overview}.
These were used to select by approximately equalizing the datasets used in the markerless vs optical data study.

\begin{table}[]
\centering
\begin{tabular}{lccc}
\hline
                                  & Subjects & Activities & Minutes \\ \hline
\multicolumn{1}{l|}{ACCAD}        & 20       & 14         & 26.74   \\
\multicolumn{1}{l|}{CMU}          & 111       & 25         & 543.49  \\
\multicolumn{1}{l|}{CNRS}         & 2        & 2          & 9.91    \\
\multicolumn{1}{l|}{DFaust}       & 10       & 12         & 5.72    \\
\multicolumn{1}{l|}{HumanEva}     & 3        & 5          & 8.47    \\
\multicolumn{1}{l|}{PosePrior}    & 3        & 10         & 20.82   \\
\multicolumn{1}{l|}{TotalCapture} & 5        & 12         & 41.10   \\
\multicolumn{1}{l|}{Transitions}  & 1        & 4          & 15.10   \\ \hline
\multicolumn{1}{l|}{THuman 2.0}   & 10       & -          & -       \\
\multicolumn{1}{l|}{Genebody}     & 50       & 50         & 8.33    \\
\multicolumn{1}{l|}{ZJUMoCap}     & 24       & 10         & 14.40   \\ \hline
\end{tabular}%
\caption{Datasets overview.}
\label{tab:datasets_overview}
\end{table}

\section{Performance Metrics \& Indicators}
\label{sec/sup:metrics}
\subsection{MoCap Metrics}
For evaluating our model's performance we resort to common metrics used in previous works as the  root mean squared error (RMSE), defined below:

\begin{equation}
\label{eq:rmse}
RMSE = \frac{1}{N}\sum_{i=1}^{N}\sqrt{\frac{1}{J}\sum_{j=1}^{J} ||\landmarks_{gt}^{(i,j)} -\landmarks_{est}^{(i,j)}||_2},
\end{equation}
with $N$ being the number of samples in the dataset, and
$J$ is the number of joints in each sample.
We follow the same notation for all the equations below.

Apart from RMSE, we use a PCK-like metric (\textit{i.e.}~distance accuracy metric), which measures the percentage of predicted keypoints that fall within a certain distance threshold $\tau$ from their ground-truth positions:

\begin{equation}
PCK = \frac{1}{N} \sum_{i=1}^{N} \frac{1}{J} \sum_{j=1}^{J} [ ||\landmarks_{gt}^{(i,j)} -\landmarks_{est}^{(i,j)}||_2 < \tau].
\end{equation}

In our experiments, we used three variants of PCK, namely PCK1, PCK3 and PCK7 with $\tau$ set to $10 mm$, $30 mm$, and $70 mm$ accordingly.

Finally, we use an angular metric defined in \cref{eq:angular_error}:

\begin{equation}
\label{eq:angular_error}
MAE = \frac{1}{N}\sum_{i=1}^N \frac{1}{J} \sum_{j=1}^{J} d(R_{gt}^{(i,j)}, R_{est}^{(i,j)}),
\end{equation}
where $d$ is the geodesic distance between each joint's rotation matrix $R_{gt}^{i}$ and $R_{est}^{i}$.

\subsection{Synthesis Metrics}
Inspired by C. Guo~\etal~\cite{guo2020mm}, we use two metrics to choose our best model for tail-pose generation and regression regularization, measuring quality and evaluating diversity. Regarding quality, we extract features from $1052$ generated and real samples and compute the Fr\'{e}chet Inception Distance (FID) between the feature distribution of the generated pose and poses from the THuman 2.0 test set that serve as the ``real" poses. 
To evaluate the diverse generation capability of our generative model, we generate and re-encode $1052$ samples which are then split into two subsets of the same size $N=526$. 
The diversity (DIV) is defined as the Euclidean norm of the distance between these two subsets as follows:

\begin{equation}
  \label{eq.div}
    DIV = \frac{1}{N}\sum_{i=1}^{N} \left\| v_{i} - \Tilde{v}_{i}\right\|,
\end{equation}

\noindent where $v$ and $\Tilde{v}$ correspond to re-encoded samples as vectors from a different subset.

\subsection{Performance Indicators}
The plethora of metrics makes it harder to find the best-performing model. 
To that end, we introduce a set of performance indicators, which essentially combines an error and an accuracy metric. 
Specifically, for the MoCap metrics we introduce $rmse3$ indicator, defined in \cref{eq.rmse3}:

\begin{equation}
  \label{eq.rmse3}
    rmse3 = (1 - PCK3) \times RMSE,
\end{equation}

Regarding the generative model performance, we choose our best-performing model using the indicator defined as:

\begin{equation}
  \label{eq.fid_x_div}
    synthesis = \frac{FID}{DIV}.
\end{equation}

\section{Training Data Sourcing}
\label{sec/sup:fit}
\cref{tab:real_vs_mocap_data} presents a more extensive set of experiments for the markerless vs marker-based training data study where the models are also evaluated on our ``Tail" test set.
Extra experiments are also included, namely another variant of the markerless model that was additionally trained with the ZJU-MoCap data apart from GeneBody and THuman2.0 (i.e. Markerless\#2), and another variant of the optical data, Optical\#4 trained only on the CNRS dataset.

As in the main paper, we observe that even though the best performance is offered by an optical MoCap dataset combination, the markerless alternative is close in performance and surpasses some marker-based dataset combinations.
Essentially, the quality of the data acquisition method does not seem to play a big part in the performance of the model, but instead the variance of the samples seems to be the largest performance denominator.

\begin{figure*}[!htp]

\begin{subfigure}{0.22\textwidth}
    \includegraphics[width=\textwidth]{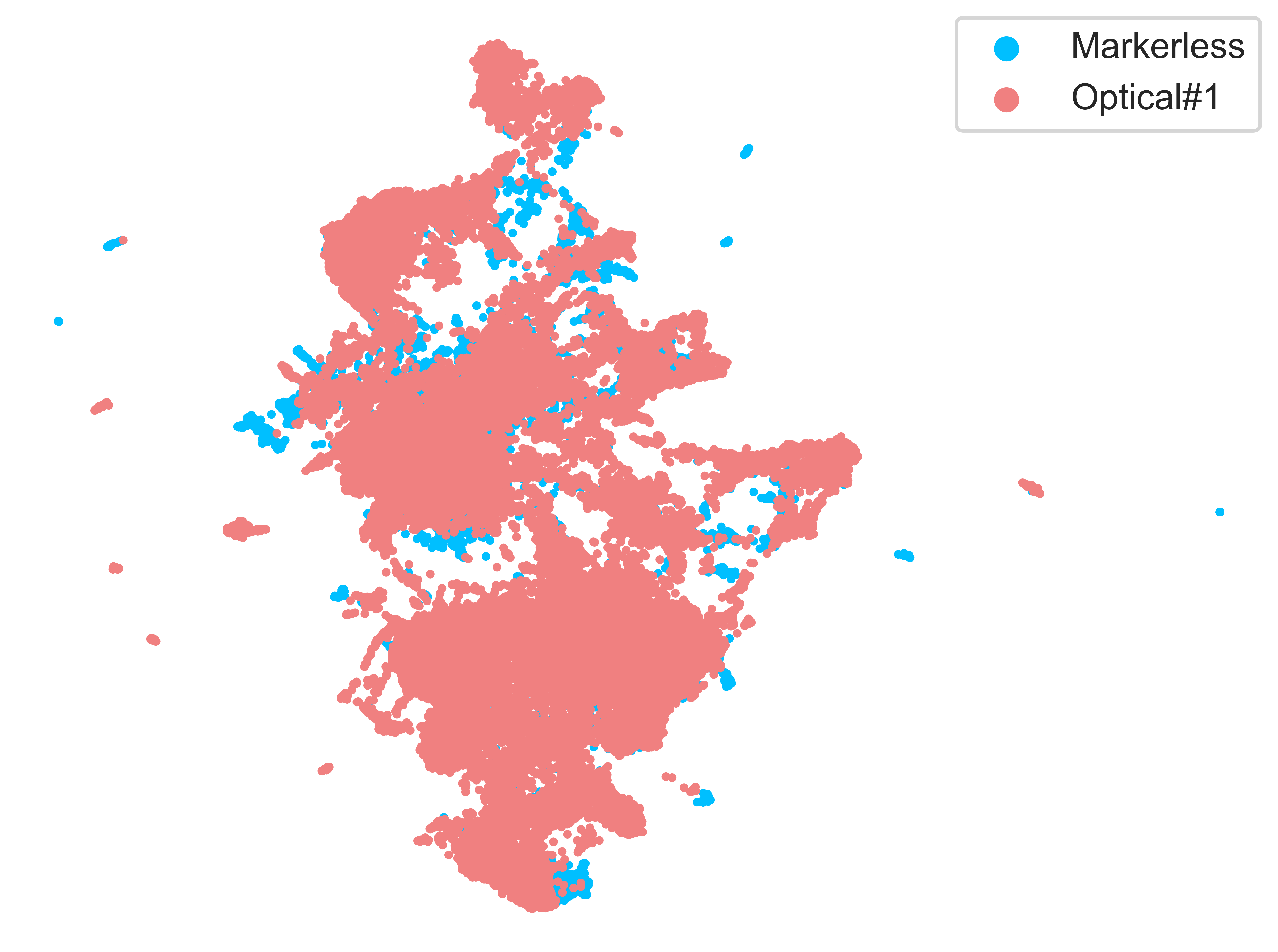}
\end{subfigure}\hfill
\begin{subfigure}{0.22\textwidth}
    \includegraphics[width=\textwidth]{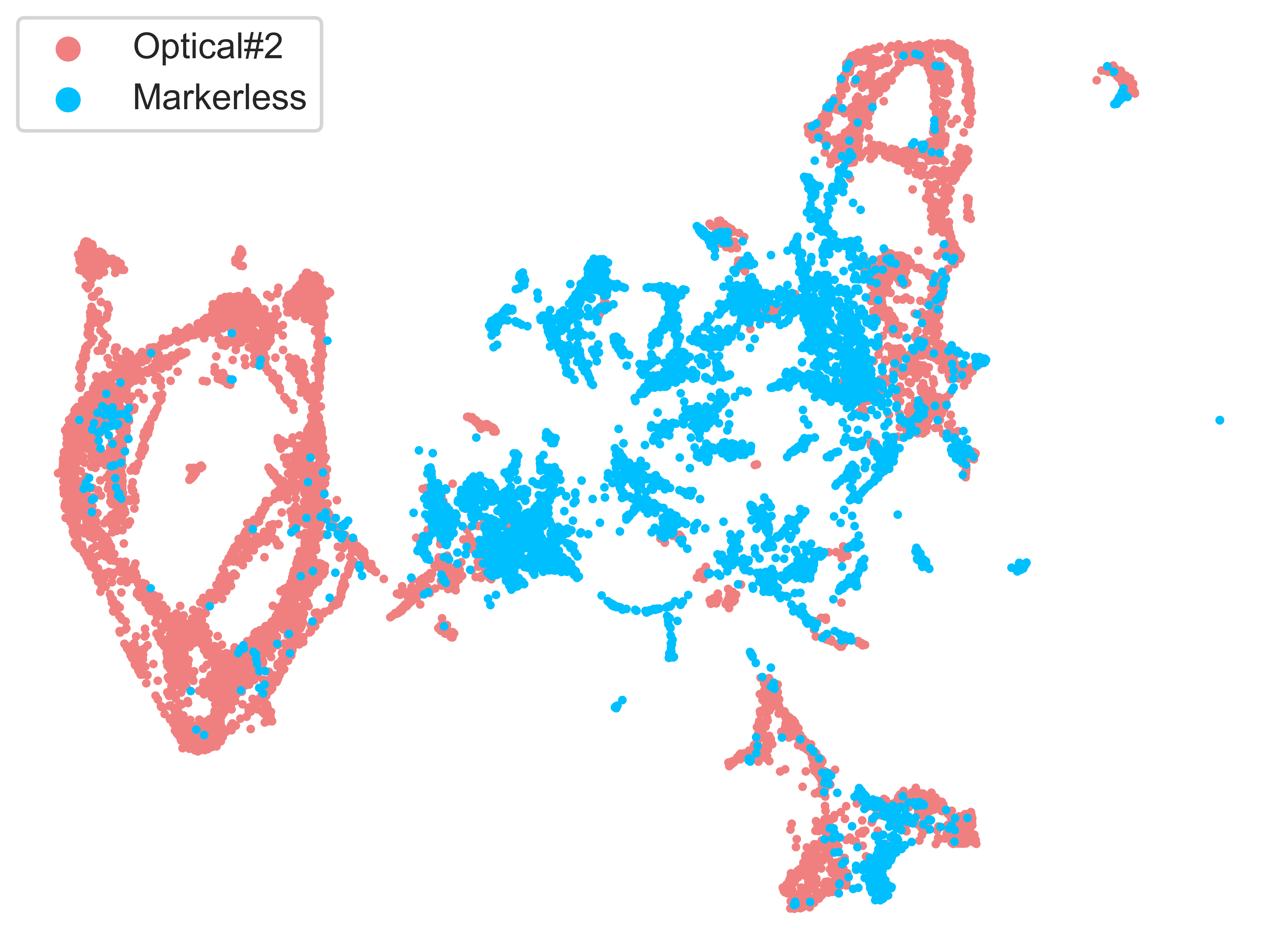}
\end{subfigure}\hfill
\begin{subfigure}{0.22\textwidth}
    \includegraphics[width=\textwidth]{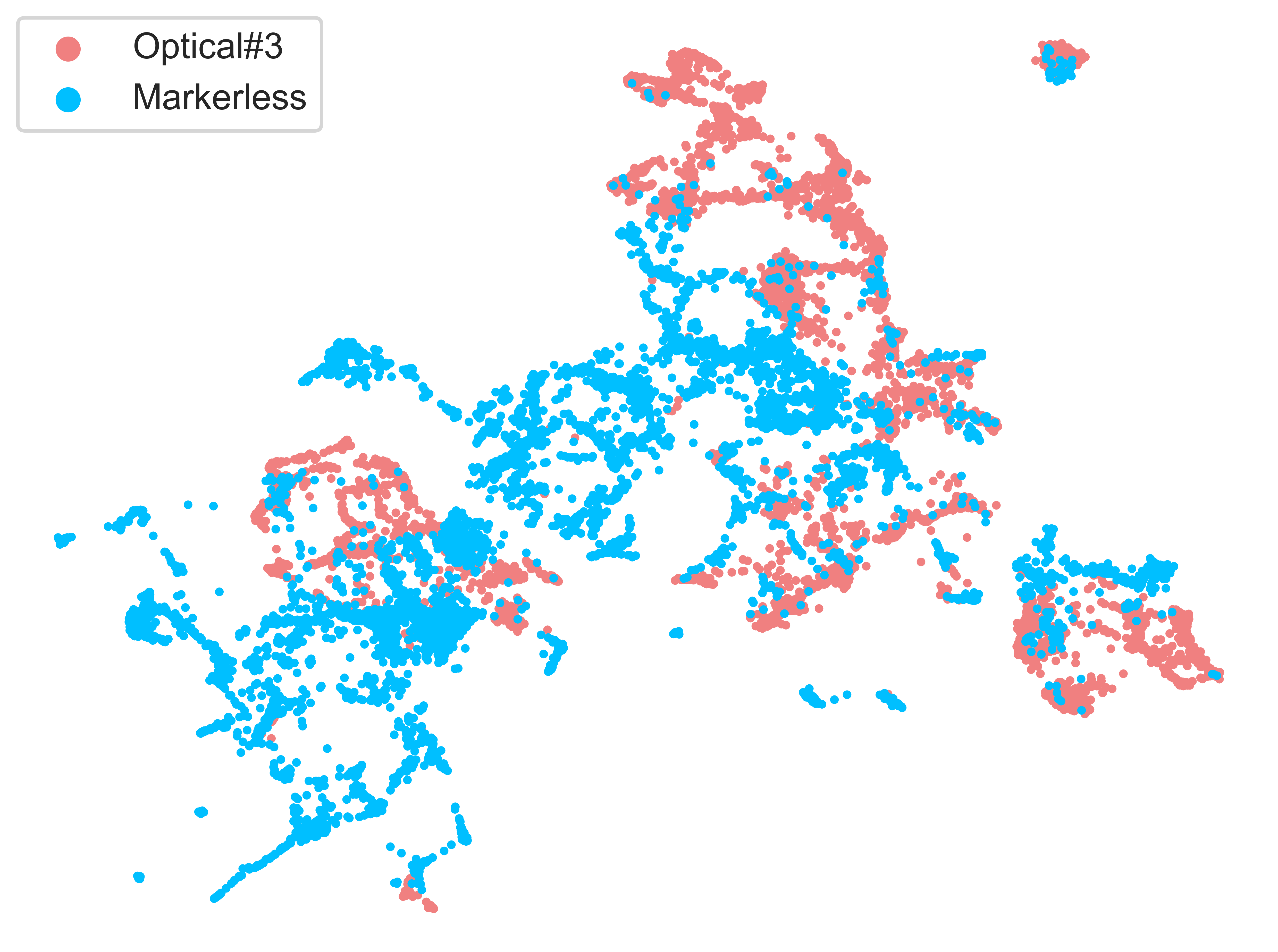}
\end{subfigure}\hfill
\begin{subfigure}{0.22\textwidth}
    \includegraphics[width=\textwidth]{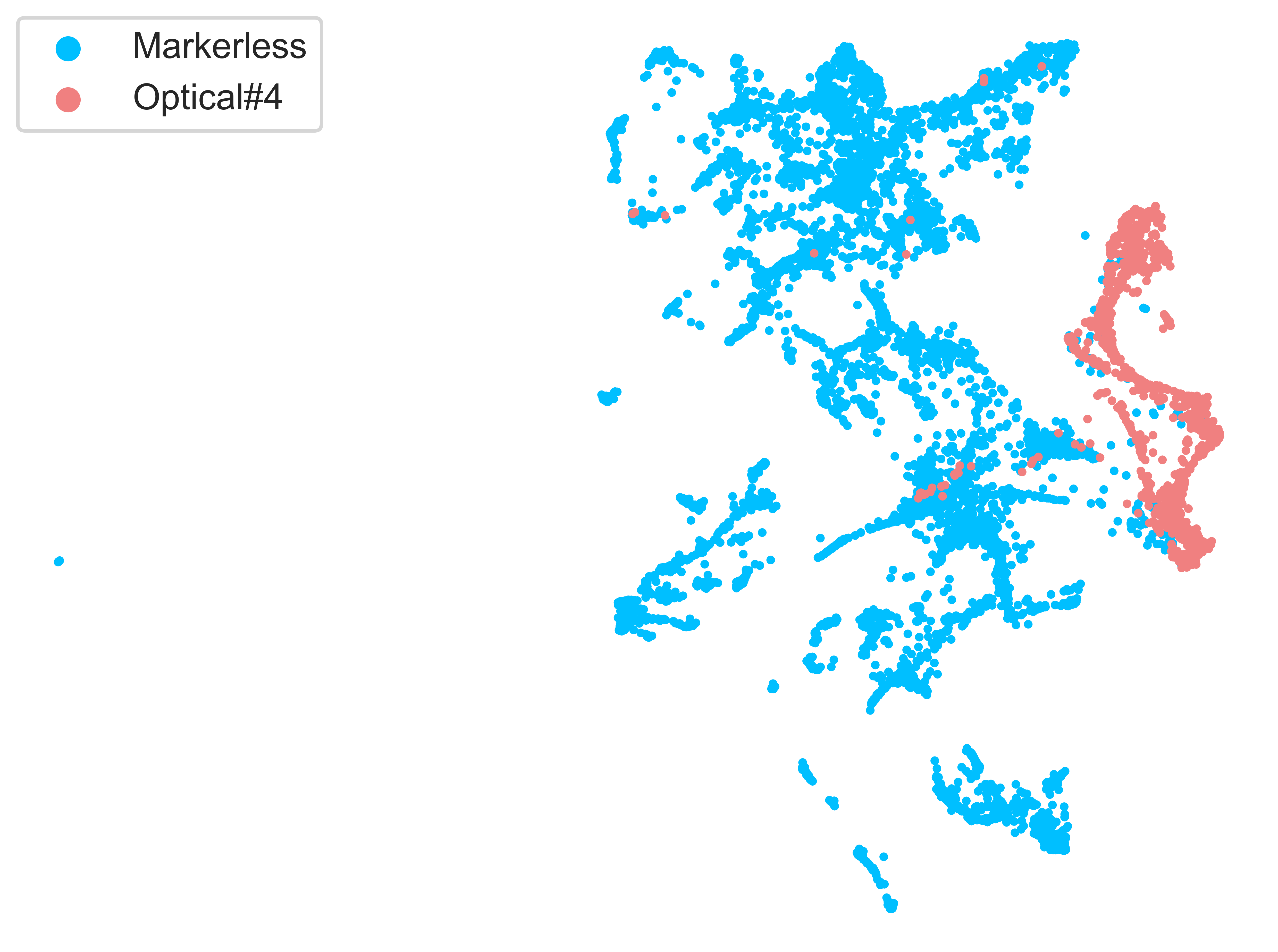}
\end{subfigure}

\begin{subfigure}{0.22\textwidth}
    \vspace{10pt}
    \includegraphics[width=\textwidth]{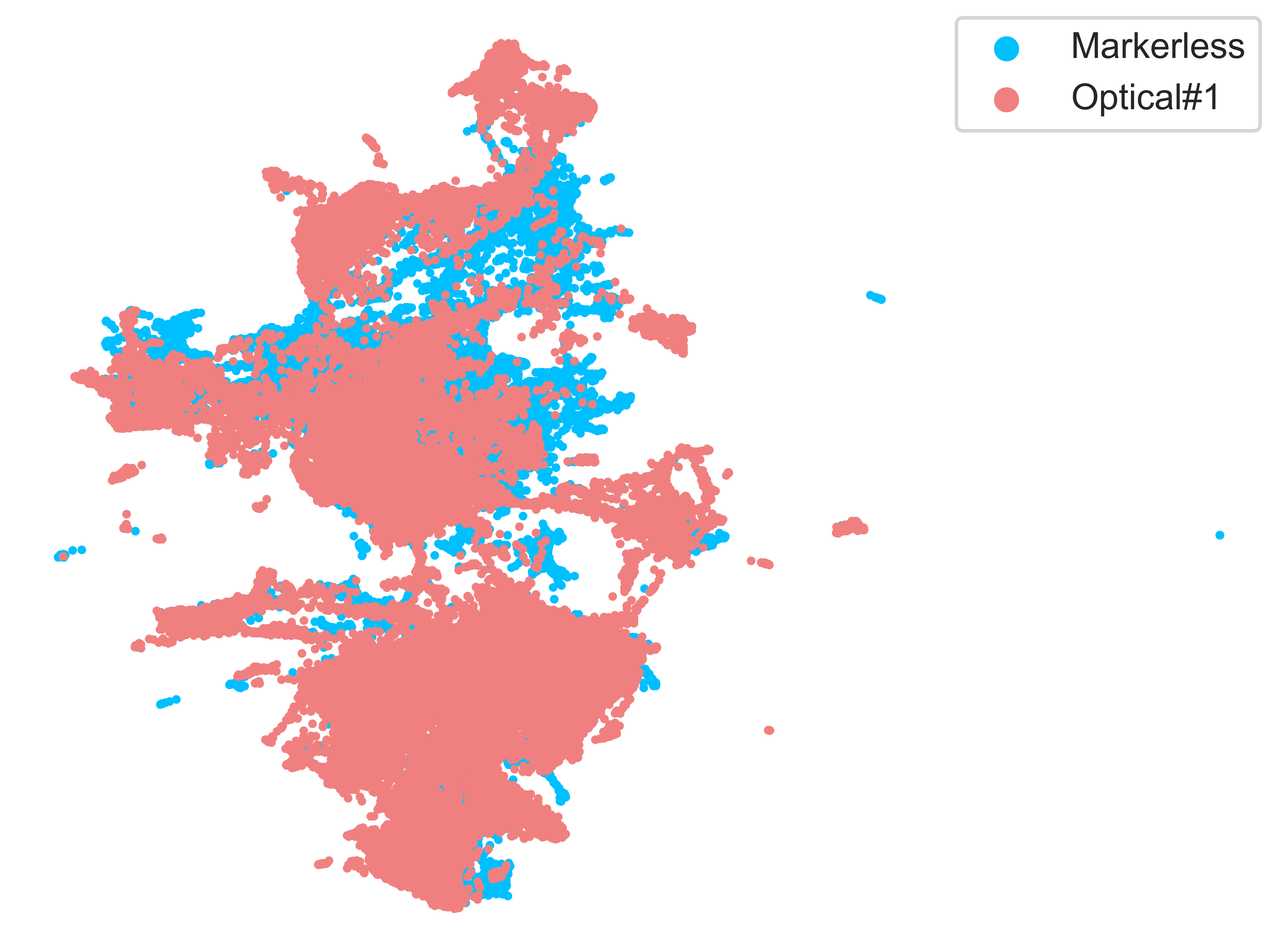}
\end{subfigure}\hfill
\begin{subfigure}{0.22\textwidth}
    \vspace{10pt}
    \includegraphics[width=\textwidth]{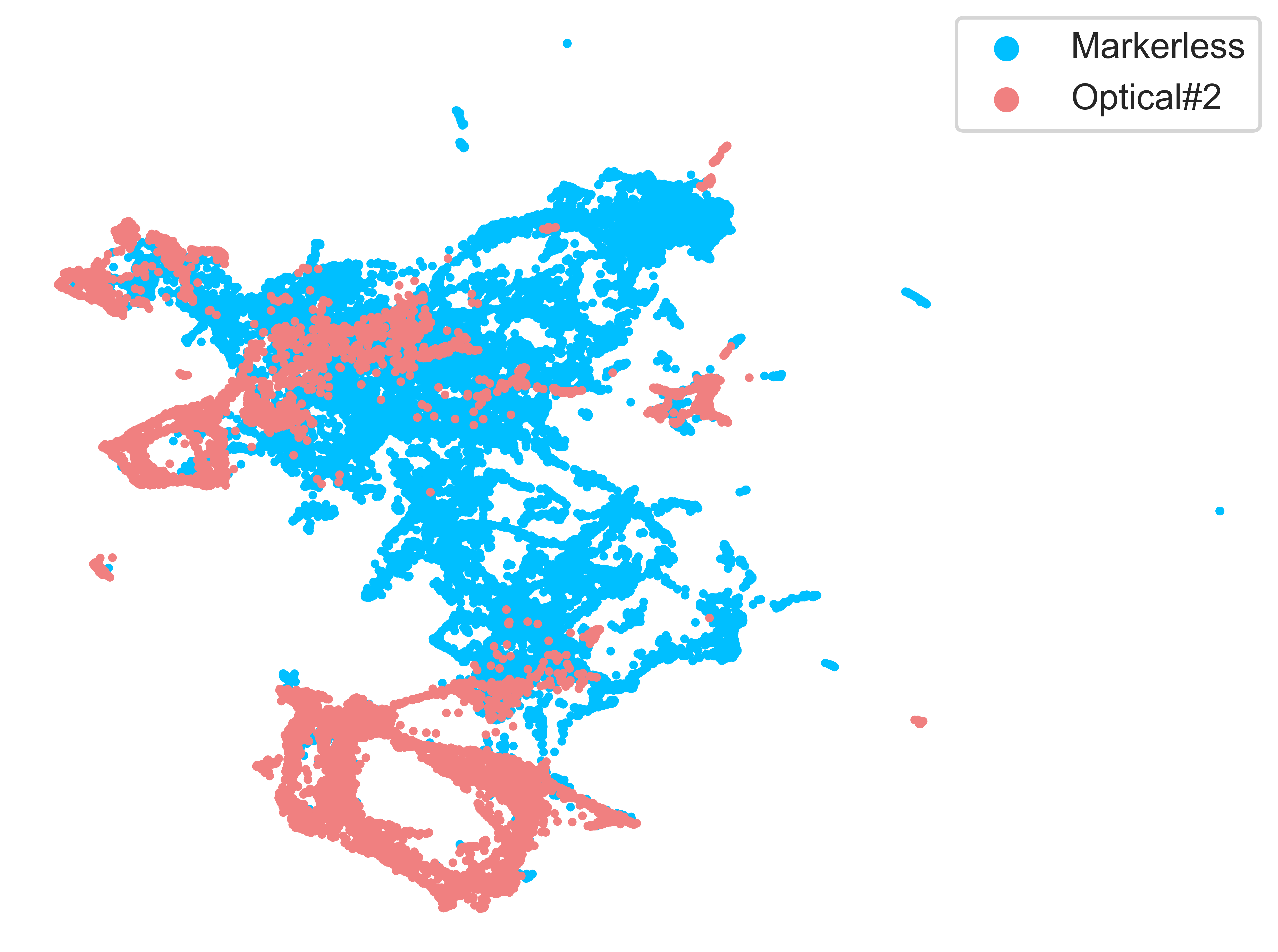}
\end{subfigure}\hfill
\begin{subfigure}{0.22\textwidth}
    \vspace{10pt}
    \includegraphics[width=\textwidth]{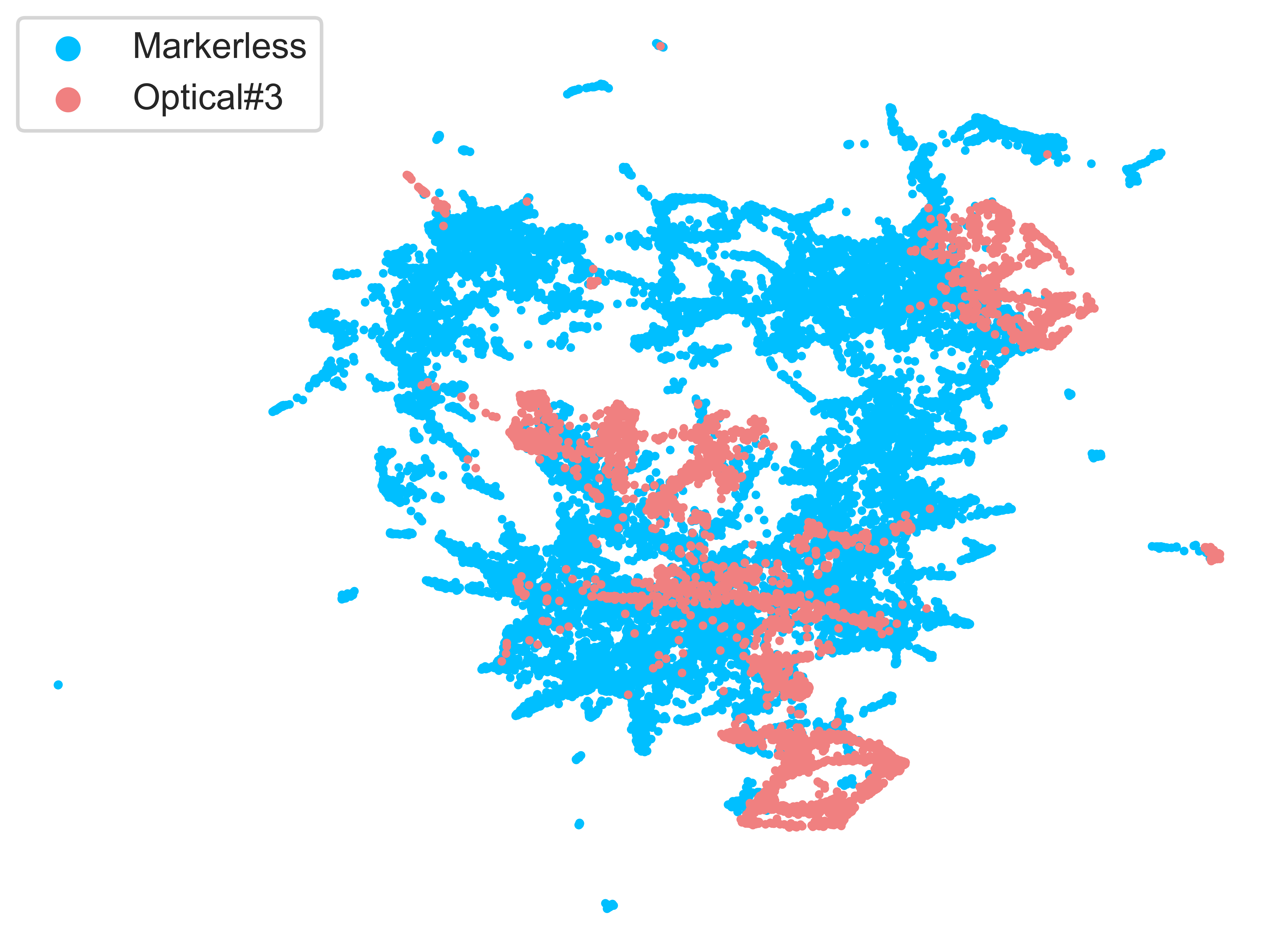}
\end{subfigure}\hfill
\begin{subfigure}{0.22\textwidth}
    \vspace{10pt}
    \includegraphics[width=\textwidth]{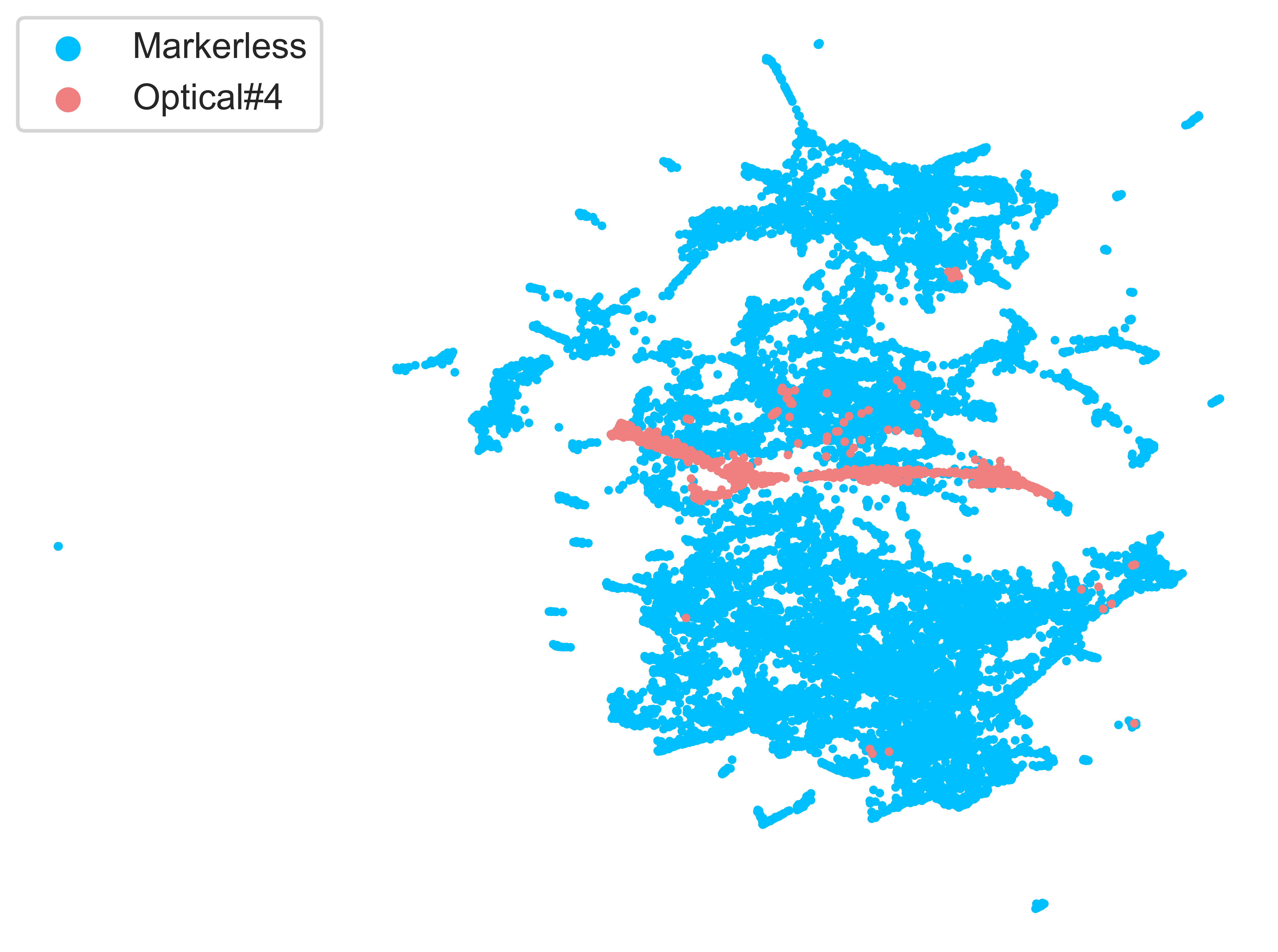}
\end{subfigure}
    
\caption{UMAP projections~\cite{umap} on datasets collected using high-end MoCap systems and others collected from a multiview markerless fitting process. 
The first row uses the markerless\#1 dataset and the second row uses the markerless\#2 dataset.
It can be seen that the variability of data is independent of the type of acquisition. }
\label{fig:embeddings}

\end{figure*}

To supplement this point, \cref{fig:embeddings} offers comparative visualizations of the encoded pose parameters $\pose$ vectors' distribution for each dataset combination. 

\begin{table}[!htbp]
\resizebox{\columnwidth}{!}{%
\begin{tabular}{cccccc}
                                    &               & \down{RMSE}          & \up{PCK1}        & \up{PCK3}        & \up{PCK7}        \\ \hline
\multirow{6}{*}{\rotatebox{90}{ACCAD}}            & Optical\#1    & \first{50.40} $mm$  & \second{36.14}\% & \first{84.89}\%  & \first{90.90}\%  \\
                                    & Optical\#2    & \third{89.99} $mm$  & \first{41.11}\%  & \second{81.18}\% & \third{86.24}\%  \\
                                    & Optical\#3    & 92.90 $mm$          & \second{39.16}\% & 79.74\%          & 86.08\%          \\
                                    & Optical\#4    & 118.2 $mm$         & 26.21\%          & 64.70\%          & 79.64\%          \\ \cline{2-6} 
                                    & Markerless\#1 & 59.40 $mm$          & 21.70\%          & 79.96\%          & 90.08\%          \\
                                    & Markerless\#2 & \second{57.40} $mm$ & 24.75\%          & \third{80.86}\%  & \second{90.40}\% \\ \hline
\multirow{6}{*}{\rotatebox{90}{Tail\#1}} & Optical\#1    & \first{23.80} $mm$  & \third{17.04}\%  & \second{86.67}\% & \first{99.26}\%  \\
                                    & Optical\#2    & \third{37.50} $mm$  & \second{19.26}\% & 76.30\%          & 95.56\%          \\
                                    & Optical\#3    & 41.30 $mm$          & \third{17.04}\%  & 70.74\%          & 94.81\%          \\
                                    & Optical\#4    & 116.8 $mm$         & 5.55\%           & 44.07\%          & 70.74\%          \\ \cline{2-6} 
                                    & Markerless\#1 & 33.50 $mm$          & 12.59\%          & \third{82.96}\%  & \second{98.52}\% \\
                                    & Markerless\#2 & \second{28.85} $mm$ & \first{20.00}\%  & \first{87.77}\%  & \third{98.14}\%  \\ \hline
\multirow{6}{*}{\rotatebox{90}{Tail\#2}} & Optical\#1    & \first{26.70} $mm$  & \second{15.26}\% & \first{84.33}\%  & \second{97.55}\% \\
                                    & Optical\#2    & 57.70 $mm$          & 13.89\%          & 71.27\%          & 89.84\%          \\
                                    & Optical\#3    & 72.80 $mm$          & \third{14.64}\%  & 67.16\%          & 86.48            \\
                                    & Optical\#4    & 123.8 $mm$         & 5.16\%           & 44.63\%          & 71.54\%          \\ \cline{2-6} 
                                    & Markerless\#1 & \second{29.50} $mm$ & 13.43\%          & \second{82.34}\% & \first{97.68}\%  \\
                                    & Markerless\#2 & \third{33.70} $mm$  & \first{18.19}\%  & \third{82.11}\%  & \third{95.11}\%  \\ \hline
\multirow{6}{*}{\rotatebox{90}{Tail\#3}} & Optical\#1    & \first{71.40} $mm$  & \first{13.89}    & \first{57.78}\%  & \first{82.22}\%  \\
                                    & Optical\#2    & \third{300.0} $mm$  & \third{3.33}     & \third{10.56}\%  & \third{19.44}\%  \\
                                    & Optical\#3    & 300.1 $mm$          & 0.5\%            & \third{10.56\%}  & 17.22\%          \\
                                    & Optical\#4    & 309.1 $mm$         & 0.5\%            & 6.67\%           & 12.78\%          \\ \cline{2-6} 
                                    & Markerless\#1 & \second{222.0} $mm$ & \second{2.22}\%  & \second{22.78}\% & \second{40.56}\% \\
                                    & Markerless\#2 & 248.0 $mm$         & \second{2.22}\%  & 16.11 \%         & 30.33\%          \\ \hline
\multirow{6}{*}{\rotatebox{90}{Tail\#4}} & Optical\#1    & \first{68.30} $mm$  & \second{11.30}\% & \second{59.90}\% & \second{88.36}\% \\
                                    & Optical\#2    & \third{280.2} $mm$  & 7.00\%           & \third{37.87}\%  & 60.58\%          \\
                                    & Optical\#3    & 343.5 $mm$          & 6.43\%           & 36.91\%          & 60.77\%          \\
                                    & Optical\#4    & 374.4 $mm$         & 4.07\%           & 20.25\%          & 36.33\%          \\ \cline{2-6} 
                                    & Markerless\#1 & \second{76.60} $mm$ & \third{10.68}\%  & \second{58.65}\% & \third{86.71}\%  \\
                                    & Markerless\#2 & 77.56 $mm$          & \first{13.10}\%  & \first{62.90}\%  & \first{89.23}\%  \\ \hline
\end{tabular}
}
\caption{Markerless vs optical data tested on ACCAD and tail test sets. Models trained on data sourced from a multi-view markerless fitting process perform on par with models trained on high-quality Optical data.}
\label{tab:real_vs_mocap_data}
\end{table}

\section{Balancing Regression}
\label{sec/sup:balance}
\subsection{Robust VPoser}
\label{subsec/sup:vae}
G.~Pavlakos~\etal~\cite{pavlakos2019expressive} were the first to leverage a Variational Autoencoder (VAE)~\cite{kingma2015iclr} instead of Gaussian mixture models to learn a pose prior by folding axis-angle embeddings around a Gaussian distribution. 
Apart from VAEs, pose - and by extension, motion-priors have been learned using other generative models~\cite{GAN-S} or by mapping the pose space on a surface-like manifold~\cite{tiwari2022pose}. 
However, in this paper, we choose to focus on autoencoding generative models, as the trained model operates as a rare pose generator, as well as to reconstruct poses and providing input to the relevance function of our balanced regression model (see Section 3.1 of the main paper).

As noted in the works above, VAEs have certain drawbacks; due to the lack of other constraints. 
The learned prior tends to be mean-centered while the manifold ``folded" around the Gaussian includes several ``dead" regions that could lead to non-plausible data generation. 
These drawbacks would make a fitting process hard as the prior would serve as a regularizer. 
However, we choose to focus on the controllable generation of tail samples, as well as the use of the VAE for re-weighting each sample's contribution to the batch loss during training. 
That is, we focus our experiments on comparing our VPoser variant termed Robust VPoser (RVPoser) with the model from \cite{pavlakos2019expressive} for tail-sample generation.

Our RVPoser follows a similar structure to the VPoser's, with 3 main differences: a) we do not use batch normalization \cite{ioffe2015batch} prior to the first fully-connected layer of the encoder, b) we do not use any dropout layers in the decoder, and c) we do not use any activation function after the last fully-connected of the decoder. 
We train RVPoser using the CMU, Transitions, and PosePrior datasets, while our total training loss can be decomposed into the following losses:
\begin{equation}
\label{lvae}
    \mathcal{L}_{VAE} = \lambda_{1}\mathcal{L}_{KL} + \lambda_{2}\mathcal{L}_{rec} + \lambda_{3}\mathcal{L}_{orth}
\end{equation}
\begin{equation}
\label{lkl}
    \mathcal{L}_{KL} = \Psi( D_{KL}(q_{\theta}(z|R) || \mathcal{N}(0,I)))
\end{equation}
\begin{equation}
\label{lrec}
    \mathcal{L}_{rec} = \left\|v - \hat{v}\right\|_{2},
\end{equation}
\begin{equation}
\label{lorth}
    \mathcal{L}_{orth} = \frac{Trace(R^T \hat{R}) - 1}{2},
\end{equation}

\noindent where $z\in R^{32}$ is the 32-dim latent code, $R\in \mathbb{SO}(3)^P$ is the rotation matrix for each pose parameter $P$, while $\hat{R}$ is the rotation matrix output of the decoder. 
$v, \hat{v}$ correspond to the predicted and ground truth vertices, indicating that the reconstruction term incorporates both angular and 3D joint-position errors. 
Instead of using solely the Kullback-Leibler (KL) divergence, we regularize it (as in~\cite{zhang2020arxiv}) using the Charbonnier penalty function $\Psi$, with $\Psi(x) = \sqrt{1+x^{2}}-1$~\cite{charb1994icip} to prevent posterior collapse and learn a more disentangled manifold. 
\cref{lkl,lrec} follow the VAE training scheme - \textit{e.g.,}~trading of reconstruction quality with learning a Gaussian-like manifold, while \cref{lrec,lorth} force the model to construct a valid rotation latent space. 
We complement RVPoser training with the weight-decaying version of Adam optimization~\cite{loshchilov2019iclr}, which penalizes large weights and prevents over-fitting. 

\begin{figure}[b]
    \centering
    \includegraphics[width=\columnwidth]{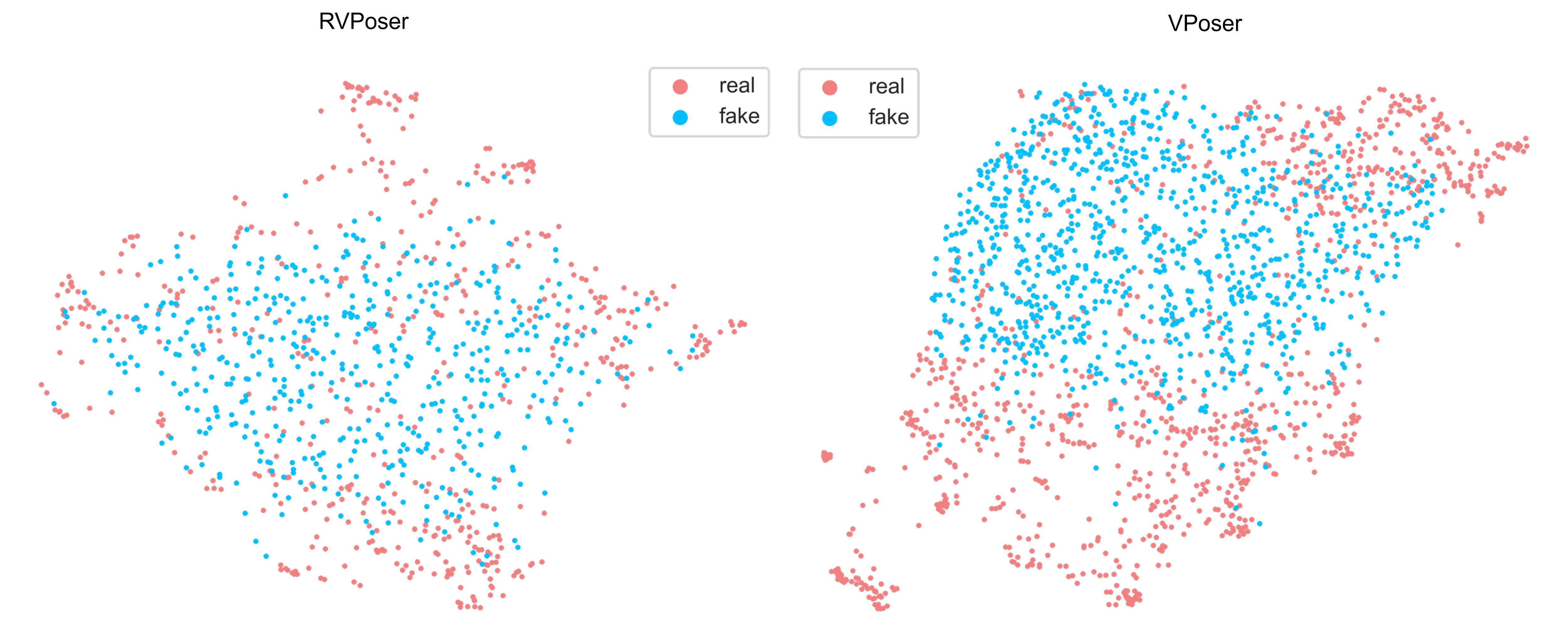}
    \caption{
    UMAP projections \cite{umap} of “real” ground truth samples and of “fake” ones generated by our RVPoser (left) and the VPoser~\cite{pavlakos2019expressive} (right) models, respectively.
    }
    \label{fig:umap_latent}
\end{figure}

We choose to evaluate the 2 models on two different settings: a) compare the models in the task of generating realistic and diverse poses, and b) compare the models as priors for the task of fitting human body parameters. We evaluate both tasks on unseen data from the THuman 2.0 dataset which comprises diverse samples with challenging poses. 
From the results presented in \cref{tab:vae}, we observe that RVPoser is able to generate more diverse and faithful poses, while also outperforming VPoser in the fitting task, improving the overall angular error and the pose prediction accuracy (except for PCK7). 
Apart from the quantitative results,  in \cref{fig:umap_latent} we show the UMAP projection~\cite{umap} of $1200$ ground truth pose vectors superposed on 1200 generated ones using VPoser and RVPoser. 
Based on the depicted result, the samples generated with our VAE variant cover significantly more space spanned by the ground truth embeddings. 
That is, our prior can generate more diverse - but still plausible - samples compared to VPoser.

\begin{table}[]
\centering
\resizebox{\columnwidth}{!}{%
\begin{tabular}{lcccccc}
 \multicolumn{1}{l}{} & \multicolumn{2}{c|}{Synthesis}     & \multicolumn{4}{c}{Fitting}        \\ \hline
 & \up{FID}  & \multicolumn{1}{c|}{\up{DIV}}   & \down{MAE}  & \up{PCK1}    & \up{PCK3}    & \up{PCK 7}   \\ \hline
VPoser~\cite{pavlakos2019expressive} & 7.94 & \multicolumn{1}{c|}{12.11} & 2.68$^{\circ}$ & 28.83\% & 89.04\% & \first{99.03\%} \\
 RVPoser (Ours)             & \first{8.57} & \multicolumn{1}{c|}{\first{14.24}} & \first{1.51$^{\circ}$} & \first{53.72\%} & \first{94.57\%} & 98.15\% \\ \hline
\end{tabular}
}
\caption{Quantitative comparison between the VPoser model from \cite{pavlakos2019expressive} and our robust variant (RVPoser) in synthesis and fitting on the THuman 2.0 test set.}
\label{tab:vae}
\end{table}

\subsection{Relevance Function}
\label{subsec/sup:relevance}
\begin{figure}[!t]
    \centering
    \includegraphics[width=\columnwidth]{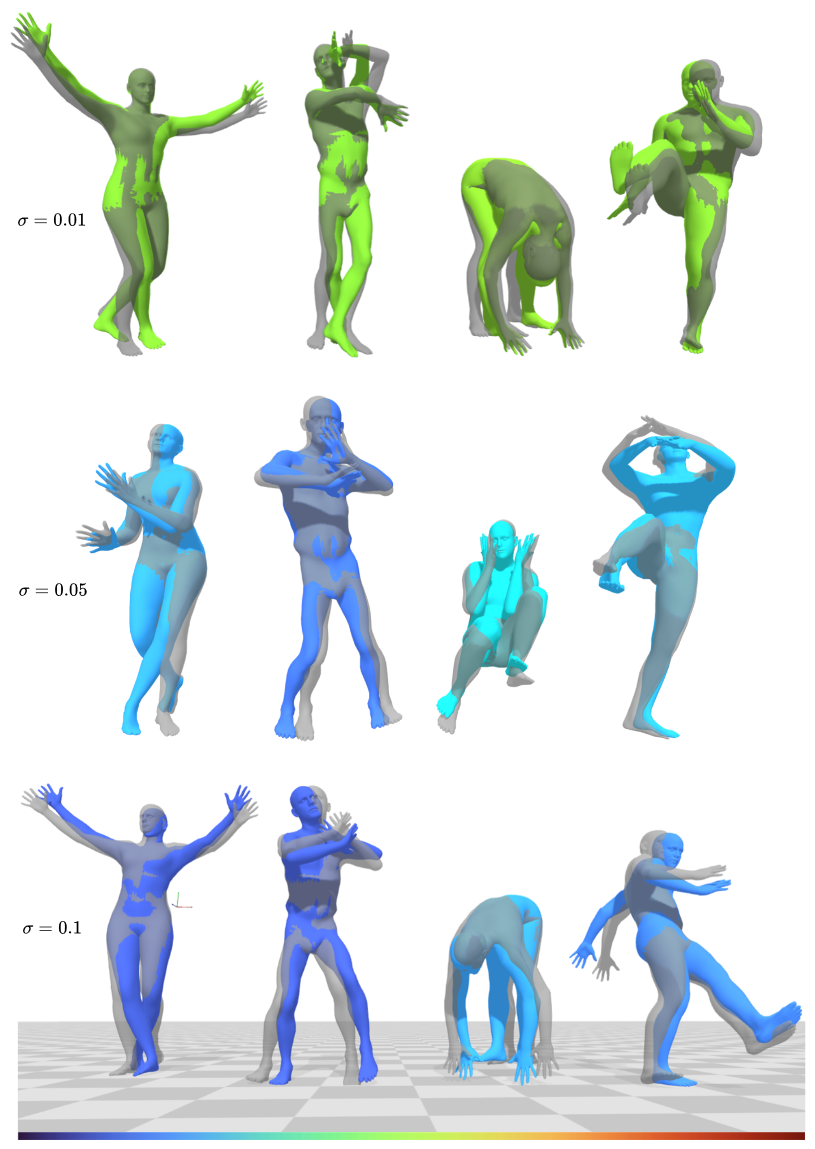}
    \caption{
    Color-coded (turbo colormap \cite{turbo} at the bottom) autoencoding $\rho$ of various poses and $\sigma$ values, using the Sigmoid-based relevance function.
    }
    \label{fig:relevance_sigmoid}
\end{figure}

As stated in the main paper, bias in sample reconstructability can be used to assign relevance to each sample as more challenging (tail) poses are hard to reconstruct accurately. 
As relevance $\rho$, we define the weight used to scale the contribution of each pose to the batch-wide loss.
That is, we need to increase the contribution of the tail poses to the batch loss for every iteration to mitigate the regression bias due to the high number of mean-like poses in our training set. 
We have experimented with 2 different relevance functions, omitting linear weighting as our goal is to boost the contribution of the poses with higher reconstruction error non-linearly.
First, we experimented with the Sigmoid function, focusing on the part that corresponds to the positive input values:

\begin{equation}
\label{eq:relevance_sigmoid}    
    \rho(\theta) = 1 + 2\bigl(\frac{e^{x}}{e^{x}+1}-0.5\bigr), \quad x = \frac{\epsilon}{\sigma},
\end{equation}

\noindent where $\epsilon$ is the normalized-RMSE, $\sigma$ is a scaling factor, and $\theta$ is the given pose parameters as defined in Eq.~(2) of the main paper. 
As shown in \cref{fig:relevance_sigmoid}, the Sigmoid-based $\rho$ - although non-linear - leads to similar error values (colorized) and thus fails to serve our cause in significantly boosting the contribution of the least faithfully reconstructed samples. 
To achieve this, we experiment with a relevance function that scales the error contribution exponentially:

\begin{equation}
\label{eq:relevance_exp}    
    \rho(\theta) = e^{\nicefrac{\epsilon}{\sigma}}.
\end{equation}

\begin{figure}[!b]
    \centering
    \includegraphics[width=\columnwidth]{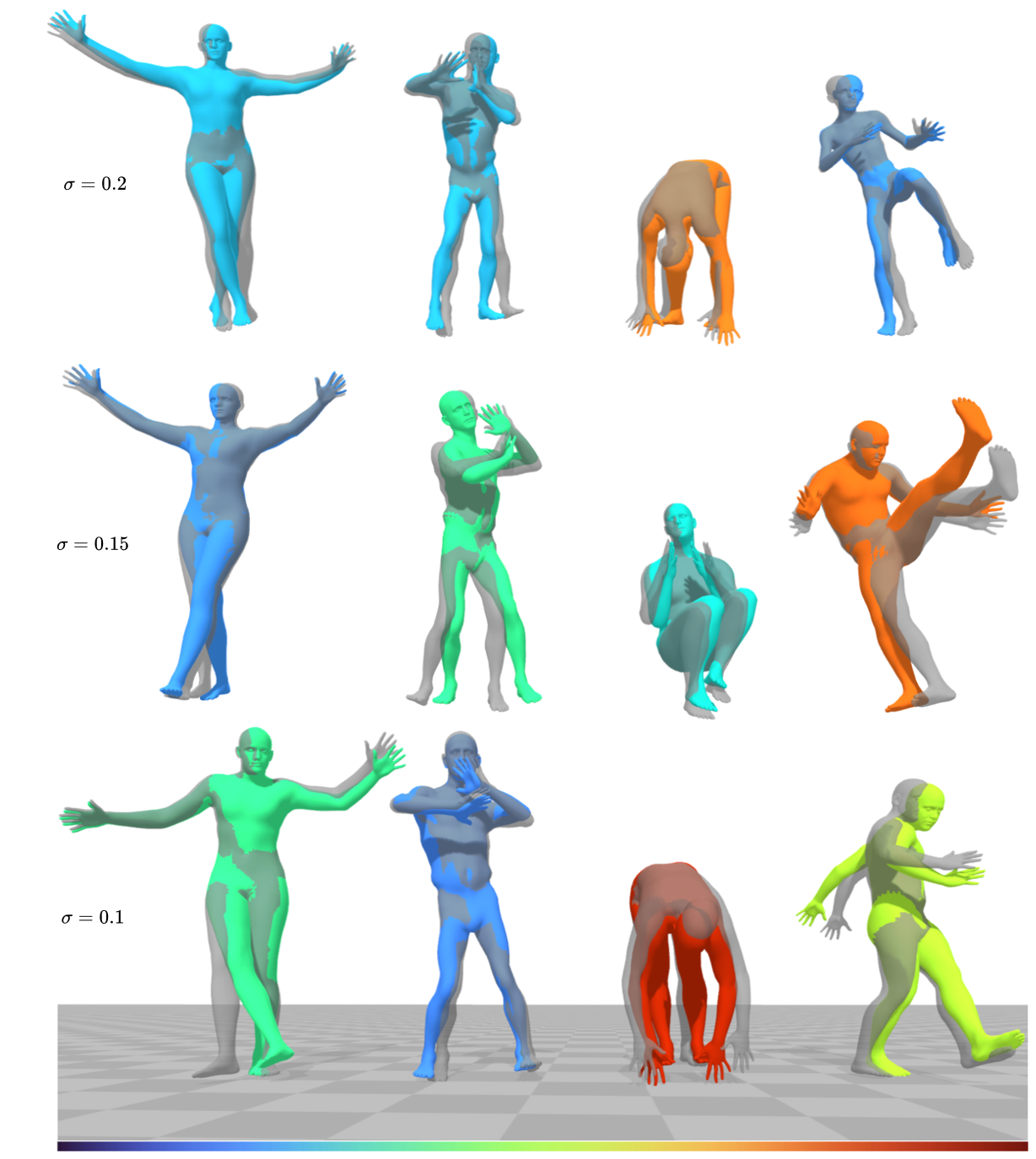}
    \caption{
    Color-coded (turbo colormap \cite{turbo} at the bottom) autoencoding $\rho$ of various poses and $\sigma$ values, using the Exponential-based relevance function.
    }
    \label{fig:relevance_exp}
\end{figure}

Note that since the exponential function does not have an upper limit, we clamp the result at $\rho(\theta)=3$, so the effective range of the weighting function is $[1,3]$, while for the Sigmoid-based relevance function $\rho \in [1,2]$ range.
From the exemplar samples depicted in \cref{fig:relevance_exp}, it can be observed that the exponential relevance function achieves our original goal as it seems to assign a significantly larger weight to higher reconstruction error (colorized). 
Note that the performance of each relevance function for different $\sigma$ values is also depicted in \cref{fig:relevance_sigmoid,fig:relevance_exp}.

\subsection{Orthogonality Investigation}
\label{subsec/sup:orthogonality}
Tab.~2 of the main paper presents the performance of our model against the baseline model (no oversampling or relevance function used) and the same model trained with the Balanced Mean Square Error (BMSE) from \cite{ren2022balanced}. 
Here, we present further details that help us explore the orthogonality of 2 of the contributions of our paper, namely the oversampling and re-weighting through reconstructability methods, as well as the performance of our best model when trained using the BMSE regression loss. 

As shown in \cref{tab:ddist_mix_ablation}, the `Ours' model performs better than the `Sampling' (\textit{i.e.}~oversampling synthetic data) and `Relevance' (\textit{i.e.}~re-weighting the loss) models for both THuman 2.0 and ``tail" test sets. 
This indicates that there is an underlying synergy between oversampling and re-weighting that is horizontal for simple, challenging, and rare poses. We also observe that both variants improve the baseline, while the oversampling variant seems to perform slightly better than the re-weighting one. This result is in line with the feedback from the prior work in unbalanced regression. For the rest of the orthogonality experiments, we choose the `Ours' model as our best-performing one.

Obviously, we have just scratched the surface of the general picture of balancing a regression task and we will keep investigating the complex relationships between different methods that attempt to ``unskew" unbalanced distributions.

\begin{table}[!b]
\centering
\resizebox{\columnwidth}{!}{%
\begin{tabular}{@{}llcccc@{}}
\multicolumn{2}{l}{} & \down{RMSE}                           & \up{PCK1}                        & \up{PCK3}                            & \multicolumn{1}{l}{\up{PCK7}}   \\ \midrule
     & Base          & 21.4 $mm$                         & 28.69\%                     & 92.08\%                         & 98.60\%                     \\
     & Sampling      & \second{20.4} $mm$                         & \third{29.69}\%                     & \third{92.78}\%                         & \second{98.80}\%                     \\
     & Relevance     & \third{20.6} $mm$                         & \second{30.99}\%                     & \second{92.79}\%                         & 98.61\%                     \\
     & Ours          & \first{19.1} $mm$           & \first{32.38}\%                     & \first{93.55}\% & \first{99.11}\%                     \\
     & \cite{ren2022balanced}          & 22.2 $mm$     & 25.51\% & 91.90\%     & \third{98.62}\% \\
\midrule
     & Base          & 35.8 $mm$                         & 22.04\%                     & 80.27\%                         & 94.31\%                     \\
     & Sampling      & \second{31.0} $mm$                         & \second{26.34}\%                     & \second{83.90}\%                         & \second{95.76}\%                     \\
     & Relevance     & 33.9 $mm$                         & \third{23.61}\%                     & 81.00\%                         & \third{95.21}\%                     \\
 & Ours          & \first{29.3} $mm$ & 23.42\%                     & \first{84.70}\% & \first{97.24}\%                     \\
     & \cite{ren2022balanced}          & \third{32.9} $mm$    & \first{27.66}\% & \third{81.98}\%     & 94.92\% \\
\end{tabular}
}
\caption{Imbalanced regression ablation. `Sampling' and `Relevance' variants are combined in `Ours' model, while the results of \cite{ren2022balanced} are presented for reference.}
\label{tab:ddist_mix_ablation}
\end{table}

\subsection{Sampling Ablation}
\label{subsec/sup:synthesis}
\begin{table}[]
\centering
\resizebox{\columnwidth}{!}{%
\begin{tabular}{@{}llcccc@{}}
\multicolumn{2}{l}{} & \down{RMSE}                           & \up{PCK1}                        & \up{PCK3}                            & \multicolumn{1}{l}{\up{PCK7}}   \\ \midrule
     & Base          & \second{21.4 $mm$}                       & 28.69\%                     & 92.08\%                         & \second{98.60\%}                     \\ \cmidrule{2-6}
     & Random      & \second{21.5 $mm$}                         & \first{31.60\%}                     & \third{92.49\%}                         & \second{98.60\%}                     \\
     & LERP     & \third{21.6 $mm$}                         & \third{29.48\%}                     & \second{92.68\%}                         & \third{98.58\%}                     \\
\multirow{-4}{*}{\rotatebox{90}{TH2}} & SLERP          & \first{20.4} $mm$            & \second{29.69}\%                     & \first{92.78}\% & \first{98.80}\%                     \\ \midrule
     & Base          & \third{35.8 $mm$}                         & 22.04\%                     & \third{80.27\%}                         & 94.31\%                     \\ \cmidrule{2-6}
     & Random      & \third{35.8} $mm$                         & \third{23.00\%}                     & \second{81.81\%}                         & \second{95.70\%}                     \\
     & LERP     & \second{33.5 $mm$}                         & \second{25.02\%}                     & 79.82\%                         & \third{95.22\%}                     \\
\multirow{-4}{*}{\rotatebox{90}{Tail}} & SLERP          & \first{31.0} $mm$ & \first{26.34}\%                     & \first{83.90}\% & \first{95.76}\%                     \\ \bottomrule
\end{tabular}
}
\caption{Alternative sampling methods ablation. `SLERP' variant corresponds to the `Sampling' variant in \cref{tab:ddist_mix_ablation}, while `Base' corresponds to the baseline model (\textit{i.e.}~no synthetic samples).}
\label{tab:sampling_ablation}
\end{table}

Our `Sampling' and `Ours' models consist of a specific strategy for sampling from a learned latent space in order to generate diverse, rare, and plausible poses. 
As stated in Section 3.1 of the main paper, this strategy is based on non-linear sampling between $2$ or more anchor samples. 
That is, \ST{we choose samples using statistical thresholding and use them as anchor samples}, avoiding using them in any training or test set. 
Our sampling strategy is to randomly sample a latent vector and add it to one of the anchor vectors.
This helps us achieve extra diversity versus (re)using the anchor vector as is. The next step is to pick a latent sample from the intermediate space between 2 anchor neighborhoods. 
For this purpose, we choose geometric spherical linear interpolation (SLERP) with alternative blending factors in the $[0,1]$ range and compare it with its linear variant `LERP' and the simple random (\textit{i.e.}~no anchors used) sampling (`Random').

\cref{tab:sampling_ablation} presents the performance of our `Sampling' model using each of the 3 different sampling methods on the THuman 2.0 and custom tail test sets, as well as the performance of the `Baseline' for reference. 
\ST{From the results, we can verify that the geometric SLERP helps allows for a safer traversing of the hypersphere-shaped manifold avoiding the dead regions between anchors.} 
This conclusion is supported especially by the performance of SLERP on the ``Tail" set, where the sampling neighborhood can be truly ``away" from the mean of the manifold. 
Another interesting feedback from the presented results is the performance drop of the `Random' variant when tested on the tail set compared with the results for THuman 2.0. This result demonstrates the difference between having to operate on diverse - but possibly still close to the mean - poses and having to estimate rare and complex poses. 
A visual representation of the 3 sampling methods is depicted in Figure 4 of the main paper.

\section{Extra Solving Experiments}
\label{sec/sup:solving}
In the following \cref{tab:solving_vs_sota_seed_200} we compare the performance of our model to a dataset generated with a different seed following \cite{mocap-solver} (denoted as SEED200). 
We observe that the results do not significant vary from those presented in the main paper.

\begin{table}[t]
\resizebox{\columnwidth}{!}{%
\begin{tabular}{c|ccccc}
& \down{RMSE}        & \down{JPE}         & \up{PCK1}        & \up{PCK3}       & \up{PCK7}        \\ \hline
\cite{mocap-solver} & \second{18.20} $mm$ & \second{14.80} $mm$ & 37.19\%          & 85.38\%         & \first{99.37}\%  \\
\cite{democap}      & 22.27 $mm$          & 17.08 $mm$          & \first{49.86}\%  & 88.98\%         & 97.26\%          \\
Ours                & \first{17.90} $mm$ & \first{14.20} $mm$ & \second{48.93}\% & \first{92.55}\% & \second{98.84}\%
\end{tabular}%
}
\caption{
Direct joint solving on CMU \cite{cmuWEB} test set with a different seed (SEED200 from \cite{mocap-solver}) than in the main paper.
}
\label{tab:solving_vs_sota_seed_200}
\end{table}

\section{Landmarks and fitting ablation}
\label{sec/sup:landmarks}
As demonstrated, our noise-aware fitting method is more robust to various types of noise, whether originating from the data, $n_{d}$, the model's inference, $n_{m}$, or both. 
The results in \cref{tab:fit_results} show that our approach maintains its performance across different noise sources, while the method proposed in \cite{barron2019general} may require hyperparameters tuning.

\begin{figure*}[!htp]

\begin{subfigure}{0.33\textwidth}
    \includegraphics[width=\textwidth]{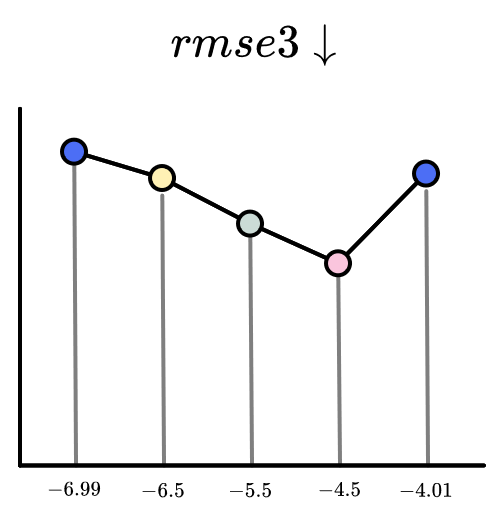}
    \caption{With $\alpha_{range}$ $\in$ [-7,-4], we search for the best $\alpha_{init}$ value.}
\end{subfigure}\hfill
\begin{subfigure}{0.33\textwidth}
    \includegraphics[width=\textwidth]{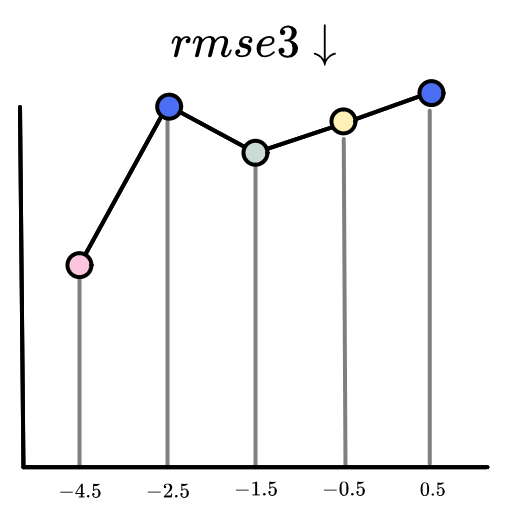}
    \caption{With $\alpha_{range}$ $\in$ [-7, 2], we search for the best $\alpha_{init}$ value. }
\end{subfigure}\hfill
\begin{subfigure}{0.33\textwidth}
    \includegraphics[width=\textwidth]{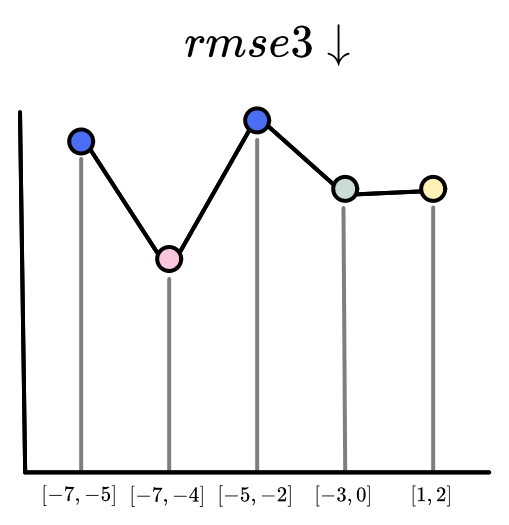}
    \caption{We initialize $\alpha$ to the mean value of $\alpha_{range}$, and search for its best range.}
\end{subfigure}
 \caption{Ablation on $\alpha$ values.}
 \label{fig:alpha_ablation}
\end{figure*}

In addition, we present results that are optimized using both $\markers$ and $\joints$, which further improves performance. 
Our method also has the advantage of adapting the influence of markers and joints on the fit dynamically, which reduces the burden of hyperparameter tuning. In \cref{fig:uncertainty_vs_naive}, we qualitatively compare the performance of our method with that of \cite{soma}, colorised each mesh based on its distance error from the ground truth.
Finally, for a fair comparison with \cite{barron2019general} we conducted several experiments to find the best range of $\alpha$ values, as well as their initial values. \cref{fig:alpha_ablation} reports the values of $rmse3$ with different values of $\alpha$.
Interestingly, we found that the best results are obtained with an $\alpha$ range of [-7, 4] and an initial $\alpha$ value of -4.5.

\begin{table}[t]
\resizebox{\columnwidth}{!}{%
\begin{tabular}{c|cc|ccccc}
\hline
                            & $n_d$                   & $n_m$                   & \down{RMSE}       & \down{MAE}              & \up{PCK1}        & \up{PCK3}        & \up{PCK7}        \\ \hline
\cite{mosh,amass}           & \multirow{4}{*}{\cmark} & \multirow{4}{*}{\xmark} & 30.10 $mm$          & 3.49$^{\circ}$          & 11.79\%          & 66.85\%          & 98.34\%          \\
\cite{barron2019general}    &                         &                         & 30.80 $mm$          & 3.10$^{\circ}$          & 12.71\%          & 67.06\%          & 97.71\%          \\
Ours $(\markers)$           &                         &                         & \second{28.90} $mm$ & \second{2.98}$^{\circ}$ & \second{14.71}\% & \second{69.86}\% & \second{98.18}\% \\
Ours $(\markers | \joints)$ &                         &                         & \first{23.40} $mm$  & \first{2.29}$^{\circ}$  & \first{19.66}\%  & \first{81.06}\%  & \first{99.11}\%  \\ \hline
\cite{mosh,amass}           & \multirow{4}{*}{\xmark} & \multirow{4}{*}{\cmark} & 20.60 $mm$          & 1.93$^{\circ}$          & 28.71\%          & 89.03\%          & \first{99.05}\%  \\
\cite{barron2019general}    &                         &                         & 21.71 $mm$         & 1.91$^{\circ}$          & 36.38\%          & 87.75\%          & 98.22\%          \\
Ours $(\markers)$           &                         &                         & \second{18.70} $mm$ & \second{1.85}$^{\circ}$ & \second{41.99}\%   & \second{90.95}\%   & \second{98.81}\%   \\
Ours $(\markers | \joints)$ &                         &                         & \first{18.50} $mm$  & \first{1.49}$^{\circ}$  & \first{42.18}\%  & \first{91.44}\%  & 98.56\%          \\ \hline
\cite{mosh,amass}           & \multirow{4}{*}{\cmark} & \multirow{4}{*}{\cmark} & 23.80 $mm$          & 2.03$^{\circ}$          & 24.26\%          & 85.63\%          & \first{98.22}\%  \\
\cite{barron2019general}    &                         &                         & 24.87 $mm$         & 1.94$^{\circ}$          & 31.99\%          & 84.05\%          & 97.00\%  \\
Ours $(\markers)$           &                         &                         & \second{22.40} $mm$ & \second{1.79}$^{\circ}$ & \second{36.01}\% & \second{87.14}\% & 97.53\%          \\
Ours $(\markers | \joints)$ &                         &                         & \first{21.90} $mm$  & \first{1.52}$^{\circ}$  & \first{36.67}\%  & \first{88.09}\%  & \second{97.69}\% \\ \hline
\end{tabular}%
}
\caption{
Noisy landmark fitting on THuman 2.0.
}
\label{tab:fit_results}
\end{table}

\section{Additional Qualitative Results}
\label{sec/sup:qualitative}
We present additional qualitative results comparing our direct regression approach to labeling \cite{soma} in the THuman 2.0 and ``Tail" sets. 
These additional results further reinforce the case that a labeling method's errors are more detrimental to fitting performance, even in cases with no noise, as is evident in the \cref{fig:ours_vs_soma}.
Finally, \cref{fig:in_the_wild_qualitative} presents qualitative results using real-world data acquired from the developed system presented in \cref{sec/sup:system}, including both model predictions and post-fitting body results, showcasing the benefits of the noise-aware fitting process.

\begin{figure*}[!htp]

\begin{subfigure}{0.3\textwidth}
    \includegraphics[width=\textwidth]{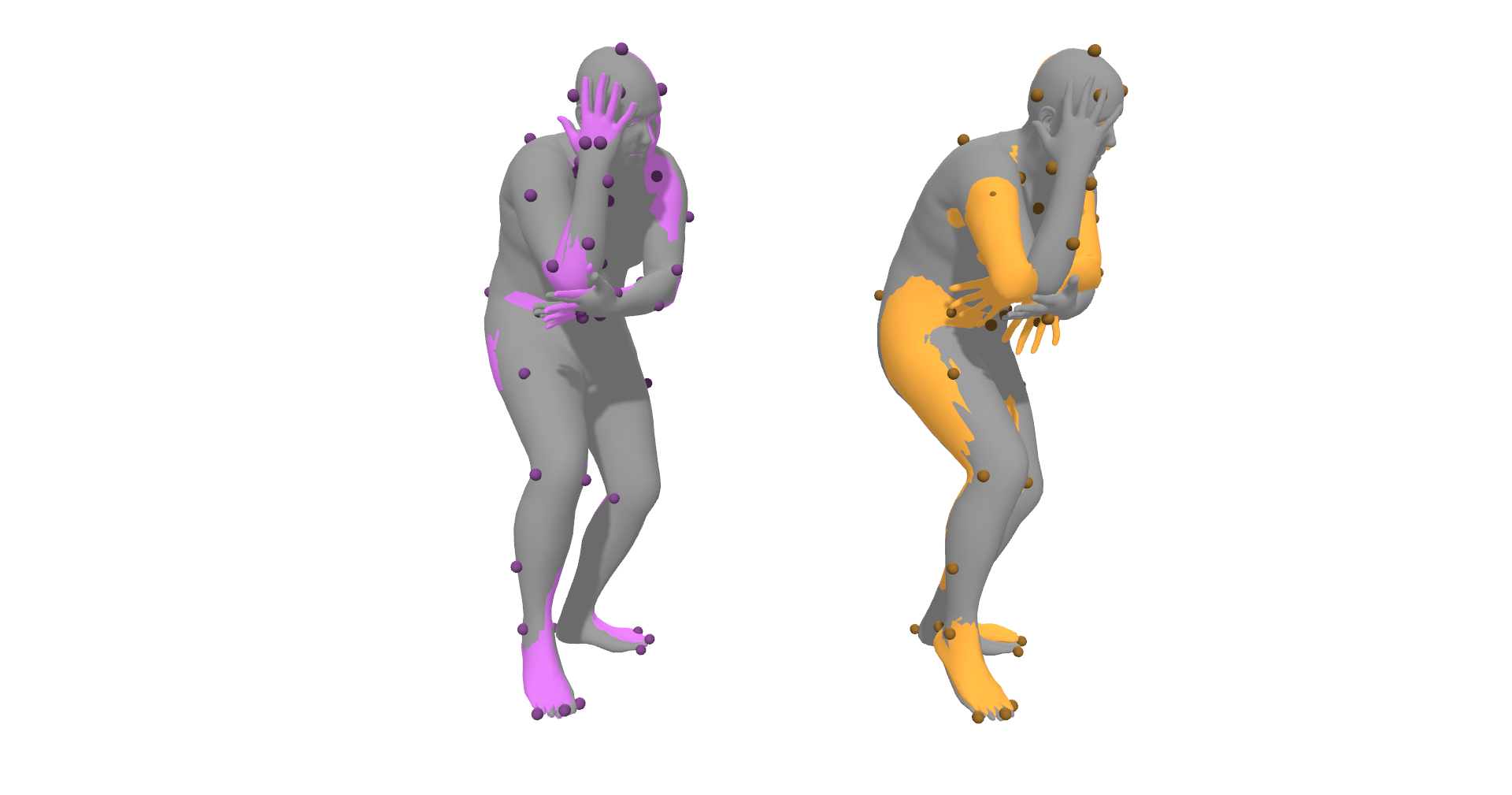}
\end{subfigure}\hfill
\begin{subfigure}{0.3\textwidth}
    \includegraphics[width=\textwidth]{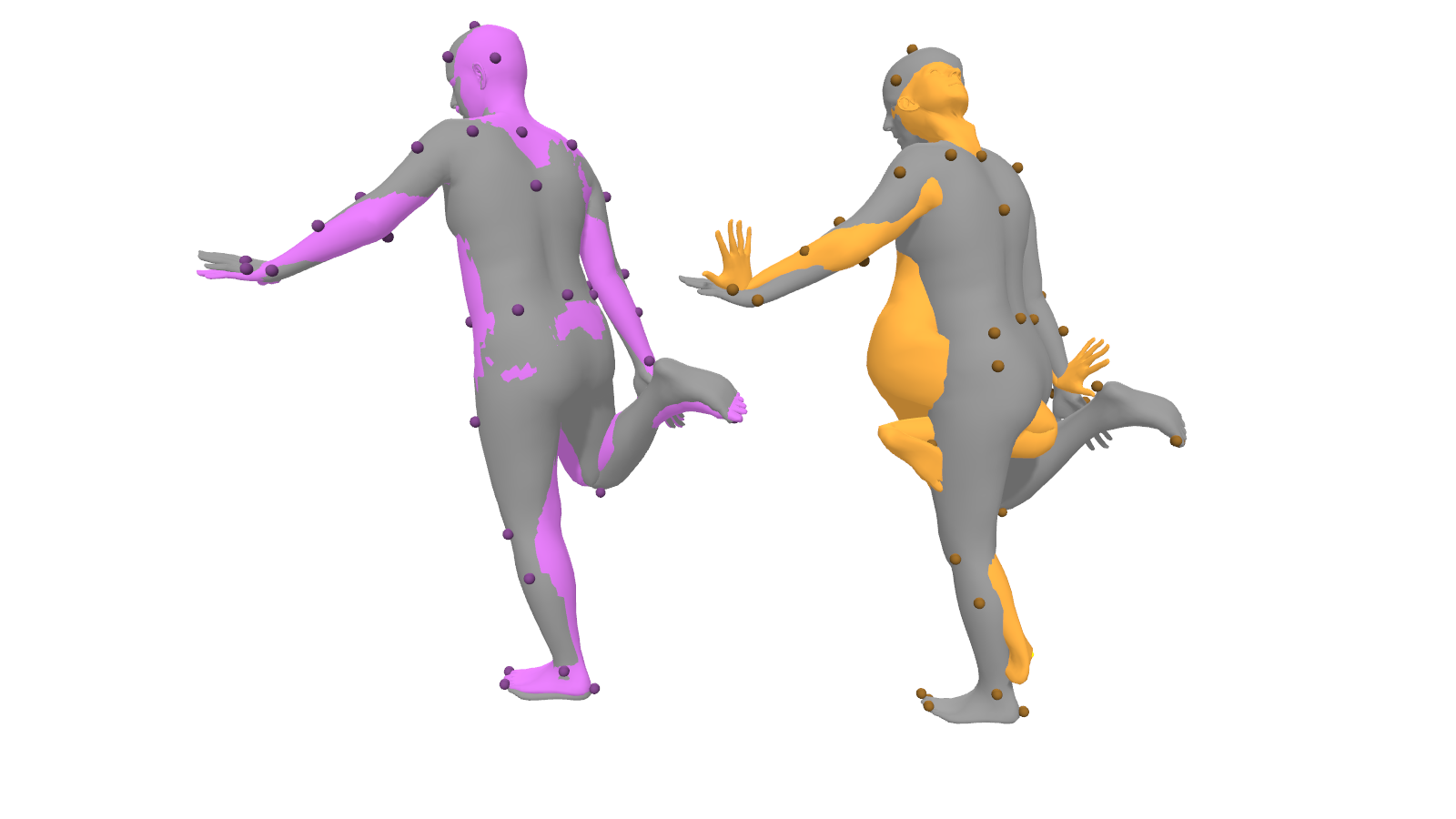}
\end{subfigure}\hfill
\begin{subfigure}{0.3\textwidth}
    \includegraphics[width=\textwidth]{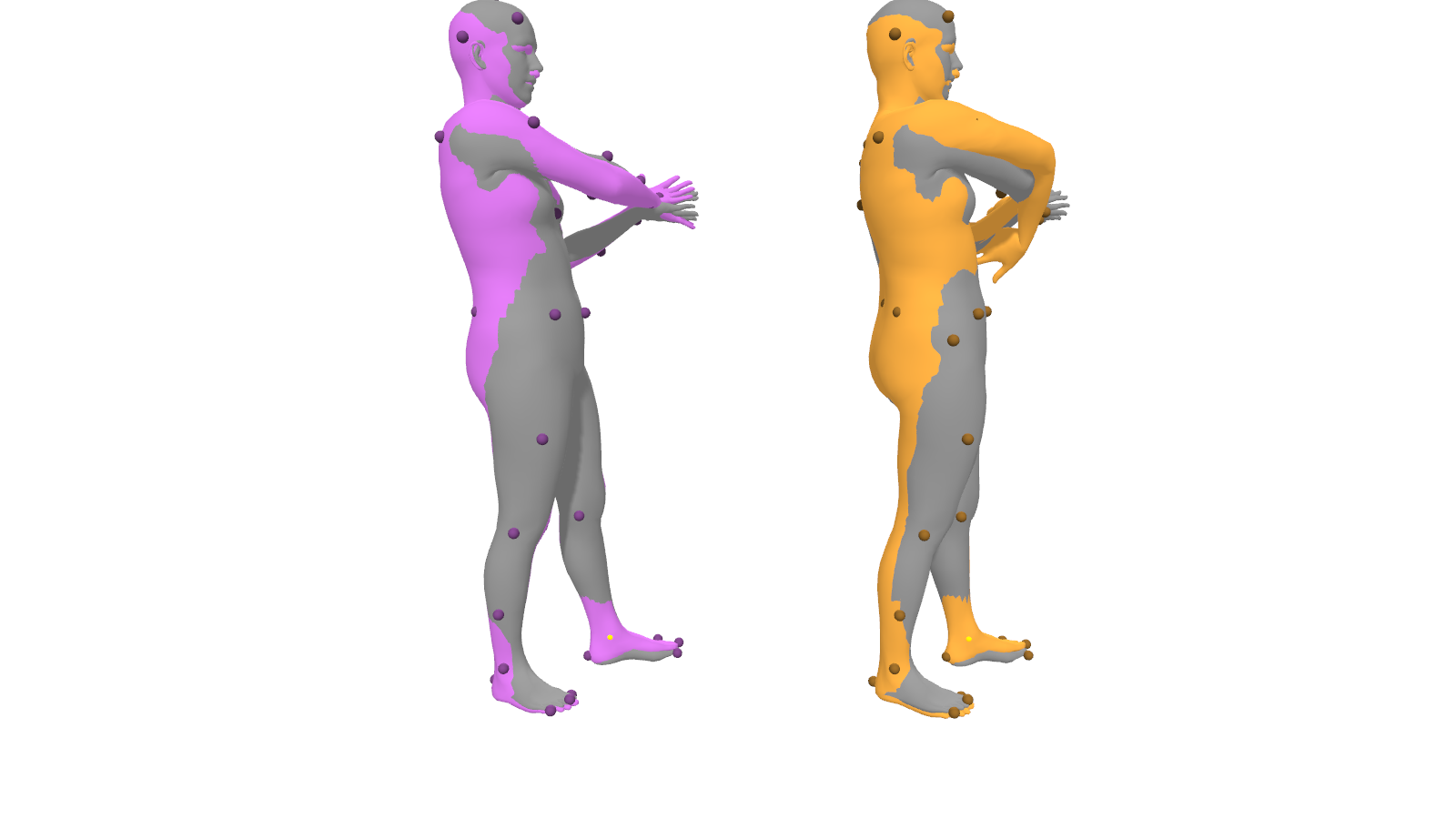}
\end{subfigure}\hfill

\begin{subfigure}{0.3\textwidth}
    \vspace{10pt}
    \includegraphics[width=\textwidth]{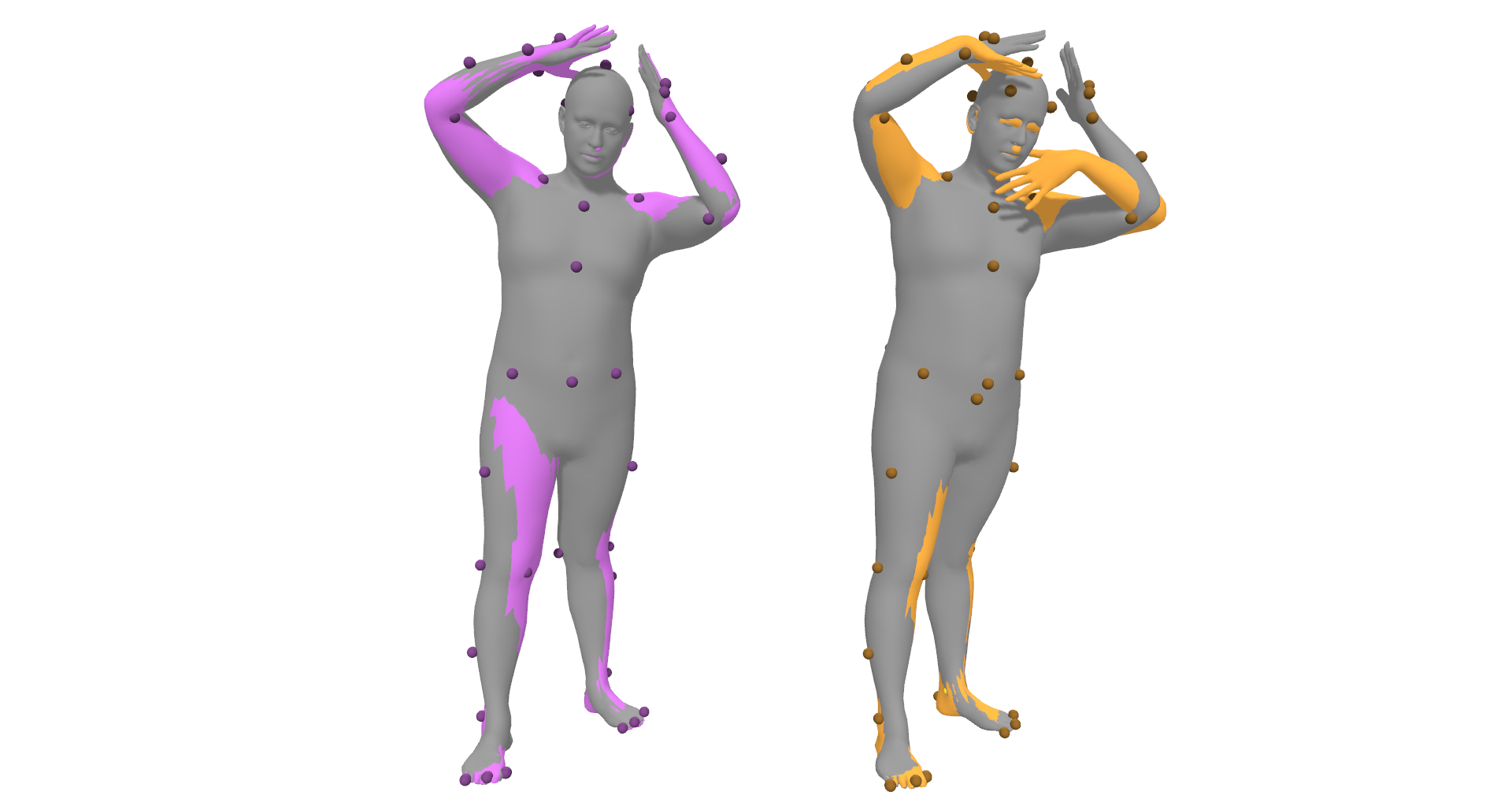}
\end{subfigure}\hfill
\begin{subfigure}{0.3\textwidth}
    \vspace{10pt}
    \includegraphics[width=\textwidth]{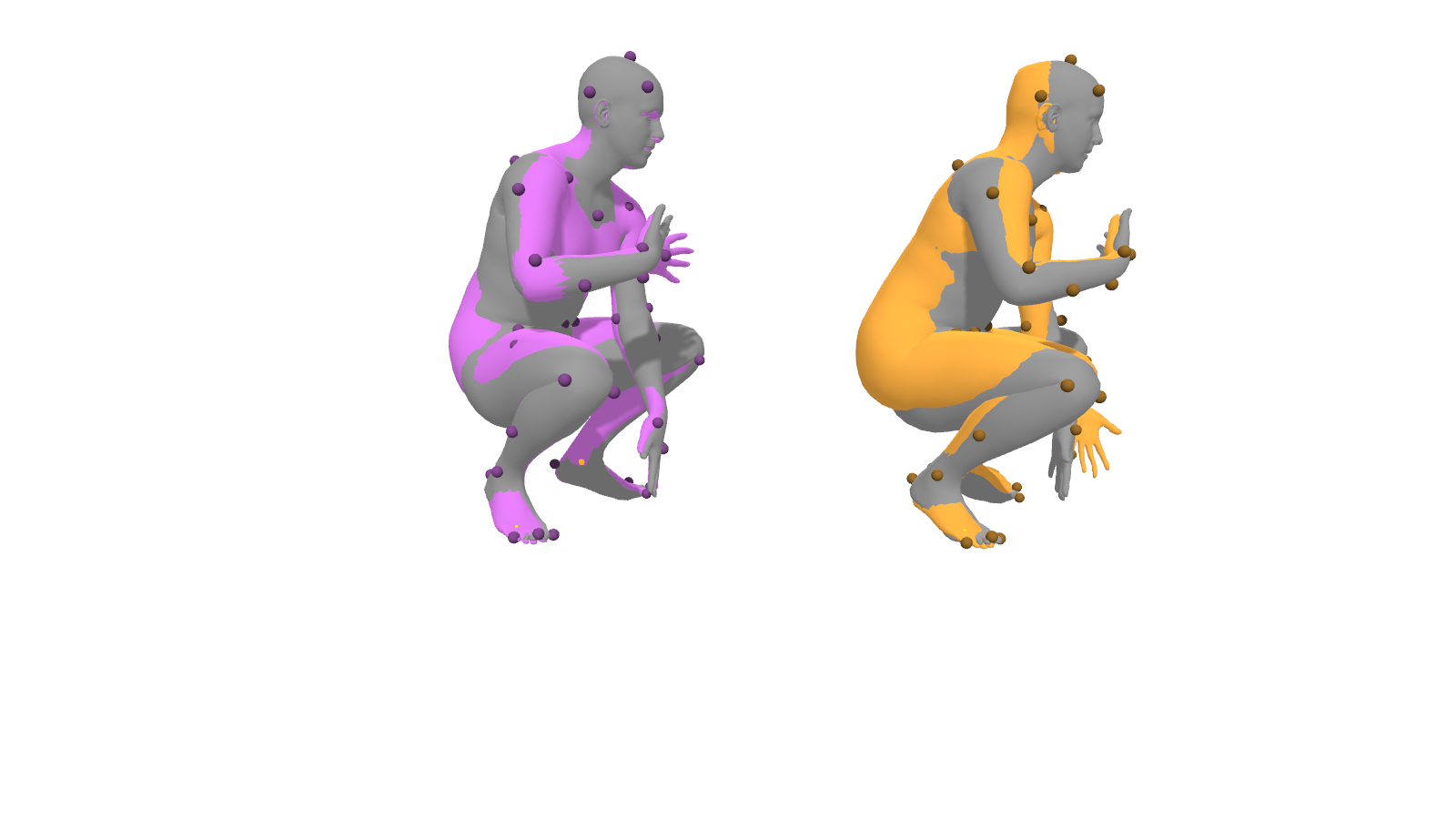}
\end{subfigure}\hfill
\begin{subfigure}{0.3\textwidth}
    \vspace{10pt}
    \includegraphics[width=\textwidth]{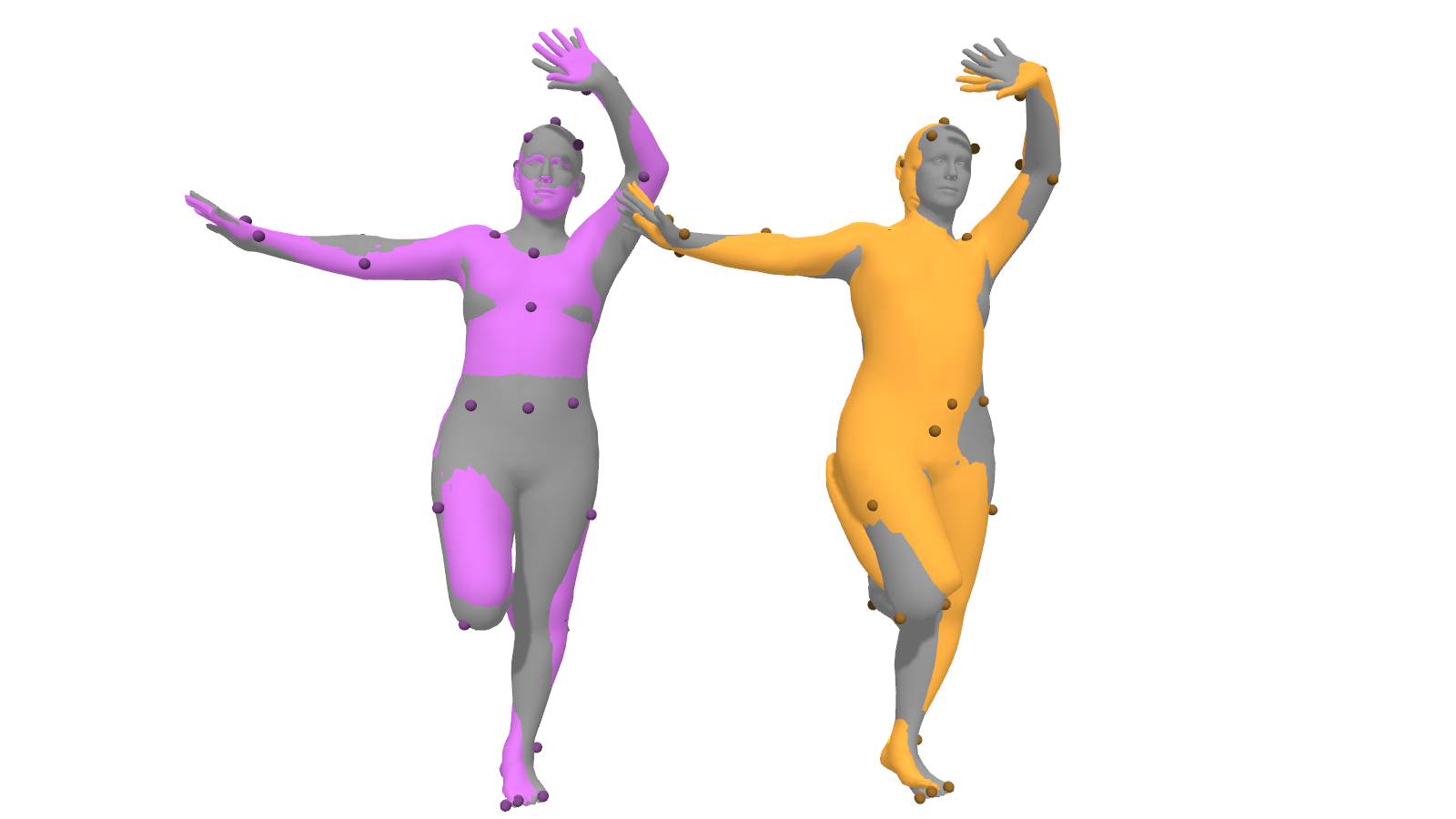}
\end{subfigure}\hfill

\begin{subfigure}{0.3\textwidth}
    \vspace{10pt}
    \includegraphics[width=\textwidth]{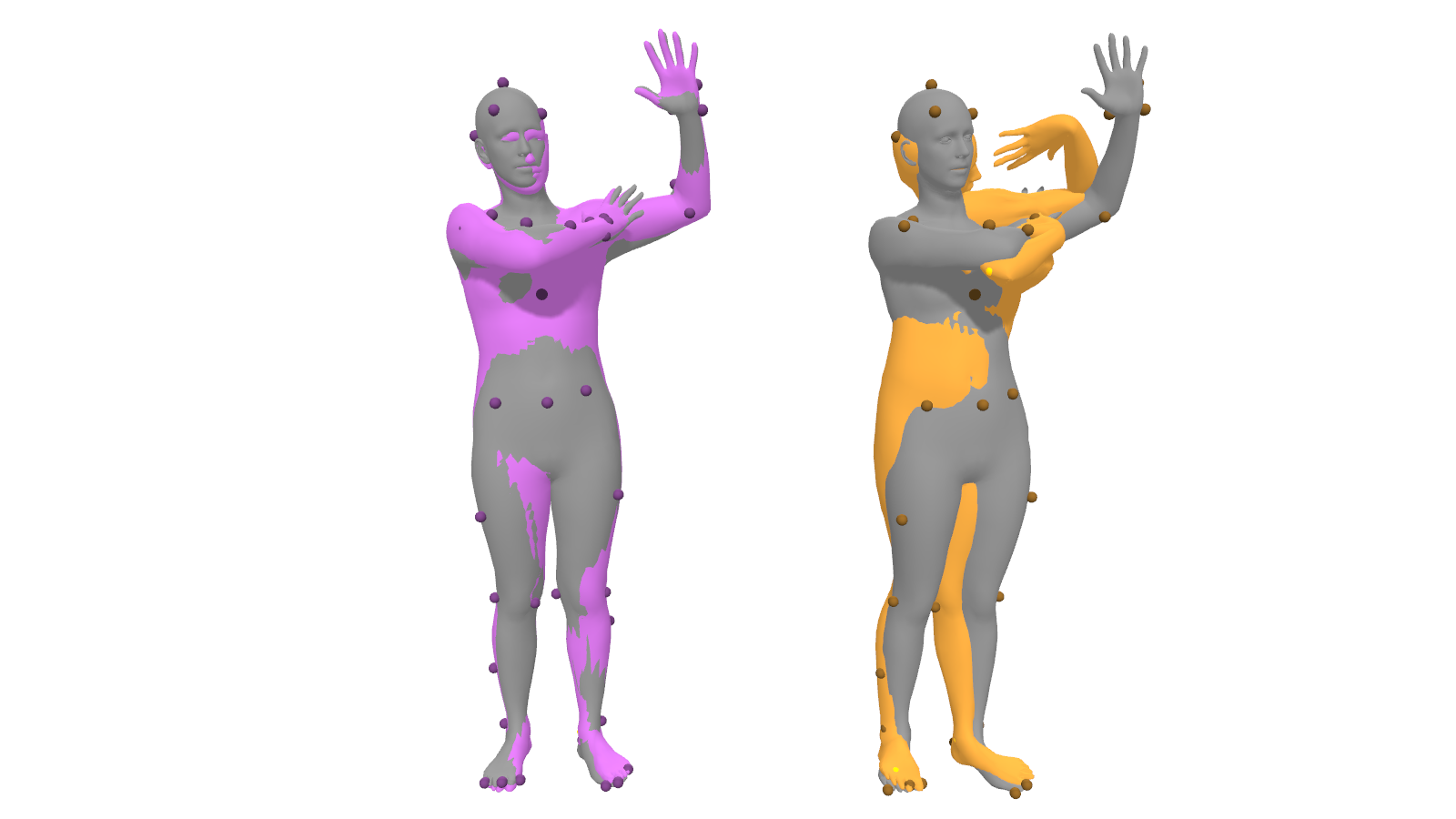}
\end{subfigure}\hfill
\begin{subfigure}{0.3\textwidth}
    \vspace{10pt}
    \includegraphics[width=\textwidth]{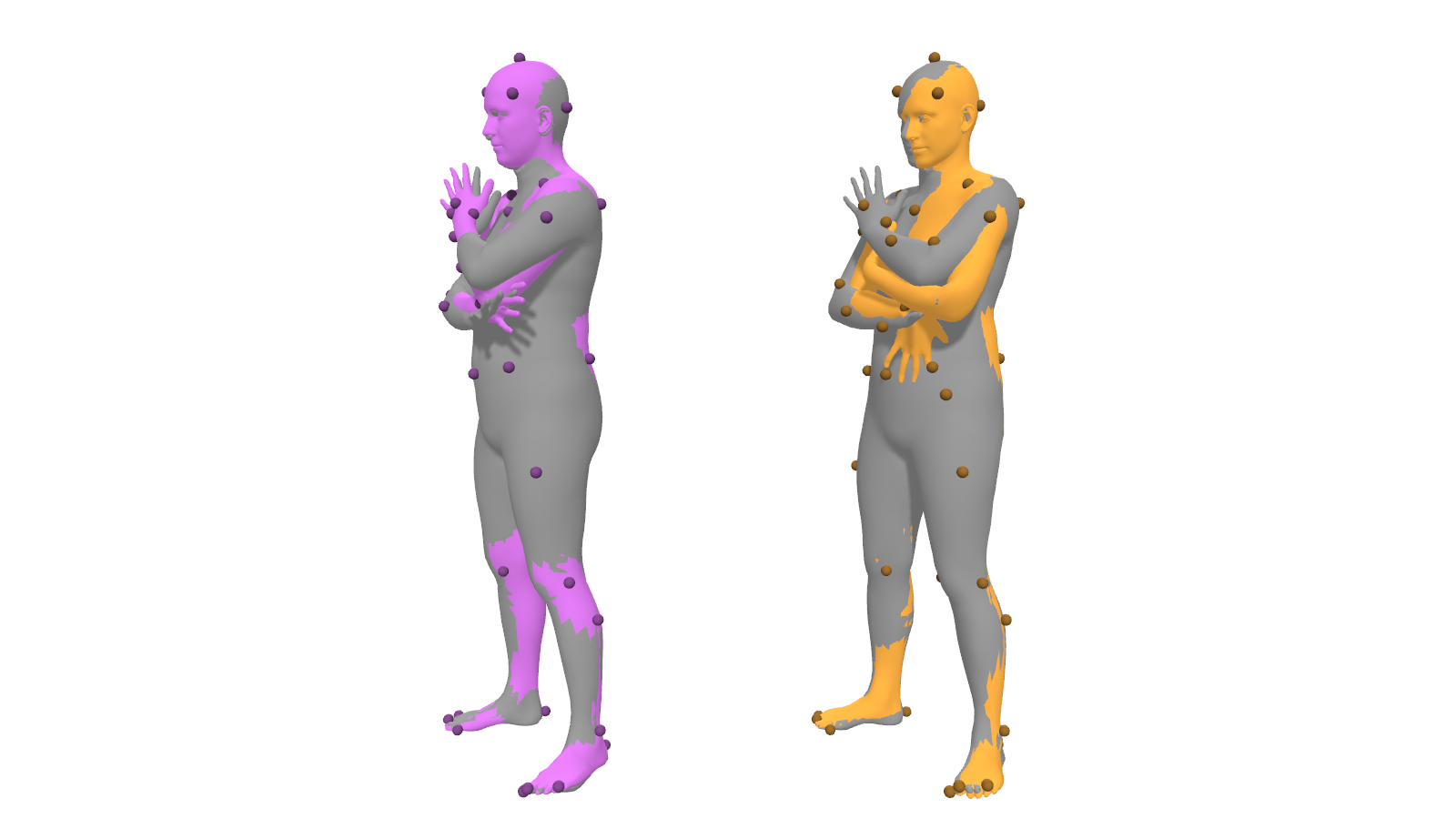}
\end{subfigure}\hfill
\begin{subfigure}{0.3\textwidth}
    \vspace{10pt}
    \includegraphics[width=\textwidth]{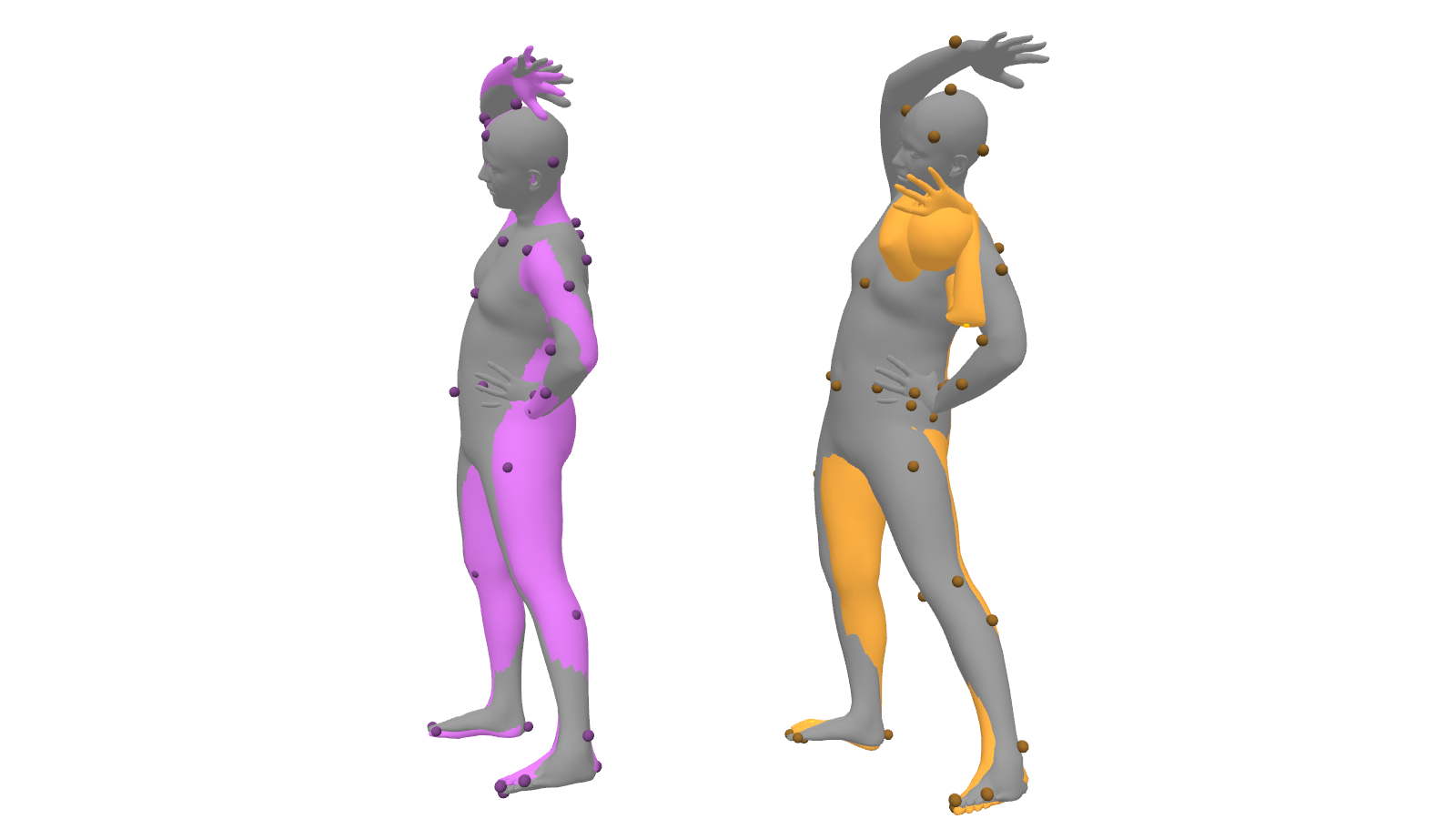}
\end{subfigure}

\begin{subfigure}{0.3\textwidth}
    \includegraphics[width=\textwidth]{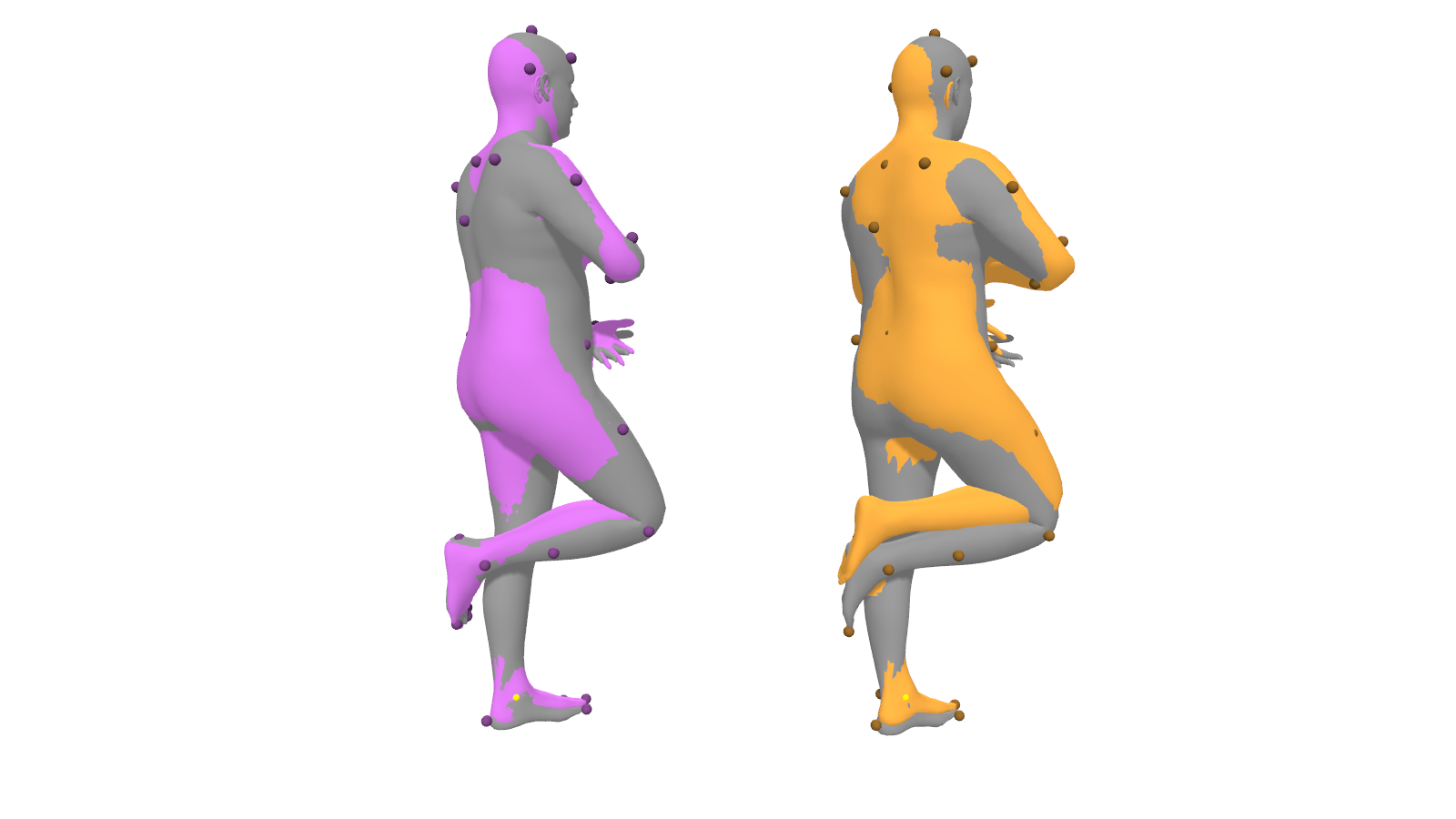}
\end{subfigure}\hfill
\begin{subfigure}{0.3\textwidth}
    \vspace{10pt}
    \includegraphics[width=\textwidth]{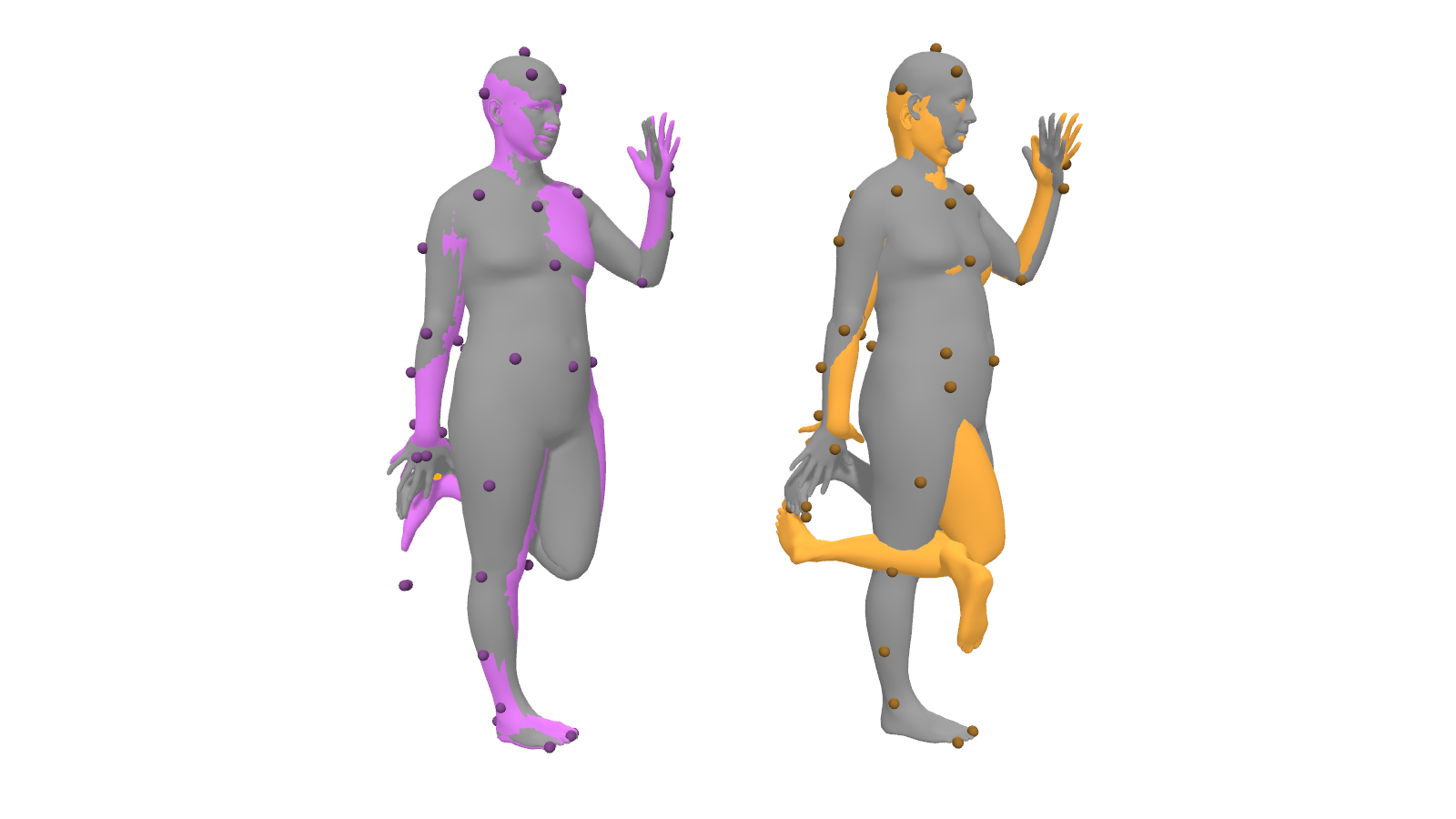}
\end{subfigure}\hfill
\begin{subfigure}{0.3\textwidth}
    \vspace{10pt}
    \includegraphics[width=\textwidth]{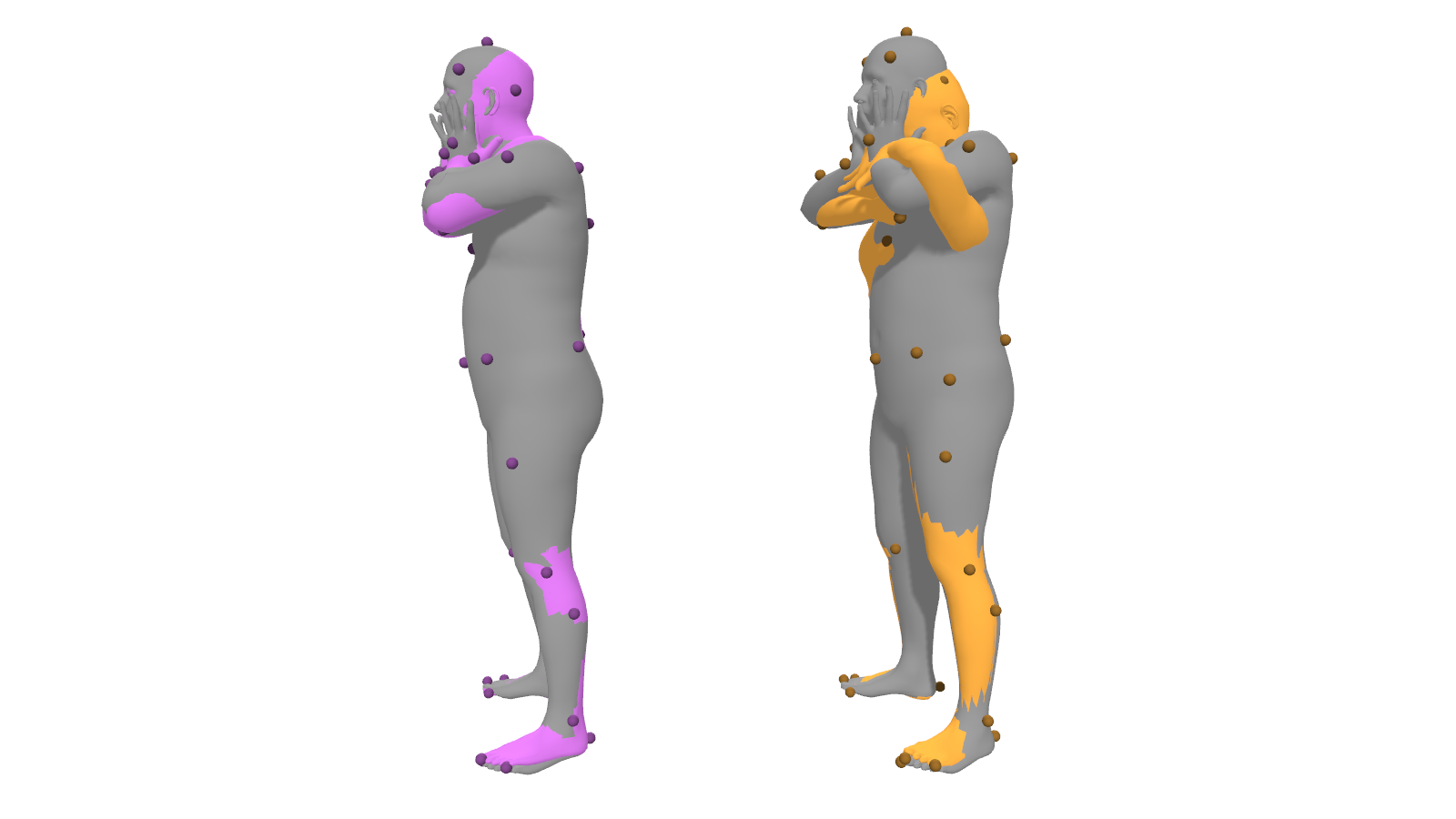}
\end{subfigure}
    
\caption{Fits to our \textcolor{ours}{regressed} versus SOMA \textcolor{soma}{labeled} markers. The fitting process is more sensitive to labeling errors.}
\label{fig:ours_vs_soma}

\end{figure*}

\begin{figure*}[!htp]

\begin{subfigure}{0.3\textwidth}
    \includegraphics[width=\textwidth]{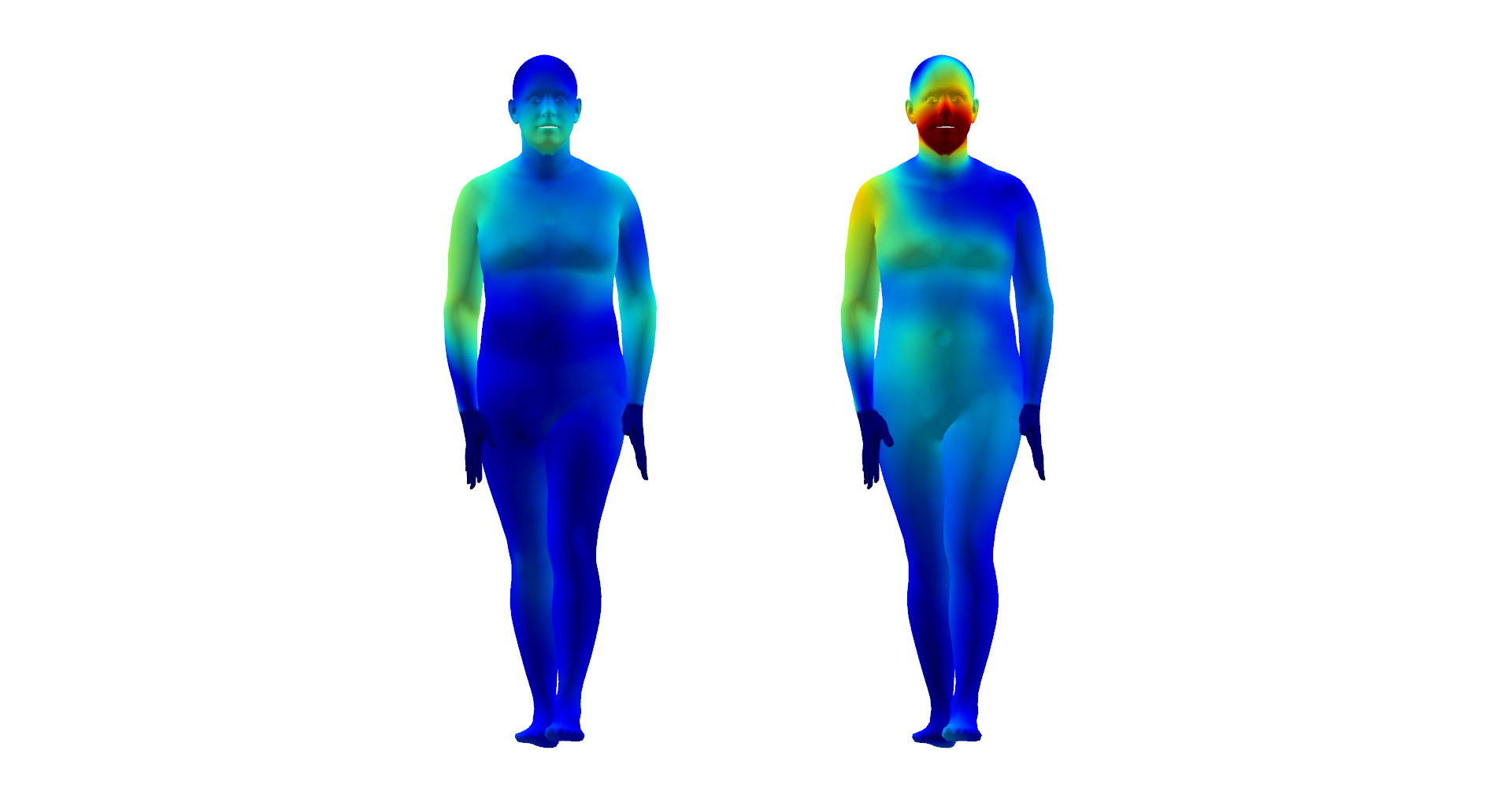}
\end{subfigure}\hfill
\begin{subfigure}{0.3\textwidth}
    \includegraphics[width=\textwidth]{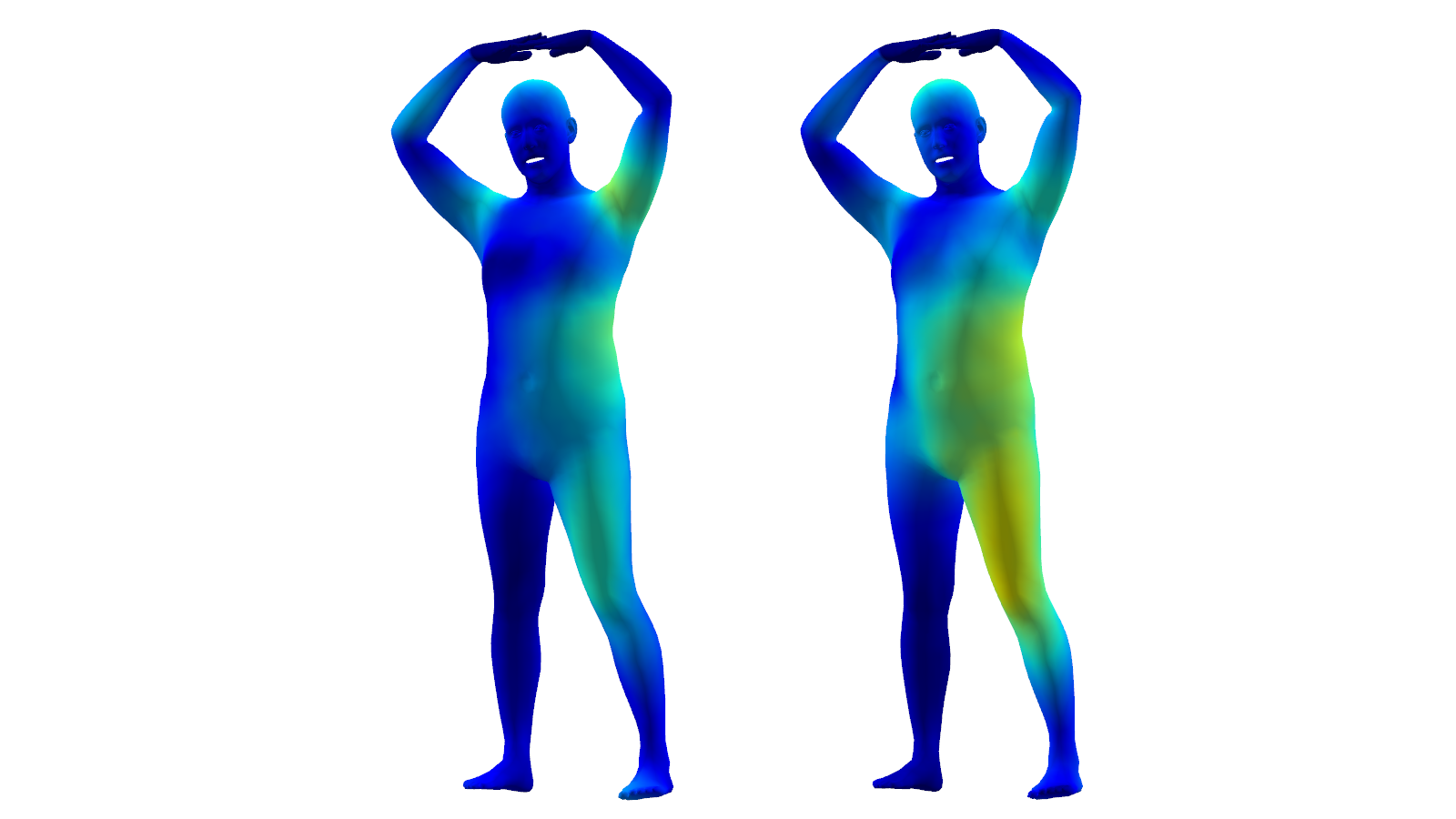}
\end{subfigure}\hfill
\begin{subfigure}{0.3\textwidth}
    \includegraphics[width=\textwidth]{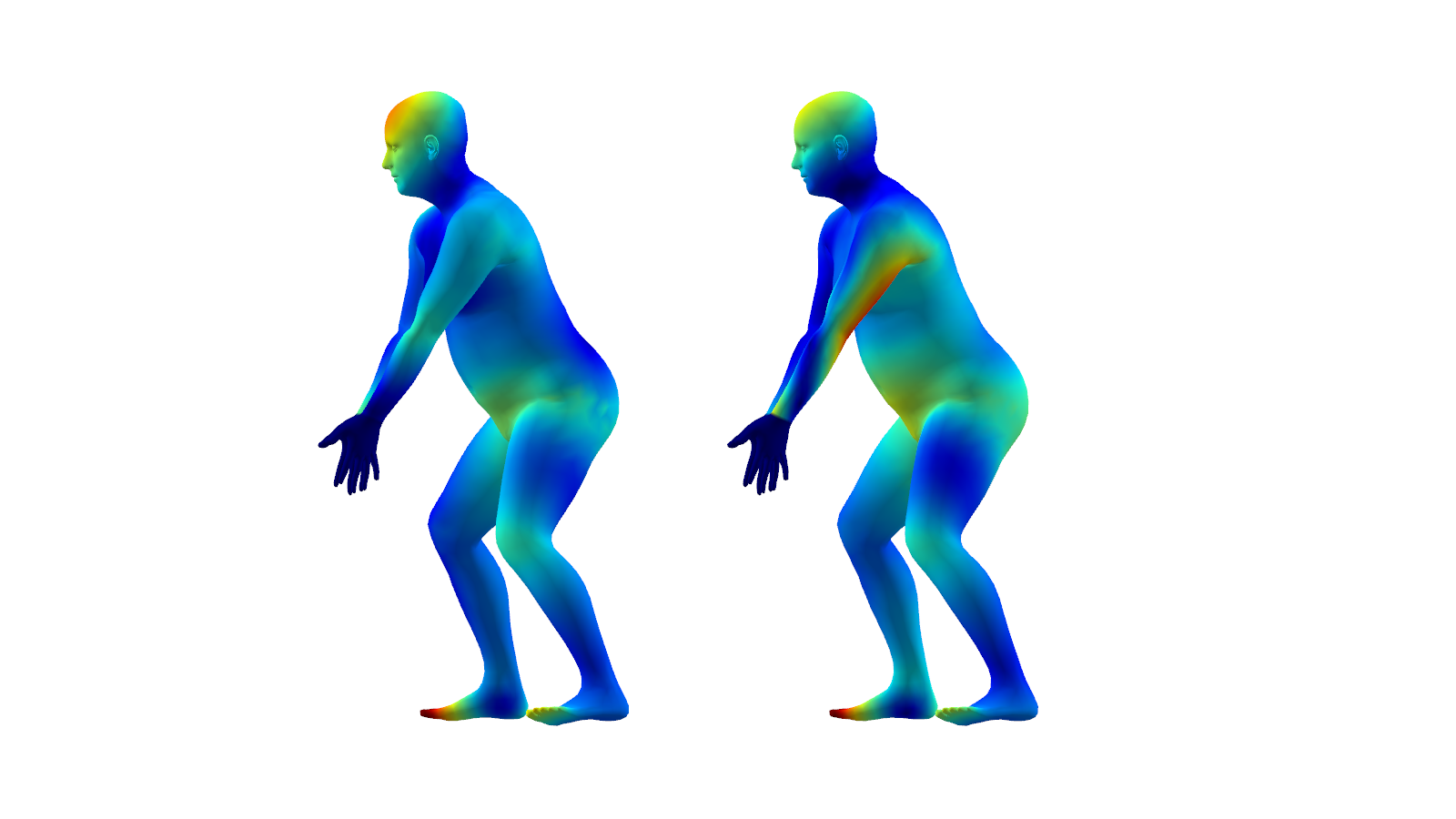}
\end{subfigure}\hfill

\begin{subfigure}{0.3\textwidth}
    \vspace{10pt}
    \includegraphics[width=\textwidth]{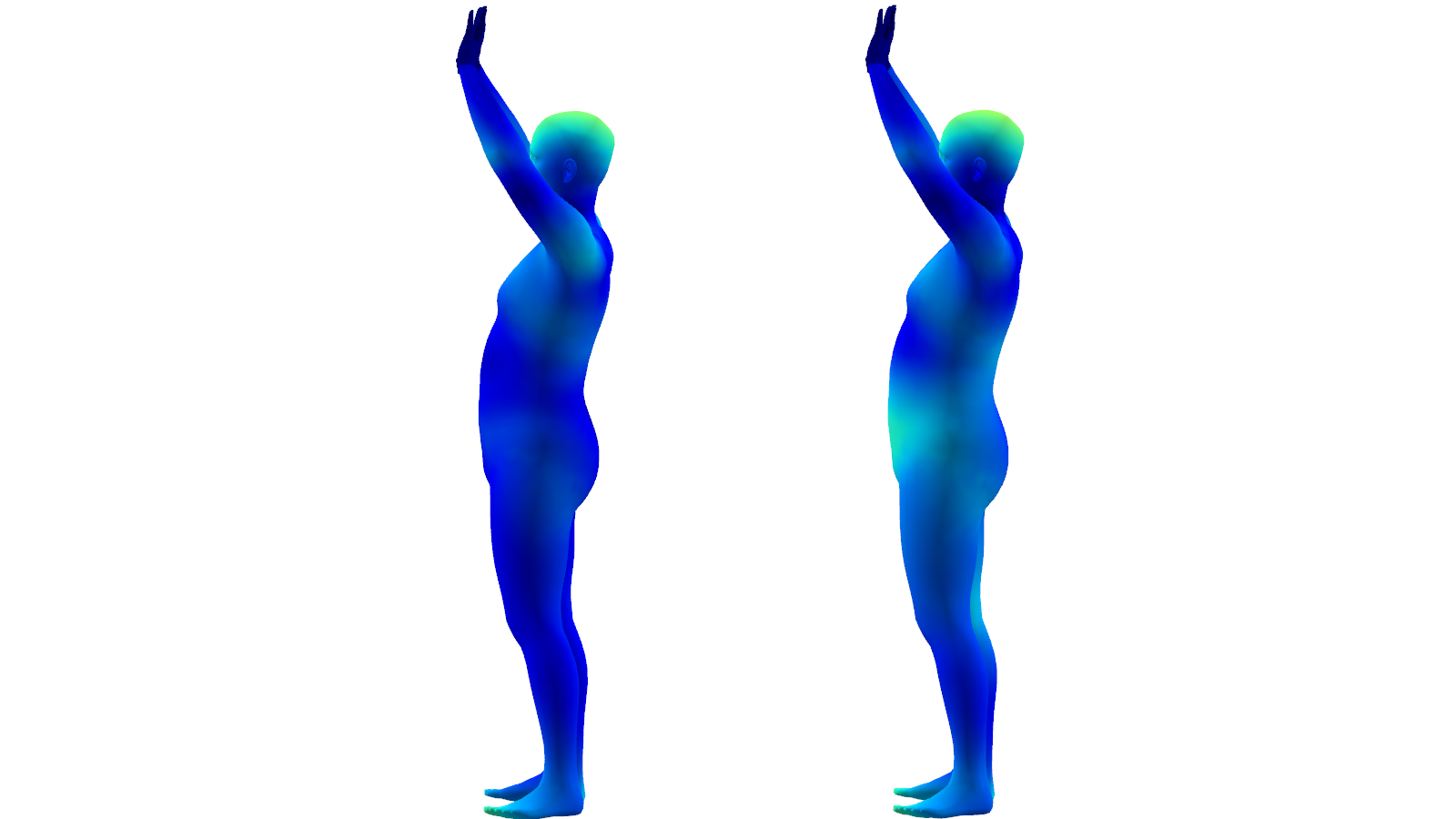}
\end{subfigure}\hfill
\begin{subfigure}{0.3\textwidth}
    \vspace{10pt}
    \includegraphics[width=\textwidth]{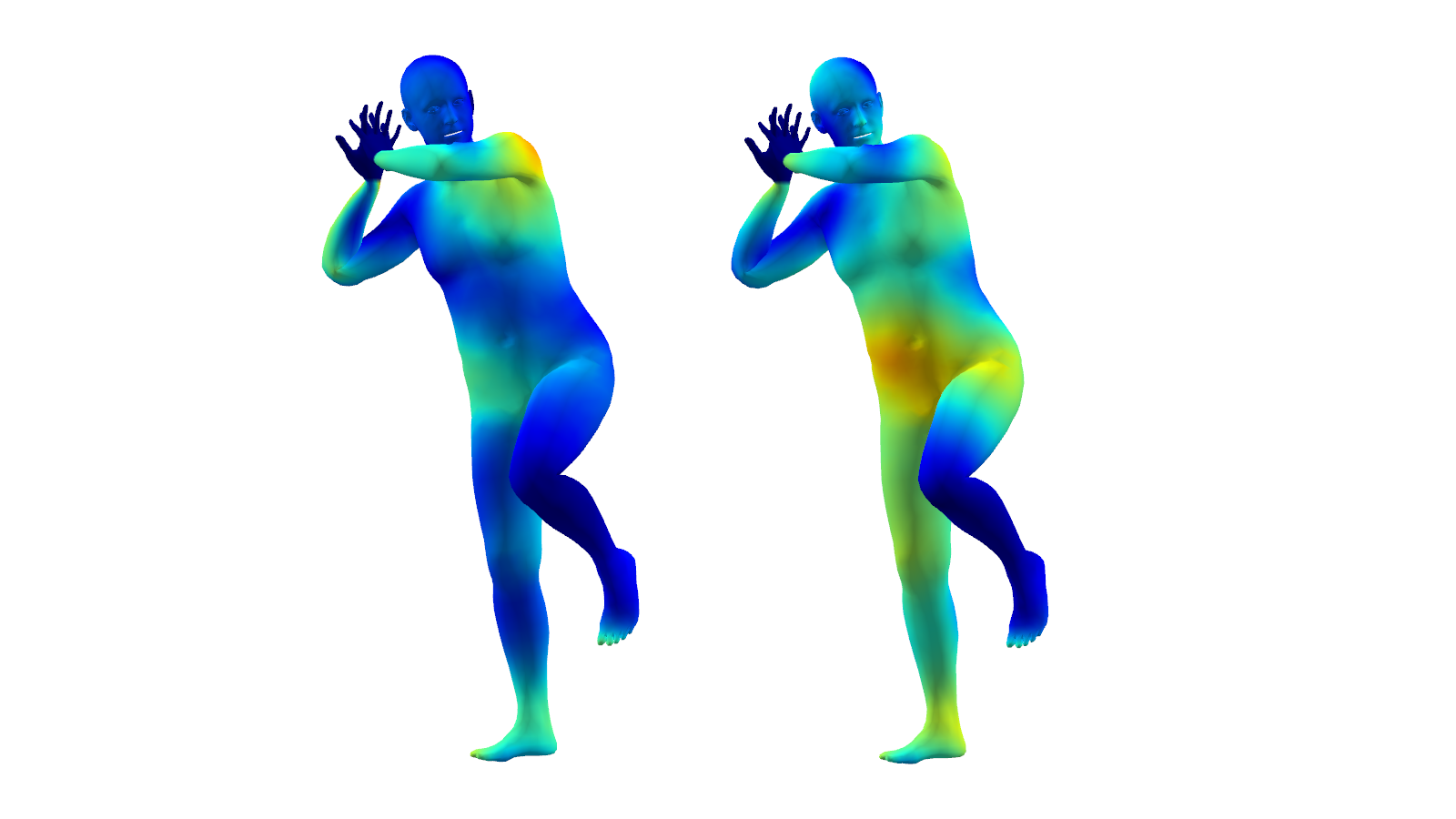}
\end{subfigure}\hfill
\begin{subfigure}{0.3\textwidth}
    \vspace{10pt}
    \includegraphics[width=\textwidth]{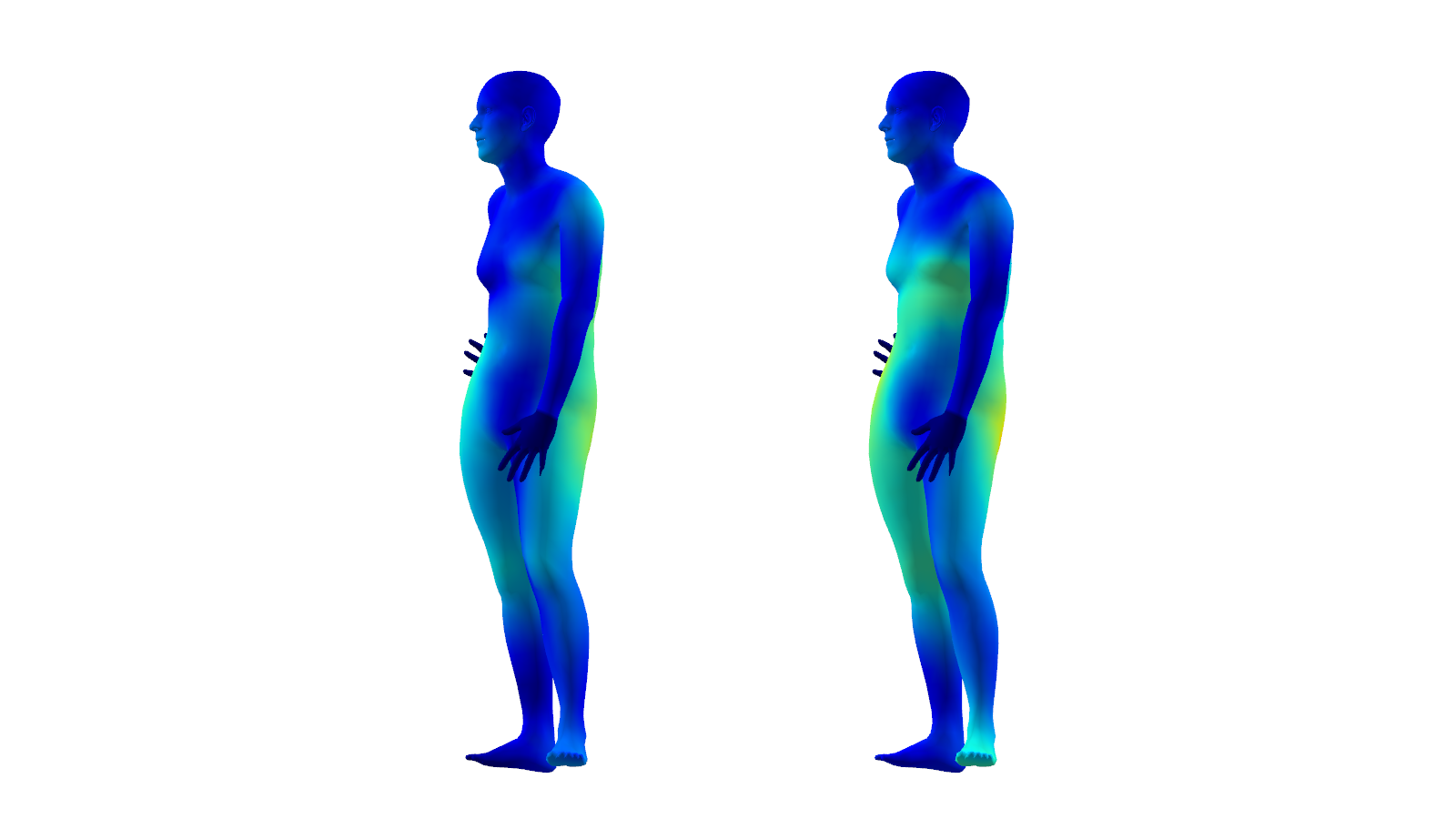}
\end{subfigure}\hfill

\begin{subfigure}{0.3\textwidth}
    \vspace{10pt}
    \includegraphics[width=\textwidth]{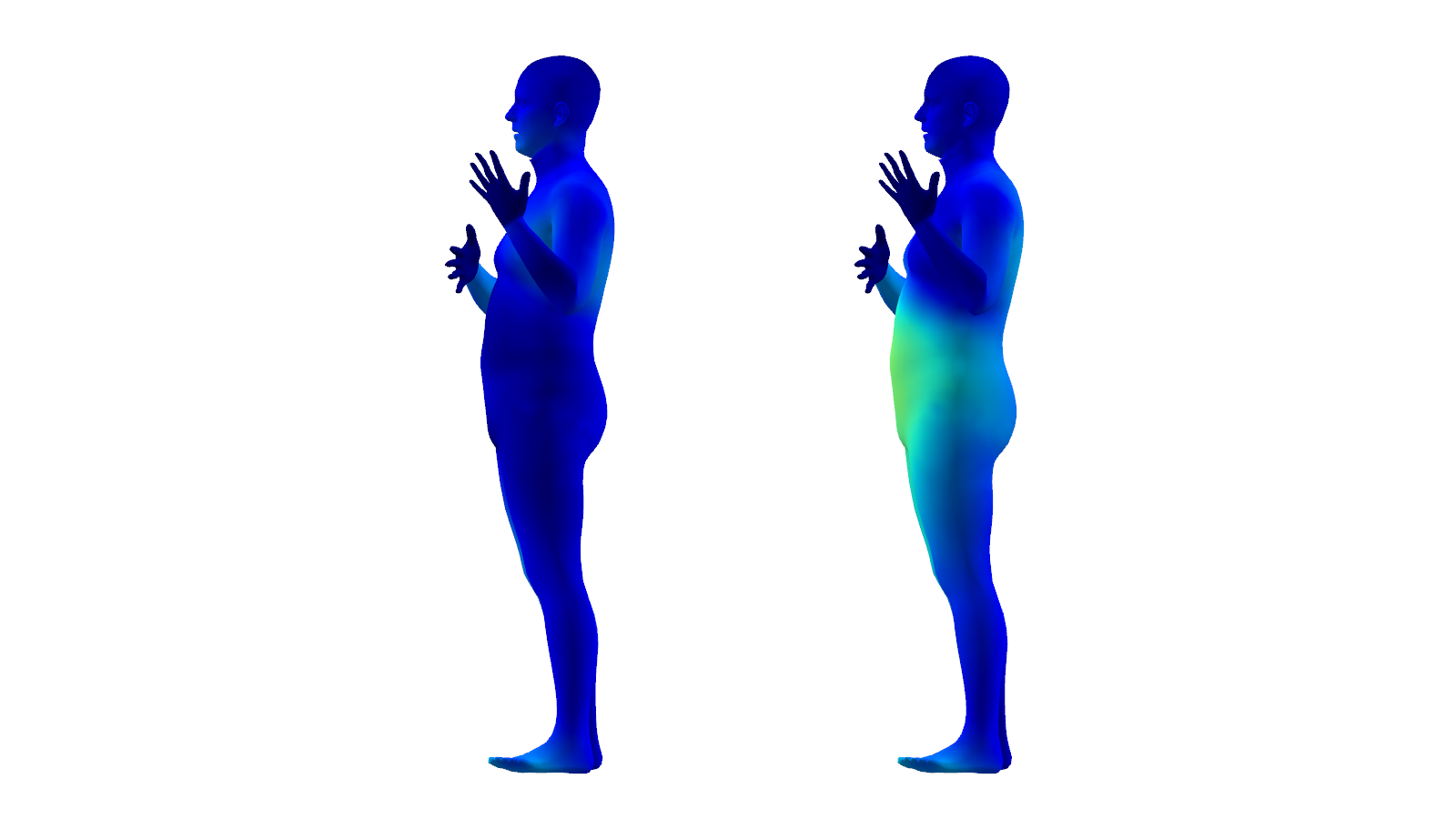}
\end{subfigure}\hfill
\begin{subfigure}{0.3\textwidth}
    \vspace{10pt}
    \includegraphics[width=\textwidth]{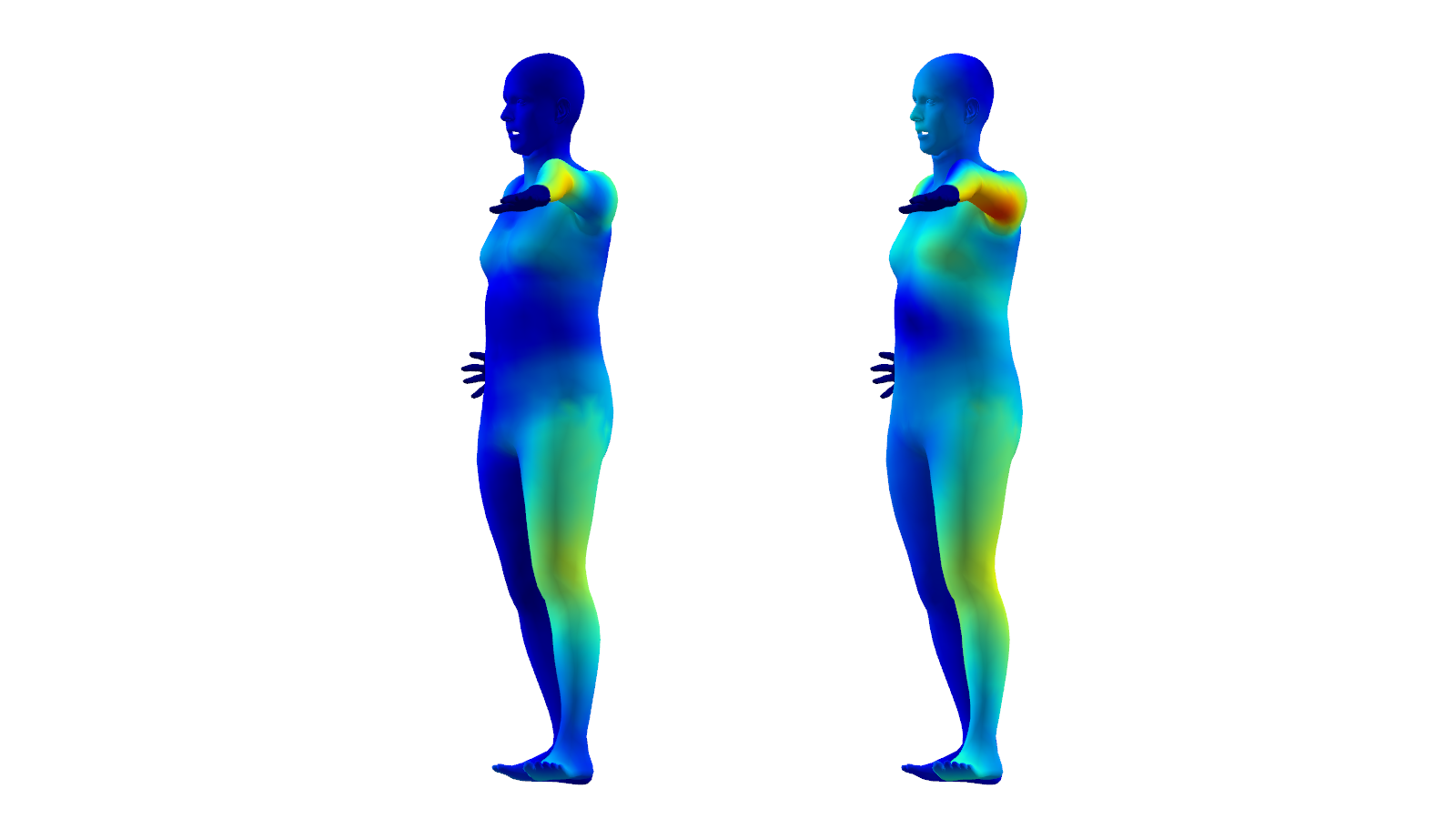}
\end{subfigure}\hfill
\begin{subfigure}{0.3\textwidth}
    \vspace{10pt}
    \includegraphics[width=\textwidth]{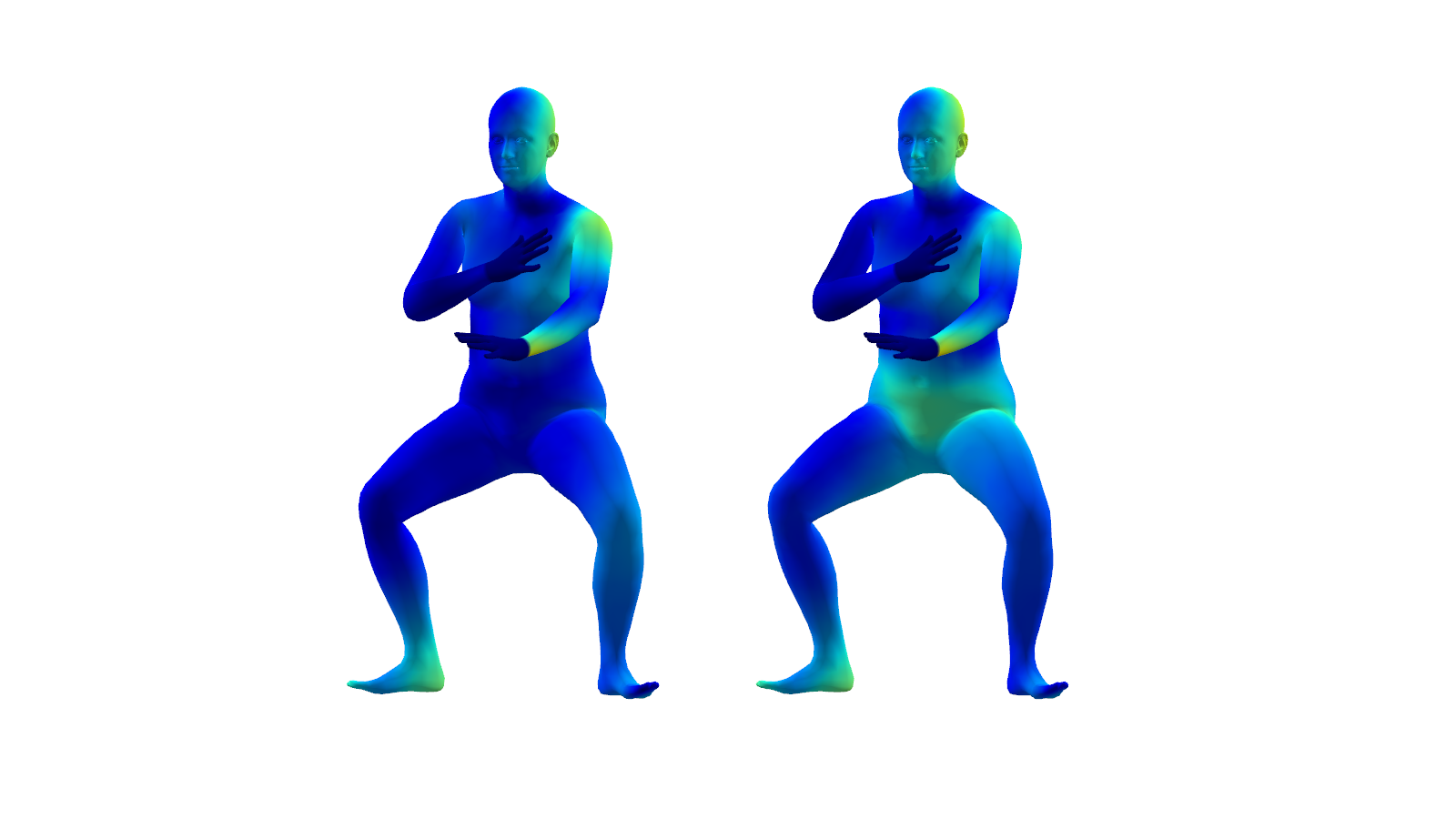}
\end{subfigure}

\begin{subfigure}{0.3\textwidth}
    \includegraphics[width=\textwidth]{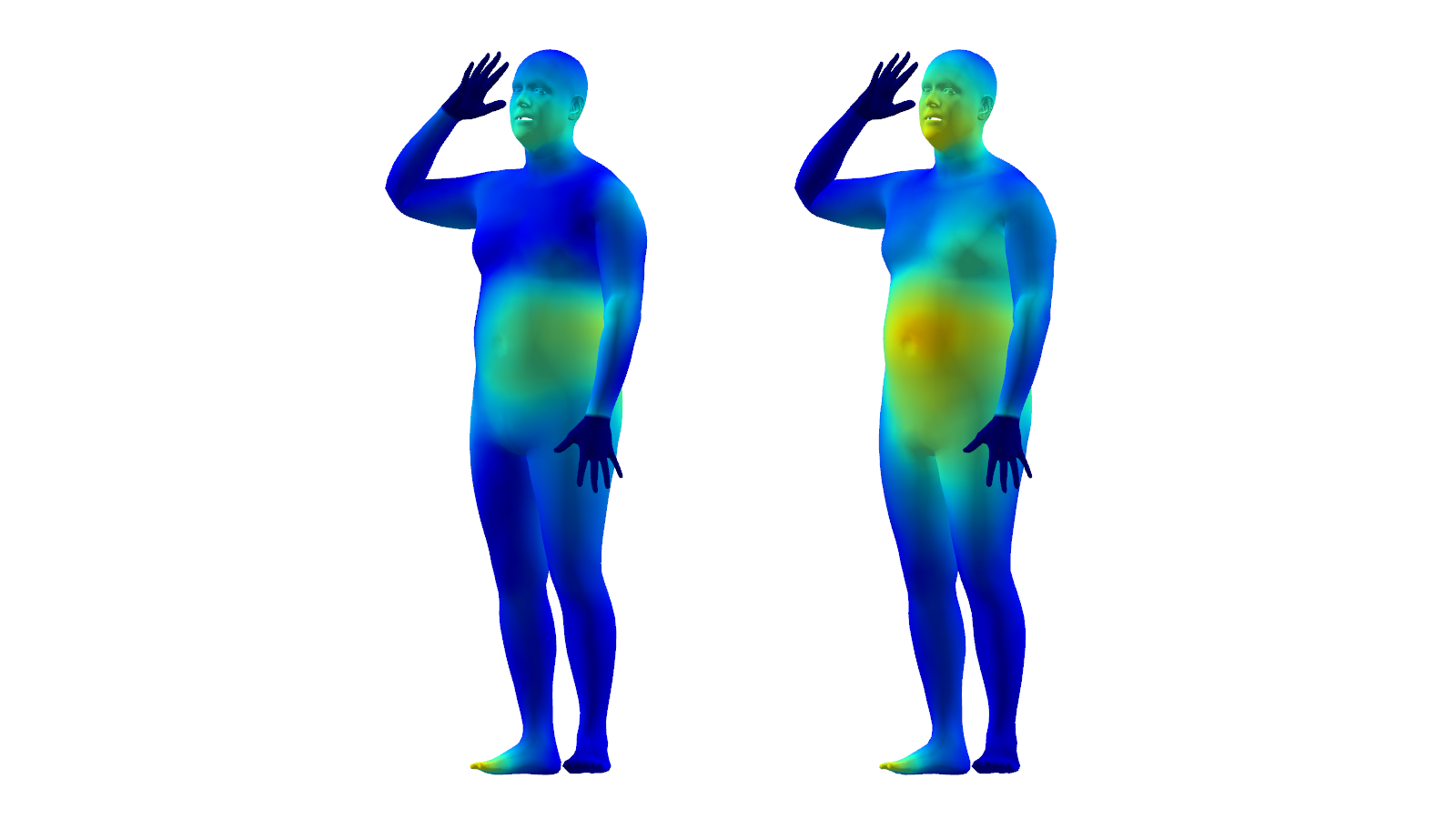}
\end{subfigure}\hfill
\begin{subfigure}{0.3\textwidth}
    \vspace{10pt}
    \includegraphics[width=\textwidth]{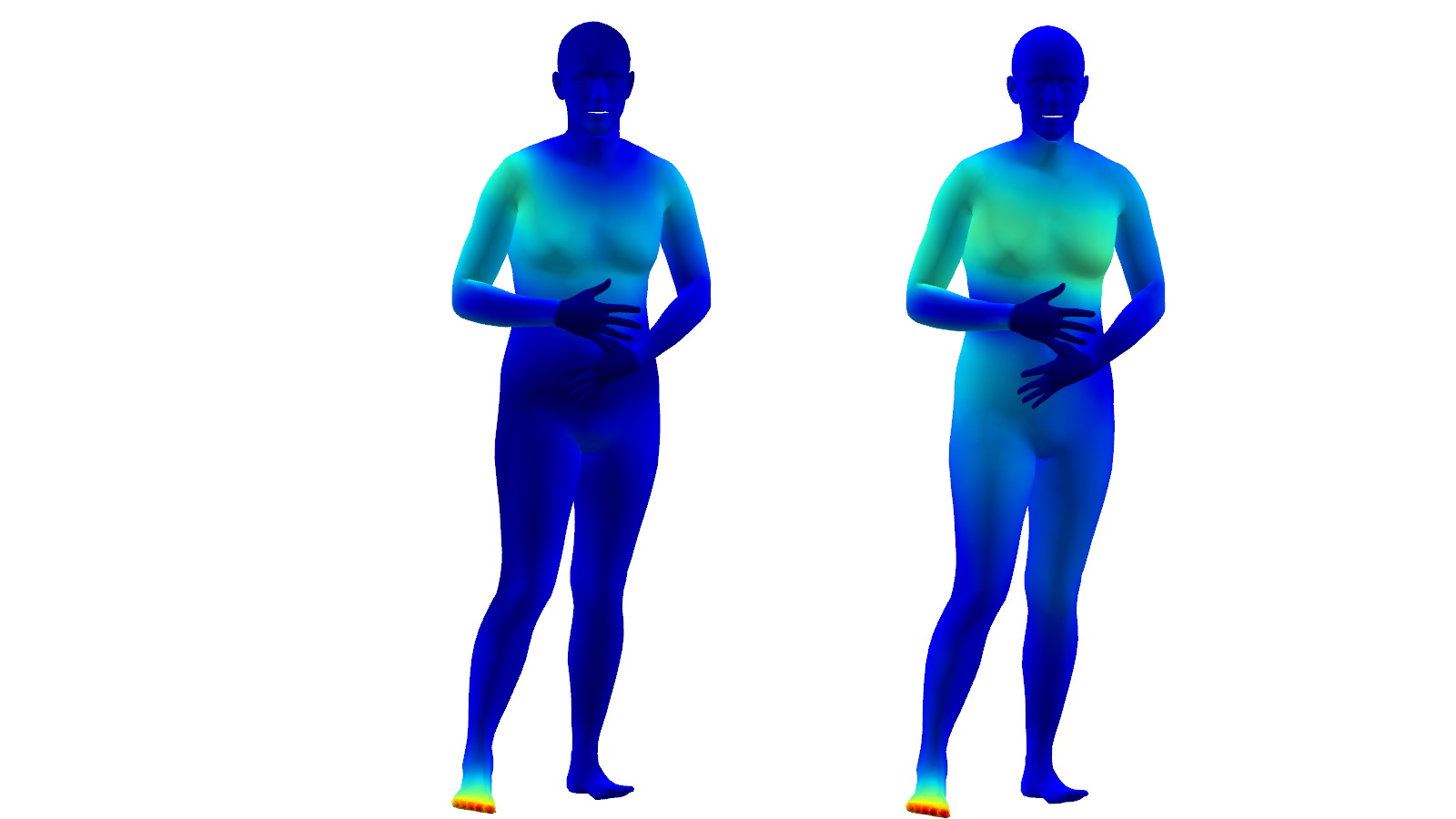}
\end{subfigure}\hfill
\begin{subfigure}{0.3\textwidth}
    \vspace{10pt}
    \includegraphics[width=\textwidth]{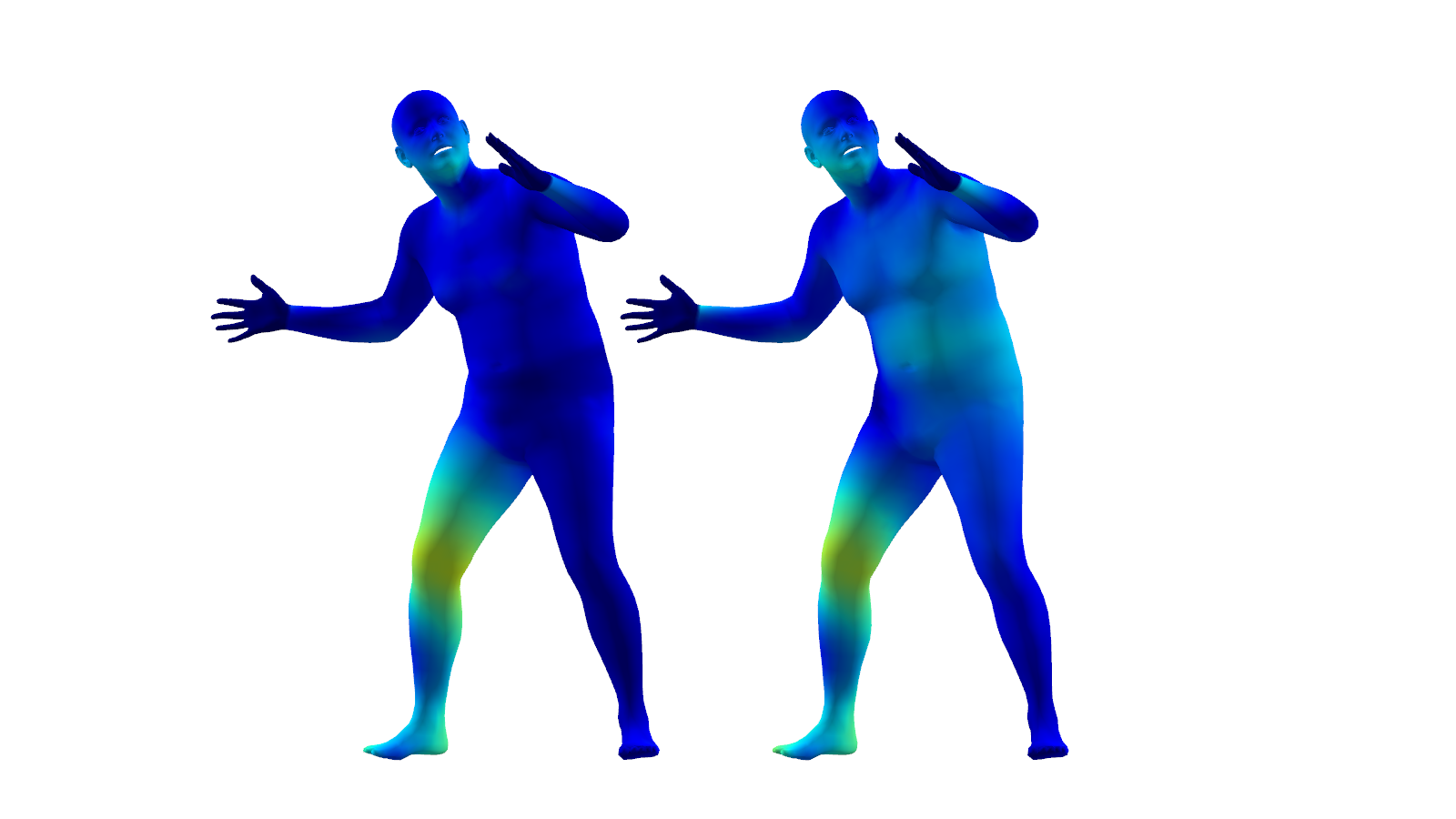}
\end{subfigure}
    
\caption{The figure shows the qualitative results of our noise-aware fitting method on the left and the method proposed in \cite{mosh} on the right. Each mesh is colored using a Jet color map based on the Euclidean distance error metric from the ground truth mesh.}
\label{fig:uncertainty_vs_naive}

\end{figure*}

\section{System Details}
\label{sec/sup:system}
We develop a multi-sensor acquisition system, equipped with 3 Microsoft Kinect for Azure depth sensors, to demonstrate our model's results in real-time.
The system connects $K$ hardware synchronized time-of-flight (ToF) sensors $k$, $k \in \{1, \dots , K\}$, spatially aligns them by performing extrinsic parameter calibration, and fuses the marker measurements in real-time, producing an unstructured point cloud $\mathbf{m} \in \mathbb{R}^{M \times 3}$, with $M$ being the number of marker estimates.

This process crucially relies on first acquiring 3D position marker measurements from a ToF sensor.
The sensor $k$ produces a stream of an infrared image $\mathbf{I}(\mathbf{p}) \in \mathbb{R}$ as well as a pixel-registered depth map $\mathbf{D}(\mathbf{p}) \in \mathbb{R}$, where each pixel $\mathbf{p} \in \mathbb{N}^{2}$ is defined in the image domain $\Omega := W \times H$ of width $W$ and height $H$ (the subscript $k$ is omitted for the sake of notational simplicity).
Using the factory calibrated intrinsic parameters of the sensor, the depth map is straightforwardly transformed to a structured point cloud $\mathbf{P} \in \mathbb{R}^3$, with $\mathbf{P}(\mathbf{p}) = \mathbf{K} \mathbf{G}(\mathbf{p}) \mathbf{D}(\mathbf{p})$, with $\mathbf{K} \in \mathbb{R}^{3 \times 3}$ being the intrinsic camera parameters matrix, and $\mathbf{G} \in \mathbb{N}^3$ the homogeneous coordinates image grid.

We exploit this one-to-one mapping between the infrared image $\mathbf{I}$ and the structured point cloud $\mathbf{P}$ to extract the marker positions $\mathbf{m}_k$.
Relying on the retro-reflective properties of markers that return the light emitted by the ToF projector, we identify the marker pixels after applying binary thresholding and contour detection \cite{suzuki1985topological} on the infrared image.
While measurements are undefined on the actual marker position due to the ToF depth estimation principles, we observe that the measurements around the actual marker position are well-defined.
Thus, for each contour we sample the structured point cloud to extract a point measurement, aggregating them into a vector $\mathbf{v} \in \mathbb{R}^{V \times 3}$, with $V$ being the number of the contour points.
As spurious outliers can be included in this vector due to fore/background issues and imperfect pixel sampling, we perform Median Absolute Deviation (MAD) outlier rejection \cite{leys2013detecting} using the $z$-coordinate (depth) of each point, and the average the remaining points to extract the final marker position estimates $\mathbf{m}_k$.

Using  $\mathbf{m}_k$, the system calibrates the sensors by running bundle adjustment using a simple calibration wand with a marker attached to a stick.
Then, gravity alignment is achieved by placing $3$ markers in a $\Gamma$ shape on the floor and extracting the long and short edge cross product as the up vector, transforming all extrinsic transforms to align with it.
With the sensors spatially aligned, all marker estimates are fused in a single unstructured point cloud $\mathbf{m}$.
To account for slight calibration errors, we perform point cloud clustering with a radius of $1cm$, which results in the actual model input.
Evidently, this process is a cascade of numerous estimation errors, the inherent measurement noise that influences the calibration process, and the clustering itself which also adjusts the final estimates.
Additionally, we only use $K=3$ sensors, which accentuates the problem since information fusion is not that effective with such a sparse number of viewpoints. 

\begin{figure*}[!htbp]
\centering
\begin{subfigure}{0.9\textwidth}
    \vspace{-50pt}
    \centering\includegraphics[width=\textwidth]{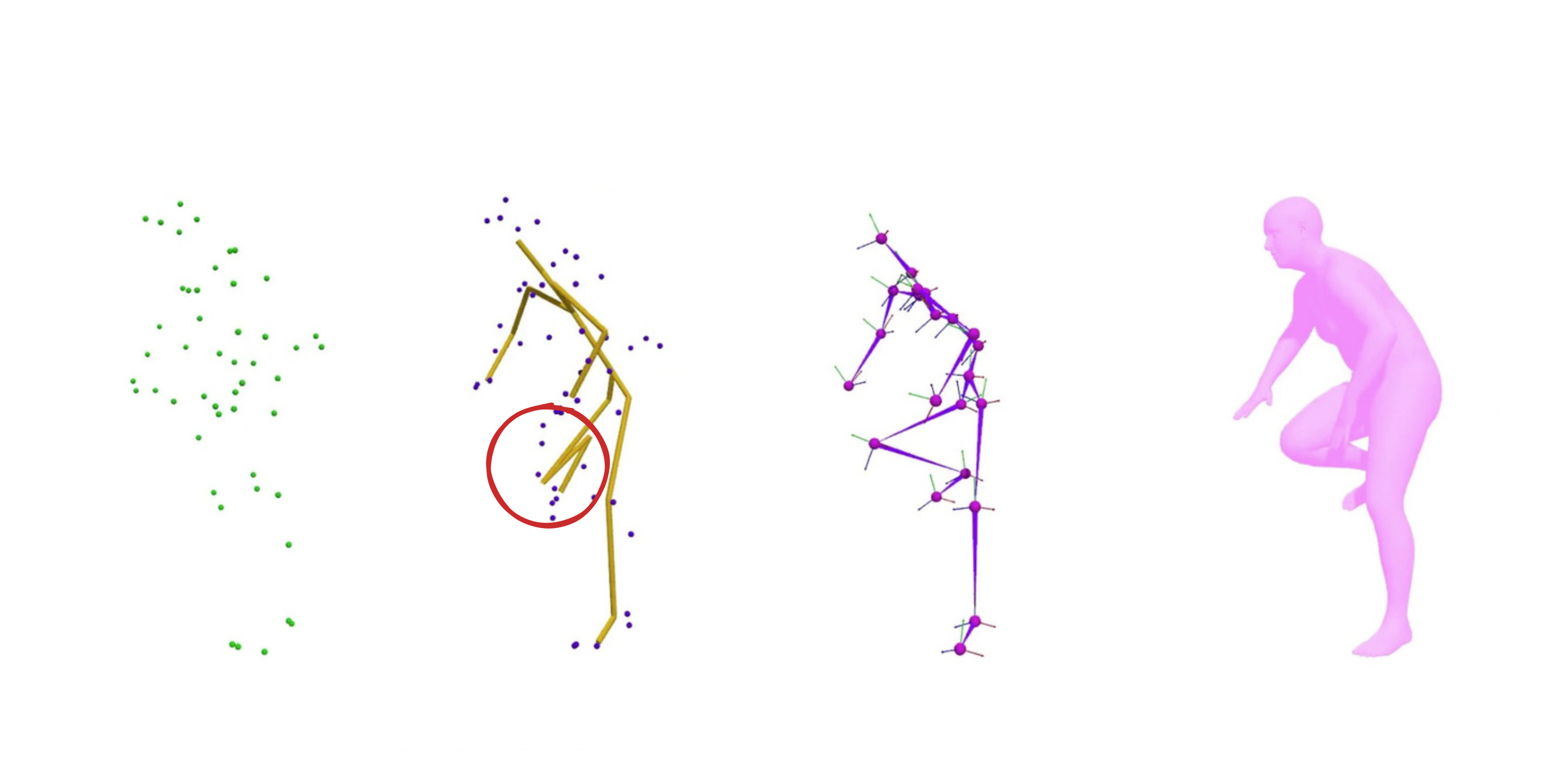}
\end{subfigure}\hfill
\begin{subfigure}{0.9\textwidth}
    \vspace{-30pt}
    \centering\includegraphics[width=\textwidth]{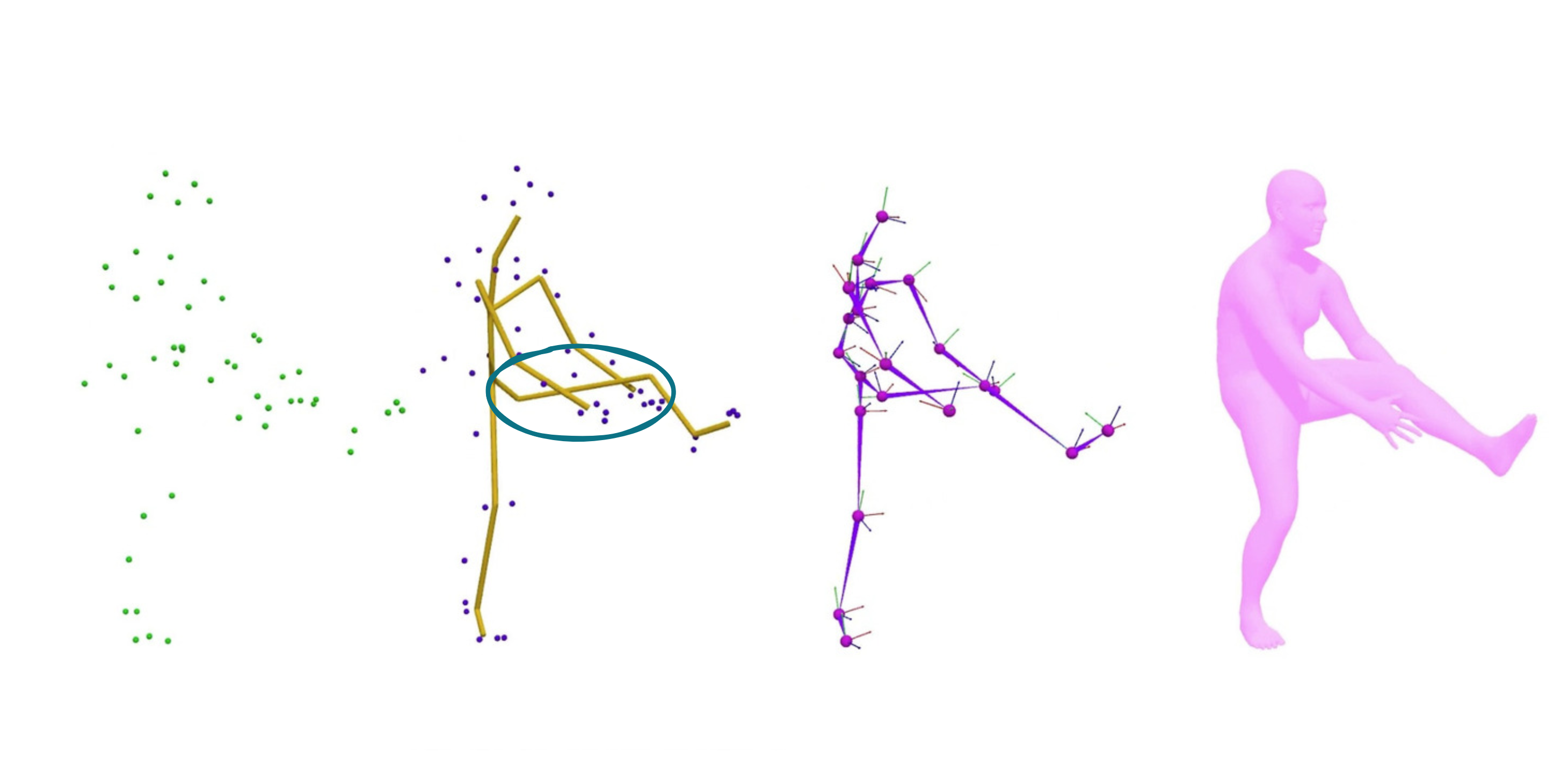}
\end{subfigure}\hfill
\begin{subfigure}{0.9\textwidth}
    \vspace{-20pt}
    \centering\includegraphics[width=\textwidth]{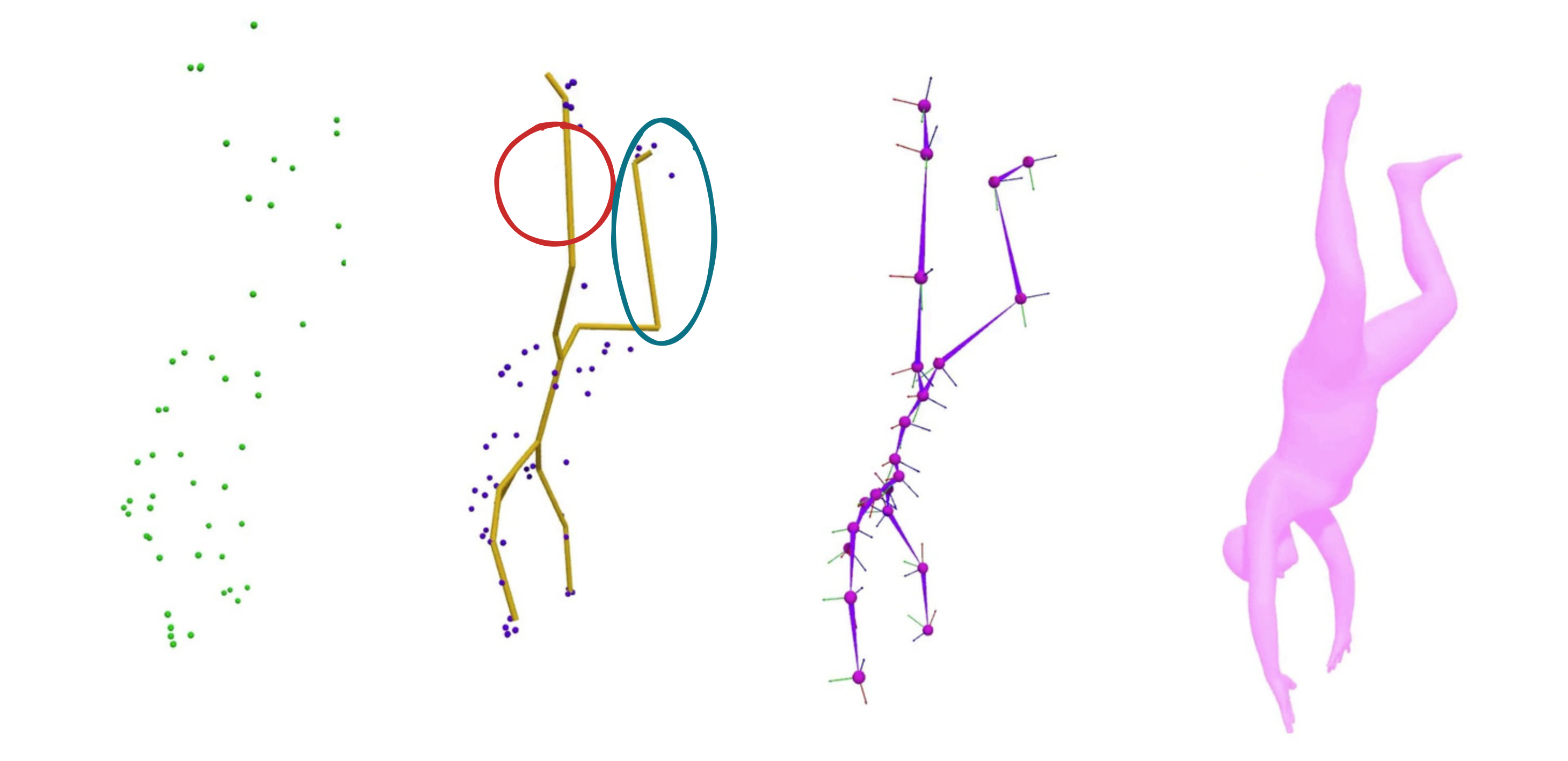}
\end{subfigure}\hfill
    
\caption{
Additional qualitative results of our system in the wild using a setup comprising a very sparse set of low-cost sensors.
Starting from the left, we present the raw input collected from our multi-sensor 
acquisition system (\cref{sec/sup:system}), with the raw (unfiltered) estimated $\landmarks_{est}$ from our model following.
The last 2 columns present the fitted $\theta_{est}$ pose and shape $\beta_{est}$ parameters. 
As our real-time model only implicitly learns the human skeleton, this can lead to \textcolor{unrealistic_results}{unrealistic results}. 
To address this, the noise-aware fitting approach introduces human body constraints, resulting in more accurate and realistic results. 
Furthermore, it adequately handles \textcolor{missing}{missing} or \textcolor{missing}{incorrectly} inferred landmarks.
}
\label{fig:in_the_wild_qualitative}
\end{figure*}

\end{document}